\documentclass[10pt,letterpaper]{article}

\usepackage[
    letterpaper,
    top=1in,
    bottom=1in,
    left=1in,
    right=1in
]{geometry}

\usepackage[T1]{fontenc}
\usepackage{lmodern}
\usepackage{microtype}
\usepackage{textcomp}
\usepackage{fancyvrb}

\usepackage{amsmath,amssymb,amsfonts}

\usepackage{graphicx}
\graphicspath{{./figures/}}

\usepackage[table]{xcolor}
\usepackage{soul}

\usepackage{booktabs}
\usepackage{multirow}
\usepackage{threeparttable}
\usepackage{tabularx}
\usepackage{longtable}
\usepackage{array}
\usepackage{makecell}

\usepackage{float}
\usepackage{caption}
\usepackage{placeins}

\usepackage{algorithm}
\usepackage{algpseudocode}

\floatname{algorithm}{Algorithm}
\renewcommand{\thealgorithm}{\arabic{algorithm}}

\usepackage{enumitem}

\usepackage[most]{tcolorbox}

\usepackage{cite}
\usepackage{url}
\usepackage[hidelinks]{hyperref}

\usepackage{indentfirst}
\usepackage{fancyvrb}
\usepackage{listings}

\lstdefinestyle{qaexample}{
    basicstyle=\ttfamily\footnotesize,
    breaklines=true,
    breakatwhitespace=false,
    columns=fullflexible,
    keepspaces=true,
    showstringspaces=false,
    frame=none
}

\definecolor{kgplain}{RGB}{255,249,219}   
\definecolor{kgrepair}{RGB}{255,235,214}  
\definecolor{cgplain}{RGB}{226,242,255}   
\definecolor{cgrepair}{RGB}{226,246,234}  

\newcommand{\entitytext}[1]{\sethlcolor{yellow!25}\hl{#1}}
\newcommand{\reltext}[1]{\sethlcolor{green!18}\hl{#1}}

\newcommand{\entitymath}[1]{\colorbox{yellow!25}{\ensuremath{\textit{#1}}}}
\newcommand{\relmath}[1]{\colorbox{green!18}{\ensuremath{\textit{#1}}}}

\usepackage[most]{tcolorbox}
\usepackage{xcolor}
\usepackage{enumitem}
\usepackage{amsmath}

\newcommand{\triple}[3]{%
\ensuremath{\langle \textit{#1},\ \textit{#2},\ \textit{#3} \rangle}%
}

\newcommand{\triplehl}[3]{%
\ensuremath{\langle \entitymath{#1},\ \relmath{#2},\ \entitymath{#3} \rangle}%
}

\newcommand{\gainlow}{\textcolor{yellow!70!orange}{\scriptsize$\uparrow$}}
\newcommand{\gainmid}{\textcolor{green!60!black}{\scriptsize$\uparrow\uparrow$}}
\newcommand{\gainhigh}{\textcolor{green!60!black}{\scriptsize$\uparrow\uparrow\uparrow$}}

\title{
\textbf{Repair Before Reinforce: Context-Augmented Knowledge Graph Reasoning
for Multi-Hop 
Question Answering}
}

\author{
Tharaka D. Fonseka and Niraj K. Jha\\
{\small Dept. of Electrical and Computer Engineering, Princeton University}
}

\date{}
\makeatletter
\def\bstctlcite#1{\@bsphack
  \@for\@citeb:=#1\do{%
    \edef\@citeb{\expandafter\@firstofone\@citeb}%
    \if@filesw\immediate\write\@auxout{\string\citation{\@citeb}}\fi}%
  \@esphack}
\makeatother
\begin{document}

\bstctlcite{IEEEexample:BSTcontrol}
\maketitle


\begin{abstract}

Question-answering often requires reasoning across multiple connected 
facts rather than retrieving a single isolated relation. Knowledge graphs (KGs) provide a structured way to represent such facts, but training large language models (LLMs) only on isolated KG head-relation-tail triples may limit their ability to learn the surrounding 
context needed for multi-hop reasoning. In this work, we propose a context-augmented training framework for 
multi-hop question-answering. Although generally applicable, we validate the framework in the context of disease-specific KGs, extracted using a reliable KG extraction framework called GraphMERT, for Gastroparesis and Diabetes. For each primary KG triple, we attach supporting triples extracted from the same source text chunk to form a context graph (CG). This creates two supervision settings: KG-grounded supervision, which uses only the target KG triple or path, and CG-grounded supervision, which uses the target KG triple or path together with supporting context triples.

We train the Qwen3-14B model using supervised fine-tuning (SFT) under both settings, producing KGModel and CGModel variants. To strengthen the lower-hop factual foundation of the models,
we introduce an LLM-judged, history-aware adaptive repair pipeline that identifies unresolved one-hop failures, continually fine-tunes on targeted repair examples, and removes or quarantines problematic noisy triples. This repair stage enables the models to reach 100\% accuracy on the cleaned retained one-hop validation sets. Finally, we employ reinforcement learning (RL) using lower-hop question-answer items and evaluate generalization on harder 3-hop, 4-hop, and 5-hop tasks.

Across both diseases, context-augmented supervision consistently improves multi-hop performance over KG-only supervision. RL initialized from repaired SFT checkpoints yields larger and more stable gains than RL initialized from unrepaired checkpoints. Overall, the strongest results are achieved by the context-augmented, repaired, and RL-trained models, demonstrating that reliable 
multi-hop reasoning benefits from both richer context and a repaired lower-hop knowledge foundation.


\end{abstract}

\section{Introduction}

Large language models (LLMs) have achieved substantial progress in natural language understanding, instruction-following, and question-answering, yet 
question-answering remains a challenging setting as it requires 
grounded and knowledge-intensive reasoning~\cite{KnowledgeIntensiveMultiHopQA,LLM1,LLM2,LLM3,BioGPT}. In the biomedical (and many other) contexts, questions often cannot be answered from a single isolated fact. Instead, they require connecting diseases, symptoms, mechanisms, treatments, anatomical sites, and clinical findings across multiple pieces of evidence~\cite{MultihopQA1,MultihopQA2,medhop}. This makes multi-hop reasoning especially important in settings, where a correct answer may depend on composing several related facts into a coherent reasoning chain~\cite{MultihopQA1,MultihopQA2,BioHop,medhop,MisorderedContext}.

To improve the reliability of multi-hop 
reasoning, knowledge graphs (KGs) have played an important role by organizing 
knowledge into structured and interpretable facts. A KG encodes relations as triples of the form $(h,r,t)$, where a head entity $h$ is connected to a tail entity $t$ through a relation $r$~\cite{KGSurvey}. For example, diabetes mellitus is a well-established underlying cause of gastroparesis and metoclopramide is used as a pharmacologic treatment for gastroparesis; these clinical facts can be represented in a biomedical KG as $(\textit{diabetes mellitus}, \textit{cause\_of}, \textit{gastroparesis})$ and $(\textit{metoclopramide}, \textit{treats}, \textit{gastroparesis})$, respectively~\cite{NIDDKGastroparesis,NIDDKGastroparesisTreatment}. Large biomedical resources, such as the Unified Medical Language
System (UMLS)~\cite{UMLS}, organize biomedical concepts and relations across many vocabularies making them widely used foundations for structured biomedical knowledge representation. Prior KG-based reasoning systems have shown that structured knowledge can complement language-model representations. For example, QA-GNN connects questions and candidate answers to relevant KG nodes and reasons over a joint language-graph representation~\cite{QA-GNN}. More recent work has also used KGs as structured sources of supervision rather than only as retrieval resources. KG-SFT uses external KGs to construct reasoning subgraphs for supervised fine-tuning (SFT)~\cite{kgsft}. Dedhia et al.~synthesize KG-grounded reasoning curricula from medical KG primitives~\cite{DedhiaSI} and Stephen et al.~generate curriculum-based question-answer (QA) items from a neuroscience KG~\cite{JakePaper}. BioKGQA further supports this research direction by constructing a biomedical KG question-answering dataset from PrimeKG, demonstrating that structured biomedical KGs can serve as foundations for generating QA benchmarks and supervision~\cite{biokgqa,primekg}. These works suggest that reliable KGs can serve as data-generating substrates for domain-specific language-model training.

Despite the above advantages, isolated KG triples may be insufficient for robust reasoning. A single triple may state that a relation exists, but it often omits the surrounding evidence needed to interpret that relation correctly. For example, a disease-symptom relation may be more meaningful when considered together with related mechanisms, anatomical sites, clinical findings, risk factors, or associated conditions. This is especially important in biomedical question-answering, where answering a question often requires not only recognizing a target relation, but also understanding how that relation fits within a local clinical context. This observation motivates context-augmented KG supervision. Instead of treating each triple as an independent fact, we ask whether a model can learn more useful representations when each target triple is presented together with supporting evidence from its source context. Based on the broader concept of context graphs (CGs) 
\cite{CGPaper}, we augment each primary triple with supporting triples extracted from the same source text chunk. The resulting CG preserves the structured form of a conventional KG while adding local evidence around each target fact.

Building on this motivation, we construct disease-specific biomedical KGs for Gastroparesis and Diabetes using GraphMERT, a recent framework for automatically distilling reliable domain-specific KGs from unstructured biomedical text~\cite{margarita}. We then augment these KGs by attaching same-chunk supporting triples to each primary triple, producing CGs for the two disease settings. From these KGs and CGs, we generate two types of QA items. In the KG-grounded setting, each item is generated from only the target triple or reasoning path. In the CG-grounded setting, each item is generated from the target triple or path together with its supporting context triples. This design enables us to directly compare whether context-augmented supervision provides a stronger training signal than standard KG-grounded supervision.

We then study this comparison across three training stages. First, we fine-tune a base LLM using either KG-grounded or CG-grounded QA items, producing KGModel and CGModel variants. Second, we introduce an LLM-judged adaptive repair stage inspired by skill-targeted adaptive training (STAT)~\cite{STAT}. This stage checks whether the initial SFT models still contain unresolved one-hop missing knowledge and applies targeted adaptive repair before employing reinforcement learning (RL). Finally, we investigate whether RL is more effective when initialized from a repaired SFT checkpoint rather than from an unrepaired SFT checkpoint.



This final stage is motivated by recent work on the complementary roles of SFT and RL. Prior work has argued that ``SFT memorizes, RL generalizes,'' suggesting that SFT learns task format and imitation patterns, whereas RL better encourages generalizable behavior~\cite{SFTMemorizesRLGeneralizes}. In KG-grounded reasoning, prior work has also suggested that KG paths can provide useful reward signals for compositional reasoning~\cite{YuvalPaper,JakePaper}. Motivated by these findings, we use lower-hop RL to optimize for both answer correctness and KG-path alignment, and then evaluate whether this improves generalization to harder 3-hop, 4-hop, and 5-hop question-answering.

The main contributions of this article are as follows:

\begin{itemize}
    \item We propose a context-augmented KG supervision framework that attaches supporting triples from the same source text chunk to each primary KG triple, producing CG-grounded supervision alongside standard KG-grounded supervision.

    \item We construct disease-specific Gastroparesis and Diabetes KG/CG datasets and generate structured multiple-choice QA items spanning 1-hop to 5-hop reasoning, grounded entirely in verifiable KG/CG paths.

    \item We introduce an LLM-judged, history-aware adaptive repair pipeline that improves the one-hop SFT foundation through continual targeted repair while removing or quarantining problematic noisy triples, ultimately reaching perfect accuracy on the cleaned retained validation set before employing RL.

    \item We compare RL initialized from unrepaired SFT checkpoints against RL initialized from repaired SFT checkpoints, directly testing whether a cleaner lower-hop factual foundation improves higher-hop generalization.

    \item Across both disease settings, we show that context-augmented SFT improves multi-hop performance, adaptive repair strengthens the one-hop foundation, and RL from repaired checkpoints yields the strongest 3-hop, 4-hop, and 5-hop results. We further evaluate robustness under answer-option shuffling to test whether the gains are sensitive to multiple-choice option order.
\end{itemize}

The rest of the article is organized as follows. In Section~\ref{sec:background_related_work}, we review the background and related work needed to understand the proposed framework. In Section~\ref{sec:methodology}, we present the methodology, including the overall pipeline, dataset construction, training procedures, adaptive repair, RL setup, and evaluation protocol. In Section~\ref{sec:results}, we report the experimental results. In Section~\ref{sec:discussion}, we discuss the main findings. In Section~\ref{sec:limitations}, we describe the limitations of the proposed approach and outline future work. Finally, Section~\ref{sec:conclusion} concludes the article.

\section{Background and Related Work}
\label{sec:background_related_work}

This section reviews the background and related work needed to motivate our proposed framework. Section~\ref{sec:bg_biomedical_kgs} introduces biomedical KGs and their role in question-answering. Section~\ref{sec:bg_context_augmented_supervision} discusses the limitation of isolated triples and motivates context-augmented KG supervision. Section~\ref{sec:bg_synthetic_qa} reviews how KGs can be used to generate QA supervision. Section~\ref{sec:bg_sft} discusses SFT for knowledge-grounded QA items. Section~\ref{sec:bg_targeted_repair} introduces targeted repair of missing biomedical knowledge after employing SFT. Finally, Section~\ref{sec:bg_rl} reviews RL for multi-hop reasoning. 

\subsection{Knowledge Graphs for Question-Answering}
\label{sec:bg_biomedical_kgs}

\begin{figure}[t]
    \centering
    \includegraphics[width=0.72\linewidth]{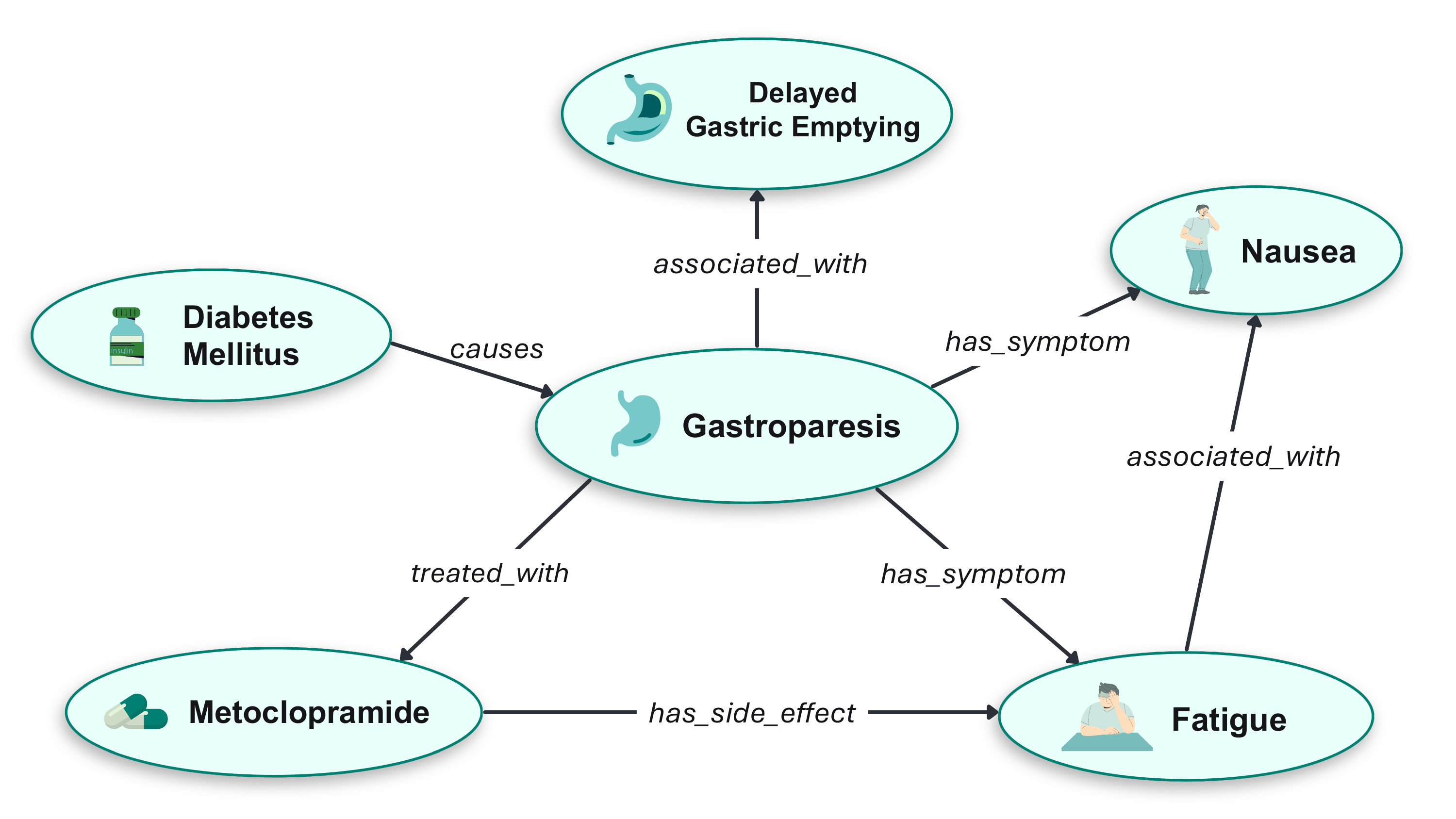}
    \caption{A subgraph of the gastroparesis KG. Nodes represent clinical concepts and directed edges represent relations.}
    \label{fig:toy_kg}
\end{figure}

KGs represent knowledge as structured relations among concepts. A KG can be represented as a set of triples:
\begin{equation}
\mathcal{G} = \{(h_i, r_i, t_i)\}_{i=1}^{N},
\label{eq:bg_kg}
\end{equation}
where $h_i$ denotes the head entity, $r_i$ denotes the relation, $t_i$ denotes the tail entity, and $N$ is the total number of triples in the graph. 
This representation converts information expressed across various documents into explicit facts that can be searched, connected, and computationally analyzed. A small subgraph of a gastroparesis KG is shown in Fig.~\ref{fig:toy_kg}. 
The value of explicit KG grounding is especially clear in high-stakes domains that require interpretability and traceability. For example, in biomedical question-answering, it is often not sufficient for a model to produce the correct answer; the answer should also be grounded in reliable evidence~\cite{WHOAIHealth,FUTUREAI,QA-GNN,hetionet}. Structured triples help meet this requirement by exposing the intermediate facts used during reasoning.

Prior work has used KGs to support question-answering by retrieving relevant entities, paths, or subgraphs and combining them with language-model representations. Graph-language reasoning methods connect questions and candidate answers to relevant KG nodes, build joint representations, and reason over both language context and KG structure~\cite{QA-GNN,JointLK,GreaseLM}. Monte Carlo Tree Search (MCTS) based approaches use KGs as external structured support for pretrained LLM agents by guiding graph-based exploration, searching over constrained KG paths, and using LLMs to evaluate or compose candidate reasoning paths before answer generation~\cite{MCTS,MCTSrekg,KGreasoningLLM}. Think-on-Graph further treats the LLM as an agent that explores KG entities and relations, retrieves promising reasoning paths, and uses those paths to support answer generation without additional model training~\cite{TOG}. These approaches show that structured knowledge can complement pretrained language models by providing explicit relational evidence during inference.


Beyond inference-time retrieval, KGs can also serve as structured sources for training and evaluation data. Recent work has used reliable KGs to generate reasoning subgraphs, explanations, KG-grounded curricula, thinking traces, QA supervision, and path-derived RL signals for language-model training~\cite{kgsft,DedhiaSI,JakePaper}. Collectively, these works show that KGs are useful not only as external evidence during inference, but also as data-generating foundations for constructing trustworthy reasoning supervision.

Together, these directions highlight the central importance of KG quality. KG construction typically requires entity extraction/normalization, relation extraction/normalization, and deduplication or consistency checking; errors at these stages can introduce noisy entities, incorrect relations, duplicate triples, or missing links that propagate into downstream QA systems~\cite{HealthcareKGSurvey,BiomedicalRelationExtractionKG}. Although earlier text-mining systems and neural models have made substantial progress on named entity recognition and relation extraction~\cite{LampleNER,BioBERT,MiwaBansalRE,semmeddb}, automatically constructing reliable domain-specific KGs from unstructured text remains challenging~\cite{AutoKGSurvey2024,OpenIESurvey2024}. In this work, we use GraphMERT as the KG extraction framework because it provides an automatic pipeline for distilling reliable domain-specific KGs from unstructured text into structured symbolic representations~\cite{margarita}. The resulting triples serve as both supervision and interpretable reasoning evidence for downstream QA. The detailed KG construction procedure and dataset statistics are provided in Section~\ref{sec:kg_construction}.

\subsection{From Isolated Triples to Context-Augmented KG Supervision}
\label{sec:bg_context_augmented_supervision}


Although KGs provide structured and interpretable knowledge, isolated triples may omit the surrounding evidence needed to interpret a relation. For example, gastroparesis is characterized by related symptoms and physiological findings, including nausea, vomiting, early satiety, postprandial fullness, bloating, abdominal pain, and delayed gastric emptying~\cite{GastroGuide}. Such adjacent facts
provide useful context for interpreting the target relation.

This motivates context-augmented KG representations. The idea of a CG was introduced as an extension of conventional triple-based KGs by associating entities and facts with additional contextual information~\cite{CGPaper}. In its general form, a CG can be written as
\begin{equation}
\mathcal{G}_{\mathrm{CG}} =
(\mathcal{E}, \mathcal{R}, \mathcal{Q}, \mathcal{E}_{C}, \mathcal{R}_{C}),
\label{eq:bg_context_graph_general}
\end{equation}
where $\mathcal{E}$ and $\mathcal{R}$ denote the sets of entities and relations, $\mathcal{E}_{C}$ and $\mathcal{R}_{C}$ denote the sets of entity contexts and relation contexts, and $\mathcal{Q}$ denotes the set of context-aware facts. An entity $e\in\mathcal{E}$ can be paired with an entity context $c^{e}\in\mathcal{E}_{C}$ to form a context-aware entity representation $\langle e,c^{e}\rangle$. A context-aware fact can then be written as $\langle h_{\tau}, r_{\tau}, t_{\tau}, c_{\tau}^{e} \rangle \in \mathcal{Q}$, where $h_{\tau},t_{\tau}\in\mathcal{E}$, $r_{\tau}\in\mathcal{R}$, and $c_{\tau}^{e}\in\mathcal{E}_{C}$ denote entity-level context associated with the entities participating in the fact $\tau$. Similarly, relation-level context can also be represented by pairing a relation $r_{\tau}$ with a relation context $c_{\tau}^{r}\in\mathcal{R}_{C}$, yielding a relation-context-aware fact $\langle h_{\tau}, r_{\tau}, t_{\tau}, c_{\tau}^{r}\rangle$.

This context-aware view inspires the main context-augmentation idea used in our work. Rather than treating each triple as a standalone fact, we represent a primary triple together with supporting triples that provide local evidence around the same statement. Specifically, for a primary triple $\tau_p$, we attach a set of supporting context triples,
\begin{equation}
\mathcal{C}(\tau_p)=\{\tau_{c,1},\tau_{c,2},\ldots,\tau_{c,m}\},
\label{eq:bg_context_set}
\end{equation}
where each $\tau_{c,j}$ is a triple that provides local evidence related to $\tau_p$. The resulting context-augmented triple can be written as
\begin{equation}
\widetilde{\tau}_p =
\left(\tau_p,\mathcal{C}(\tau_p)\right).
\label{eq:bg_context_augmented_triple}
\end{equation}
Thus, our CG formulation adapts the broader CG concept to the KG setting
by using structured supporting triples as the contextual evidence. The exact CG construction procedure, including candidate context retrieval, semantic similarity scoring, and thresholding, is described in Section~\ref{sec:context_graph_construction}.

\subsection{Question-Answer Item Generation from KGs}
\label{sec:bg_synthetic_qa}



A single triple $\tau=\langle h_{\tau}, r_{\tau}, t_{\tau}\rangle$ can support a one-hop question. A connected path $(\tau_1,\tau_2,\ldots,\tau_L)$ can support an $L$-hop question that requires composing multiple relations. Since manually generating such QA data is expensive, slow, and difficult to scale, KGs provide a structured basis for synthetic QA item generation. Dedhia et al.~use KG triples and paths as domain primitives to synthesize KG-grounded reasoning tasks and thinking traces~\cite{DedhiaSI}. Other KG-grounded approaches have also generated QA datasets, graph-query-based questions, curriculum-based QA items, and path-derived RL signals from structured KGs~\cite{biokgqa,primekg,JakePaper}.



However, synthetic QA generation can introduce quality risks, including invalid questions, ambiguous answer choices, weak or implausible distractors, answer leakage, and misalignment between the generated question and the source triple or path~\cite{PAQ,SynQA}. These issues are especially important in multiple-choice settings, where a model may exploit superficial option patterns instead of reasoning from the provided facts~\cite{OptionsShufflePaper}. Prior work on distractor generation emphasizes that distractor quality depends on plausibility, diversity, and incorrectness~\cite{distractor_quality,distractor_survey}, and synthetic instruction-generation work shows the importance of filtering invalid, low-quality, or overly similar examples before fine-tuning~\cite{selfinstruct}. Therefore, in this work, we use additional verification stages to check format quality, option quality, path-grounded reasoning, and answer correctness before using the generated QA items for SFT, adaptive repair, and RL, as described in Section~\ref{sec:qa_generation}.


\subsection{Supervised Fine-Tuning using Knowledge-Grounded QA Items}
\label{sec:bg_sft}

In knowledge-grounded question-answering, SFT adapts a pretrained language model to answer questions using structured evidence from KG-grounded or CG-grounded supervision. Let $\mathcal{D}_{\mathrm{SFT}}=\{(x_i,y_i)\}_{i=1}^{n}$ denote the SFT dataset, where $x_i$ contains a generated QA item together with its KG-grounded or CG-grounded evidence, and $y_i$ contains the expected response, including the correct answer and the associated reasoning trace when available. SFT updates the pretrained model parameters $\theta$ by minimizing the token-level negative log-likelihood:
\begin{equation}
\mathcal{L}_{\mathrm{SFT}}(\theta)
=
-\sum_{(x_i,y_i)\in \mathcal{D}_{\mathrm{SFT}}}
\sum_{j=1}^{|y_i|}
\log p_{\theta}
\left(
y_{i,j}\mid x_i, y_{i,<j}
\right).
\label{eq:bg_sft_loss}
\end{equation}
Thus, the model is trained to produce the target answer sequence conditioned on the question and its grounding evidence. Prior work has followed this direction by fine-tuning language models on KG-grounded reasoning tasks, thinking traces, and QA items synthesized from KG triples and paths~\cite{GraphGen,JakePaper,TravelKGReasoning,DedhiaSI,YuvalPaper}. This motivates the use of KG-derived QA items as a scalable way to adapt language models to domain-specific reasoning.



However, SFT alone does not guarantee that every target relation is reliably learned. Prior work suggests that fine-tuning can strongly affect output format, instruction following, and response style, but factual knowledge may still remain difficult to update reliably~\cite{lima,gekhman2024newknowledge,ghosal2024factualft}. As a result, high aggregate fine-tuning performance does not necessarily mean that all KG triples have been internalized. 
This motivates the adaptive repair stage introduced in Section~\ref{sec:LLM_Judge_SFT_Repair}, which identifies validation failures corresponding to missing knowledge and generates targeted additional supervision before RL.


\subsection{Targeted Repair of Missing Knowledge}
\label{sec:bg_targeted_repair}

Even after SFT, a language model may still fail on QA items generated from the same KG, indicating that some target facts or relations have not been reliably learned. This is problematic because standard SFT optimizes average loss over the training set, and prior work on group robustness shows that average-risk training can achieve strong overall performance even when certain subsets still have high error~\cite{jtt,GroupDRO}. Therefore, unresolved validation failures should be treated as targeted weaknesses rather than as random errors. This motivates our adaptive repair strategy, which is inspired by STAT~\cite{STAT}. In STAT, a stronger teacher model identifies remaining weaknesses in a student model and adapts the training data toward those weaknesses. A similar principle is applied to unresolved KG-specific failures instead of continuing to train uniformly on all data.

We refer to these unresolved validation failures as \textit{missing knowledge}. A missing-knowledge item is a primary triple $\tau_p$, or a reasoning pattern induced by that triple, for which the model produces an incorrect answer during validation. The goal of repair is to improve the model's coverage of the corresponding knowledge, rather than only improving its general response style. However, failed validation cases should be audited before they are reused for additional training. Prior work shows that generated instruction data should be filtered before fine-tuning and that noisy labels can negatively affect pretrained language models~\cite{selfinstruct,wang2023noiserobust}. Motivated by these risks, the LLM-judged repair step separates valid missing-knowledge items from unreliable cases. The detailed repair procedure, including failure categorization, targeted QA item generation, regression tracking, and quarantine review, is described in Section~\ref{sec:LLM_Judge_SFT_Repair}.

\subsection{Reinforcement Learning for Multi-Hop Reasoning}
\label{sec:bg_rl}



Prior work suggests that SFT can improve task imitation but may not generalize well beyond the training distribution, whereas RL with outcome-based rewards can encourage more generalizable behavior~\cite{SFTMemorizesRLGeneralizes}. More generally, SFT provides the initial domain grounding needed to learn structured facts and response format, and RL acts as a later compositional stage that helps the model combine learned facts into longer reasoning chains~\cite{SFTMemorizesRLGeneralizes,ReFT,YuvalPaper}.

RL post-training optimizes a policy $\pi_{\theta}$ using rewards assigned to generated responses. Given a question $q$ and generated response $y$, the objective can be written as
\begin{equation}
\max_{\theta}
\mathbb{E}_{q}
\mathbb{E}_{y\sim \pi_{\theta}(\cdot|q)}
\left[
R(y,q)
\right],
\label{eq:bg_rl_objective}
\end{equation}
where $R(y,q)$ is the reward function that scores the generated output. Recent language-model post-training methods commonly use policy-gradient optimization after SFT~\cite{InstructGPT,deeppath,SummaryHF}. In this work, we use Group Relative Policy Optimization (GRPO), a policy-optimization method introduced in DeepSeekMath that improves reasoning without requiring a separate value model~\cite{GRPO1,GRPO2}. 
Related work has also used RL for KG reasoning and structured reasoning paths, including relation traversal as a sequential decision problem, reward shaping for sparse or misleading rewards, and KG-path-derived reward signals for language-model reasoning~\cite{deeppath,kg_reward_shaping,YuvalPaper,JakePaper}. These works suggest that structured graph paths can provide scalable supervision for both final-answer correctness and intermediate reasoning alignment.


Furthermore, reward function design is a crucial component of RL-based multi-hop reasoning because it determines which reasoning behaviors are reinforced. Kansal et al.~show that combining final-answer correctness with KG path alignment provides a stronger signal for compositional reasoning than answer correctness alone~\cite{YuvalPaper}. An answer-only reward may not distinguish path-grounded reasoning from shortcut reasoning based on option patterns or spurious cues~\cite{YuvalPaper}. Zhu et al.~further show that penalizing incorrect generations more strongly can encourage the model to explore alternative correct trajectories~\cite{DanqiPaper}. Motivated by these findings, our RL reward evaluates both final-answer correctness and alignment with the KG path used to generate the QA item. The exact reward design is described in Section~\ref{sec:rl_setup}.



\section{Methodology}
\label{sec:methodology}

This section presents the proposed framework for context-augmented question-answering. We begin with an overview of the complete pipeline in Section~\ref{sec:overall_pipeline}. Section~\ref{sec:kg_construction} describes how the Gastroparesis and Diabetes KGs are constructed. Section~\ref{sec:context_graph_construction} introduces the CG construction procedure, where each primary KG triple is augmented with supporting triples from the same source text chunk. Section~\ref{sec:qa_generation} then explains how KG-grounded and CG-grounded QA items are generated from these structured knowledge sources.
We next describe the model training and repair stages. Section~\ref{sec:sft_setup} presents the SFT setup, including the KGModel and CGModel variants trained on full-coverage 1-hop QA items. Section~\ref{sec:LLM_Judge_SFT_Repair} describes the LLM-judged and history-aware adaptive triple repair pipeline, which identifies unresolved 1-hop knowledge gaps after SFT and generates targeted repair examples. Section~\ref{sec:rl_setup} explains the RL setup. Finally, Section~\ref{sec:evaluation_setup} describes the evaluation setup used to measure 3-hop, 4-hop, and 5-hop generalization, as well as robustness under option shuffling.

\subsection{Overall Pipeline}
\label{sec:overall_pipeline}

\begin{figure}[h]
    \centering
    \includegraphics[width=1\columnwidth]{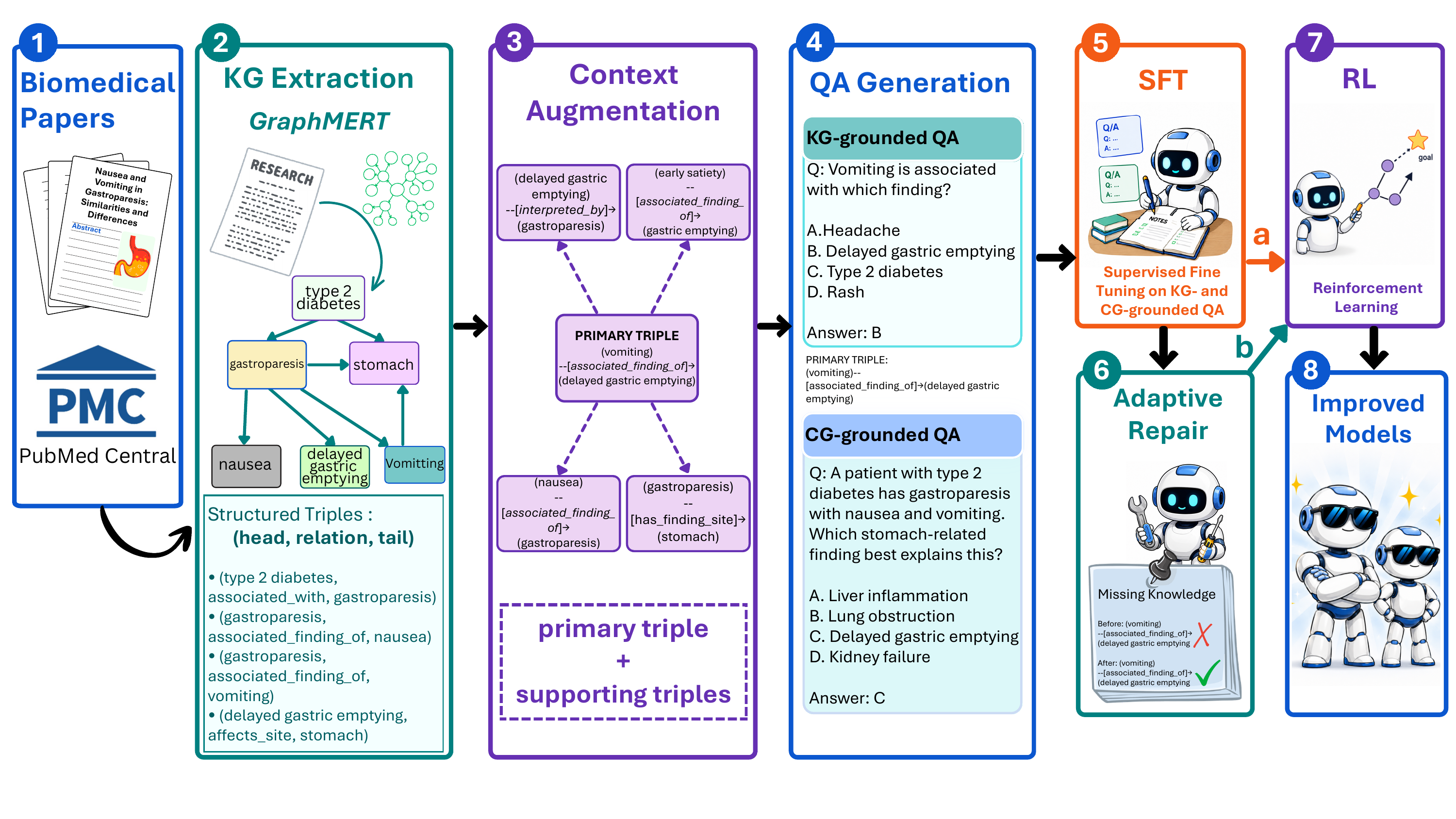}
    \caption{Overall pipeline of the proposed framework, as adapted to the biomedical arena. 
    Path (a) applies RL directly after SFT, whereas path (b) applies adaptive repair before RL.}
    \label{fig:Pipeline}
\end{figure}

The overall pipeline begins with a collection of articles. Although generally applicable, in Fig.~\ref{fig:Pipeline}, we show how the pipeline can be adapted to the biomedical arena, specifically for gastroparesis. From the corresponding articles, available from PubMed Central \cite{PubMedCentral}, we construct a disease-specific KG using GraphMERT. Each primary triple is then augmented with supporting triples from the same text chunks to form a CG. Using both the original KG triples and the context-augmented triples, we generate two types of QA items for SFT. After SFT, an adaptive repair stage is used to identify and improve missing knowledge before RL is applied to further improve multi-hop reasoning.

\subsection{Disease-Specific Knowledge Graph Construction}
\label{sec:kg_construction}
To validate our pipeline, we construct disease-specific biomedical KGs for gastroparesis and diabetes. Given biomedical text and a seed KG, GraphMERT learns both syntactic representations from text and semantic representations from KG triples, and then predicts candidate tails for head-relation pairs. A helper LLM is used only to combine GraphMERT-predicted tail tokens into coherent biomedical terms, after which candidate triples are filtered and deduplicated to obtain the final KG~\cite{margarita}. Additional implementation details of the GraphMERT pipeline for both gastroparesis and diabetes settings are provided in Appendix~\ref{app:graphmert_details}.

For gastroparesis, we run the full GraphMERT pipeline on gastroparesis-related biomedical abstracts collected from PubMed Central~\cite{PubMedCentral}. Approximately 26k open-access PubMed Central articles are retrieved to construct the disease-specific text corpus used for KG construction. The extracted outputs are assembled into an expanded gastroparesis KG containing 6,900 generated triples. After removing duplicate $(h,r,t)$ triples, the final gastroparesis KG contains 6,018 unique triples. This final deduplicated KG is used as the structured knowledge source for downstream context augmentation, QA item generation, SFT, adaptive repair, and RL.

For diabetes, we use the diabetes KG produced by the original GraphMERT pipeline. In GraphMERT, the similarity threshold $\beta$ controls how strongly candidate triples must be grounded in their originating text sequence. A higher $\beta$ produces fewer but more sequence-specific triples, whereas a lower $\beta$ allows broader semantically related triples \cite{margarita}. Since the original diabetes KG is much larger than the gastroparesis KG, we construct a size-matched diabetes subset to enable a fair comparison between the two disease settings. Specifically, we select 6,000 diabetes triples from the GraphMERT-extracted diabetes KG at $\beta=0.65$. The subset is not sampled uniformly at random; instead, it is selected using a relation-aware strategy that preserves relation diversity while maintaining graph connectivity. Each relation is assigned at least 20 triples, or all available triples if fewer than 20 exist. The remaining sampling budget is then allocated proportionately based on relation frequency. Within each relation, triples that reconnect to already selected nodes are preferred in the order: both endpoints already selected, one endpoint already selected, and neither endpoint selected. Relations with larger quotas are filled first so that frequent relations seed the node pool early. This procedure yields 6,000 sampled diabetes triples and 5,954 unique $(h,r,t)$ triples after deduplication. The statistics of the disease-specific KGs used in this work are summarized in Table~\ref{tab:kg_statistics}.


\begin{table}[h]
\centering
\caption{Statistics of the disease-specific KGs.}
\label{tab:kg_statistics}
\resizebox{\textwidth}{!}{%
\begin{tabular}{lccccc}
\toprule
\textbf{Disease} & \textbf{Source KG} & \textbf{Rows} & \textbf{Unique triples} & \textbf{Unique concepts} & \textbf{Relation types} \\
\midrule
Gastroparesis & Full GraphMERT pipeline run & 6,900 & 6,018 & 4,790 & 20 \\
Diabetes & GraphMERT-extracted KG subset & 6,000 & 5,954 & 4,397 & 28 \\
\bottomrule
\end{tabular}%
}
\end{table}


\subsection{Context Graph Construction}
\label{sec:context_graph_construction}


A useful property of the GraphMERT pipeline is that extracted triples are traceable to their originating source text sequence~\cite{margarita}. In our implementation, each processed source text chunk is assigned a unique identifier (\texttt{path\_idx}). This identifier is retained for every triple extracted from that chunk. We use this provenance information as a byproduct of the GraphMERT pipeline to retrieve candidate supporting triples. Specifically, for a primary triple $\tau_p$, all other triples extracted from the same source information chunk are treated as \textit{\textbf{candidate context triples}}. This enables us to retrieve local supporting evidence for each primary triple directly from the GraphMERT outputs, using the shared source-chunk identifier, without introducing an additional retrieval corpus.

Consider a KG triple:
$\tau = \langle h_{\tau}, r_{\tau}, t_{\tau} \rangle$.
For each primary triple and each candidate context triple extracted from the same source chunk, we convert the triple into a textual representation:
\begin{equation}
x(\tau) = h_{\tau} \; [\mathrm{SEP}] \; r_{\tau} \; [\mathrm{SEP}] \; t_{\tau}.
\end{equation}

We then encode each distinct triple string using the SapBERT model "cambridgeltl/SapBERT-from-PubMedBERT"~\cite{SapBERT}. Let $f(\cdot)$ denote the SapBERT encoder with mean pooling. The normalized embedding of a triple $\tau$ is computed as
\begin{equation}
\mathbf{z}_{\tau} =
\frac{f(x(\tau))}{\|f(x(\tau))\|_2}.
\end{equation}

Since the embeddings are $L_2$-normalized, the dot product between two triple embeddings is equivalent to cosine similarity:
\begin{equation}
s(\tau_p,\tau_c) =
\mathbf{z}_{\tau_p}^{\top}\mathbf{z}_{\tau_c},
\end{equation}
where $\tau_p$ is the primary triple and $\tau_c$ is a candidate context triple.

We first remove any candidate triple that exactly duplicates the primary triple after string normalization. The remaining candidates are scored using $s(\tau_p,\tau_c)$, sorted in descending order of similarity, and filtered using a similarity threshold $\theta$. The final context set for a primary triple $\tau_p$ is defined as
\begin{equation}
\mathcal{C}(\tau_p) =
\{\tau_c \in \mathcal{N}(\tau_p)
\mid \tau_c \neq \tau_p,\;
s(\tau_p,\tau_c) \geq \theta \},
\end{equation}
where $\mathcal{N}(\tau_p)$ denotes the set of candidate triples extracted from the same source information chunk as $\tau_p$. 
In this work, we set $\theta=0.65$, based on our preliminary experiments. Further details on the threshold selection and context-triple extraction procedure are provided in Appendix~\ref{app:context_graph_details}.

\begin{figure}[ht]
    \centering
    \includegraphics[width=1\columnwidth]{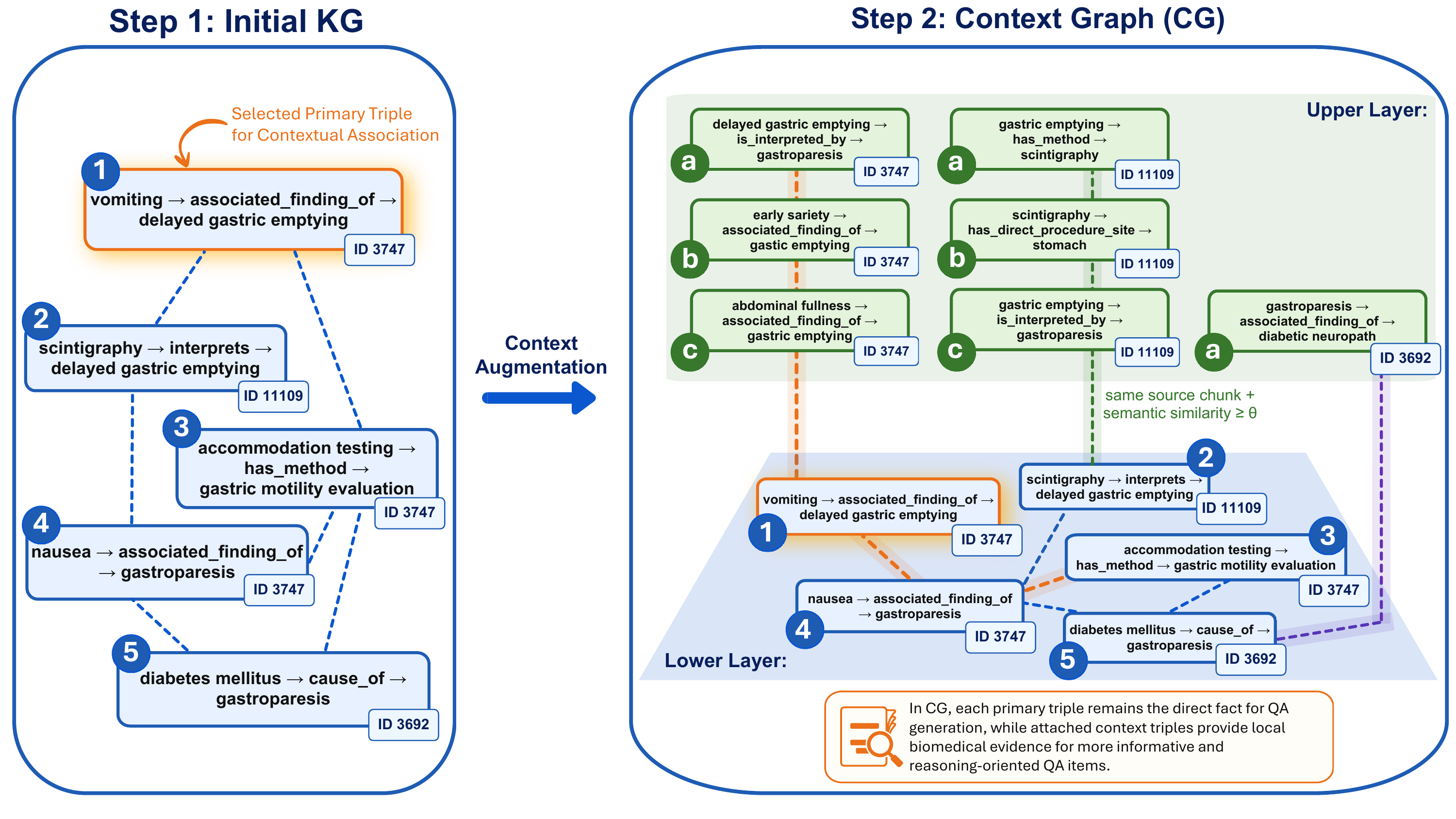}
    \caption{Illustration of the CG representation. The lower layer contains primary KG triples, while the upper layer contains supporting context triples extracted from the same source chunks and selected based on semantic similarity.}
    \label{fig:cg_subgraph}
\end{figure}

The resulting CG differs from the original KG in its structure. The original KG contains only the primary biomedical triples, where each triple is treated as an independent fact. In contrast, the CG can be viewed as a two-layer representation. The lower layer contains the original primary KG triples, while the upper layer contains supporting context triples. 
Thus, 
the attached context triples provide additional local evidence that can help generate more informative and reasoning-oriented QA items. A subgraph illustrating this two-layer CG representation is shown in Fig.~\ref{fig:cg_subgraph}.

\begin{tcolorbox}[
    colback=gray!4,
    colframe=black!60,
    title={Example of Primary and Context Triple Construction (Source ID: 3747)},
    fonttitle=\bfseries,
    arc=2mm,
    boxrule=0.6pt
]

\textbf{Source text chunk:}

\begin{tcolorbox}[
    colback=blue!4,
    colframe=blue!40,
    arc=1.5mm,
    boxrule=0.4pt
]
Gastroparesis is a condition characterized by \colorbox{yellow!25}{delayed gastric emptying} and the most common known underlying cause is \colorbox{yellow!25}{diabetes mellitus}. Symptoms include \colorbox{yellow!25}{nausea}, \colorbox{yellow!25}{vomiting}, \colorbox{yellow!25}{abdominal fullness}, and \colorbox{yellow!25}{early satiety}, which impact to varying degrees on the patient's quality of life. Symptoms and deficits do not necessarily relate to each other; hence, despite significant abnormalities in gastric emptying, some individuals have only minimal symptoms and, conversely, severe symptoms do not always relate to measures of gastric emptying.
\end{tcolorbox}

\vspace{1mm}
\textbf{Primary triple:}

\begin{tcolorbox}[
    colback=red!4,
    colframe=red!50!black,
    arc=1.5mm,
    boxrule=0.4pt
]
\[
\tau_p = \triple{vomiting}{associated\_finding\_of}{delayed gastric emptying}
\]
\end{tcolorbox}

\vspace{1mm}
\textbf{Context triples extracted from the same source chunk:}

\begin{tcolorbox}[
    colback=green!4,
    colframe=green!45!black,
    arc=1.5mm,
    boxrule=0.4pt
]
\[
\mathcal{C}(\tau_p) =
\left\{
\begin{array}{l}
\triple{delayed gastric emptying}{is\_interpreted\_by}{gastroparesis},\\[2pt]
\triple{diabetes mellitus }{cause\_of}{gastroparesis},\\[2pt]
\triple{early satiety}{associated\_finding\_of}{gastric emptying},\\[2pt]
\triple{abdominal fullness}{associated\_finding\_of}{gastric emptying},\\[2pt]
\triple{nausea}{associated\_finding\_of}{gastroparesis},\\[2pt]

\end{array}
\right\}
\]
\end{tcolorbox}

\end{tcolorbox}

The example in the box illustrates how the CG representation is constructed from a source text chunk. Details on how the CG is stored and accessed are provided in Appendix~\ref{app:cg_storage_access}.


\subsection{QA Item Generation}
\label{sec:qa_generation}

The synthetic QA item generation pipeline is inspired by the work of Dedhia et al.~\cite{DedhiaSI}. However, our setting differs in that we generate QA items under two grounding regimes, as indicated by the QA grounding regimes in the box.




\begin{tcolorbox}[
    colback=cyan!2,
    colframe=cyan!57!black,
    title={QA Grounding Regimes},
    fonttitle=\bfseries,
    arc=1.5mm,
    boxrule=0.5pt
]
\textbf{KG-grounded QA:} A QA item generated using only the target KG triple or multi-hop KG path.

\vspace{1mm}

\textbf{CG-grounded QA:} A QA item generated using the target primary triple or path together with supporting context triples extracted from the same source text chunk.
\end{tcolorbox}

We generate multiple-choice clinical QA items from two disease-specific biomedical KG settings: Gastroparesis and Diabetes. For each disease, QA items are generated under both KG-grounded and CG-grounded regimes. These two QA types enable us to compare whether context-augmented supervision improves model learning beyond supervision from isolated KG facts.

Let a sampled KG path be denoted as a sequence of triples:
$p = \left(\tau_1, \tau_2, \ldots, \tau_k \right)$.
For a 1-hop QA item, the path contains a single triple, i.e., $p=(\tau_1)$. For a multi-hop QA item, the path contains multiple connected triples. In the KG-grounded setting, the QA generation model receives only the path $p$. In the CG-grounded setting, the model additionally receives the supporting context triples
\begin{equation}
\mathcal{C}(p) = \bigcup_{\tau_i \in p} \mathcal{C}(\tau_i),
\end{equation}
where $\mathcal{C}(\tau_i)$ denotes the context triples attached to the primary triple $\tau_i$, as described in Section~\ref{sec:context_graph_construction}. The QA generation process can be summarized as
\begin{equation}
(q, a) =
\mathrm{LLM}_{\mathrm{MCQ}}
\left(
h_0,\; h_k,\; p,\; \mathcal{C}(p)
\right),
\label{eq:qa_generation}
\end{equation}
where $\mathrm{LLM}_{\mathrm{MCQ}}$ denotes the multiple-choice QA generation language model, $q$ is the generated clinical question, $a$ is the keyed correct answer, $p$ is the sampled KG path, and $h_0$ and $h_k$ denote the source and target concepts of the path. 

The QA generation prompt is designed to enforce factual grounding, clinical realism, and multiple-choice consistency. For each sampled triple or path, the prompt requires the LLM to generate a clinical vignette whose correct answer can be inferred from the provided KG or CG evidence. We enforce the following constraints during generation:

\begin{itemize}
    \item \textbf{Clinical vignette format:} The question must be written as a clinical vignette rather than a direct fact-recall question.

    \item \textbf{Path-grounded reasoning:} The question must require the model to reason over the provided triple or multi-hop path, linking the source concept $h_0$ to the target concept $h_k$. For higher-hop paths, the prompt explicitly instructs the LLM to construct the question so that the solution follows the full provided path and avoids shortcut reasoning. This prevents generated questions from being answerable through a simpler one-hop association, ensuring that the QA item tests the intended multi-hop reasoning structure.

    \item \textbf{Evidence-grounded answer:} The correct answer must be directly supported by the provided KG path. In the CG-grounded setting, the supporting context triples may provide additional background, but the primary triple or path remains the answer-bearing evidence.

    \item \textbf{Plausible distractors:} Each QA item must contain one correct answer and three plausible but incorrect answer choices.

    \item \textbf{Answer-option uniformization:} To reduce positional bias, the correct answer is approximately balanced across A, B, C, and D. This prevents the fine-tuned model from exploiting answer-position frequency instead of learning the underlying biomedical relation.
\end{itemize}

The complete QA generation pipeline consists of four stages.

\paragraph{Step 1: Multiple-choice question (MCQ) generation.}
Using the formalization presented in Eq.~\eqref{eq:qa_generation}, \texttt{Gemini-2.5-Flash} is used as $\mathrm{LLM}_{\mathrm{MCQ}}$ to generate the initial multiple-choice QA pair from the sampled source concept, target concept, KG path, and, when available, the supporting context triples. The model writes a clinical vignette with four answer options. 
After QA generation, we perform a format check and discard any pair that does not contain exactly four answer choices or does not specify a correct answer. 

\paragraph{Step 2: Quality filtering.}
A \texttt{Qwen3-1.7B-FP8} model is used as a quality filter. This stage verifies that the answer options are well-formed and that the distractors are not near-duplicates of each other or of the correct answer. QA pairs with low-quality distractors, repeated options, or overly similar choices are discarded to preserve the discriminative quality of the multiple-choice task. The remaining questions are then subjected to answer-option uniformization before thinking-trace generation. See the box for a CG-grounded QA pair.

\vspace{0.1cm}

\begin{tcolorbox}[
    breakable,
    colback=green!3,
    colframe=green!45!black,
    title={Example CG-grounded QA Pair},
    fonttitle=\bfseries,
    arc=2mm,
    boxrule=0.6pt,
    before skip=2mm,
    after skip=2mm
]

\textbf{Primary triple:}

\begin{tcolorbox}[
    colback=red!4,
    colframe=red!50!black,
    arc=1.5mm,
    boxrule=0.4pt
]
\[
\tau_p =
\triplehl{delayed gastric emptying}{associated\_finding\_of}{gastroparesis}
\]
\end{tcolorbox}

\vspace{1mm}
\textbf{Supporting context triples (Source ID: 12948):}

\begin{tcolorbox}[
    colback=green!4,
    colframe=green!45!black,
    arc=1.5mm,
    boxrule=0.4pt
]
\[
\mathcal{C}(\tau_p)=
\left\{
\begin{array}{l}
\triplehl{delayed gastric emptying}{has\_finding\_site}{stomach},\\[2pt]
\triplehl{gastroparesis}{has\_finding\_site}{stomach},\\[2pt]
\triplehl{symptomatic patients}{focus\_of}{gastroparesis},\\[2pt]
\triplehl{gastroparesis}{has\_associated\_morphology}{autonomic neuropathy}
\end{array}
\right\}
\]
\end{tcolorbox}

\vspace{1mm}
\textbf{Question:}

A 58-year-old male with a 20-year history of type 2 diabetes mellitus presents with chronic \entitytext{nausea}, \entitytext{early satiety}, and occasional postprandial \entitytext{vomiting of undigested food particles}, which has worsened over the past six months. He reports significant weight loss. Endoscopy reveals \reltext{no mechanical obstruction} or mucosal abnormalities in the \entitytext{stomach} or duodenum. His hemoglobin A1c is 9.2\%. Based on the patient's clinical presentation and initial workup, which of the following findings would be most critical in \reltext{establishing the definitive diagnosis}?

\vspace{1mm}
\textbf{Options:}
\begin{enumerate}[label=\Alph*.]
    \item \entitytext{Markedly reduced gastric emptying} \reltext{demonstrated by scintigraphy}
    \item Evidence of \entitytext{vagal nerve damage} on nerve conduction studies
    \item Presence of chronic intractable \entitytext{nausea and vomiting}
    \item Normal findings on upper gastrointestinal endoscopy
\end{enumerate}

\textbf{Answer:} A

\end{tcolorbox}

\paragraph{Step 3: Thinking-trace generation.}
\texttt{Gemini-2.5-Pro} is used to generate a detailed thinking trace. Given the clinical vignette, answer options, the source KG or CG path, the model produces a reasoning explanation that connects the question to the answer through the provided triples. This step can be written as
\begin{equation}
e =
\mathrm{LLM}_{\mathrm{trace}}
\left(
q,\; \mathcal{O},\; p,\; \mathcal{C}(p)
\right),
\label{eq:trace_generation}
\end{equation}
where $\mathrm{LLM}_{\mathrm{trace}}$ denotes the language model used for explanation generation, $q$ is the clinical vignette, $\mathcal{O}$ denotes the answer options, $p$ is the source KG path, and $e$ is the generated explanation. In KG-grounded QA generation, $\mathcal{C}(p)$ is omitted and the explanation is conditioned only on the KG path. In CG-grounded QA generation, the explanation is conditioned on both the primary path and the supporting context triples. By anchoring the explanation to the KG or CG evidence, the resulting QA item provides structured relational supervision for fine-tuning.

The trace-generation prompt instructs the model to provide a detailed explanation, include the steps leading to the answer, and explain the relationships between the concepts using the provided KG or CG evidence. The prompt also instructs the model not to explicitly mention that it is using a provided path or context. By generating explanations from the question, options, and grounding evidence alone, the thinking trace serves as a model-derived reasoning path that is later verified by the correctness filter before being used for fine-tuning.

\paragraph{Step 4: Triple-grounded correctness filtering.}
Independent grading LLMs are used for correctness filtering to verify that each generated QA item is sound and consistent with the source KG or CG evidence. Although the MCQ and thinking-trace generation stages are grounded in explicit triples or paths, errors may still occur due to ambiguous phrasing, inconclusive evidence along the path, weak distractors, or hallucinated reasoning. Therefore, we perform a final correctness check over the complete QA item, including the question, answer options, thinking trace, keyed answer, and source KG or CG path.

For this stage, the complete QA item is organized under a template prompt and given to an LLM grader. The grader is instructed to evaluate whether: (a) the keyed correct answer follows from the clinical vignette and the provided KG or CG path, and (b) every claim in the thinking trace is supported by the provided evidence without hallucination. The grader outputs a binary verdict indicating whether the QA item should be retained or discarded. To reduce the effect of idiosyncratic failures from any single grader, we use two independent grading LLMs, \texttt{Qwen3-32B} and \texttt{Qwen2.5-72B}, and retain a QA item only when both graders verify its correctness. This two-factor agreement improves robustness through cross-model consistency.

\begin{figure}[h]
    \centering
    \includegraphics[width=1\columnwidth]{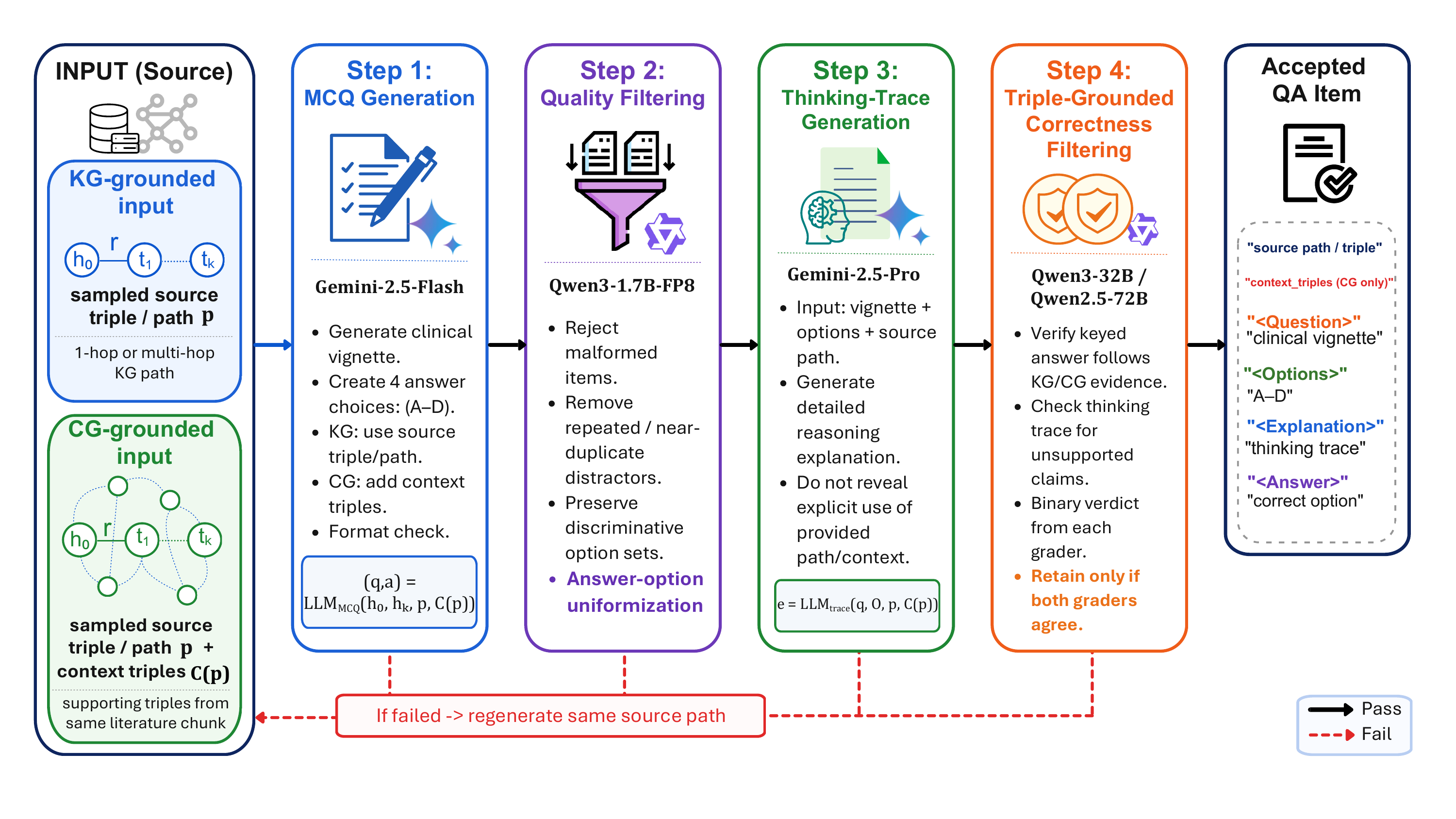}
    \caption{QA item generation workflow for KG-grounded and CG-grounded supervision. Each sampled source path is converted into a multiple-choice clinical QA item, filtered for option quality, augmented with a thinking trace, and verified using triple-grounded correctness filtering. Failed items can be regenerated using the same source path, while accepted items are used for downstream training and validation.}
    \label{fig:qa_pipeline}
\end{figure}

\vspace{0.1cm}
If a generated item fails at any stage, the same source triple or path is sent back for regeneration rather than being removed from the dataset. This is especially important for 1-hop paths, because they correspond to the basic disease-specific facts and define the core factual coverage of the graph. The complete workflow of the QA item generation process is shown in Fig.~\ref{fig:qa_pipeline}. Additional details of each stage in the QA item generation pipeline, including the prompts used for MCQ generation, quality filtering, thinking-trace generation, and correctness verification, are provided in Appendix~\ref{app:QA_items_generation}.

\vspace{0.1cm}
For each disease and grounding regime, we create two separate full-coverage 1-hop QA datasets: one for SFT training and one for validation. This process results in 6,018 KG-grounded and 6,018 CG-grounded 1-hop QA items for Gastroparesis, and 5,954 KG-grounded and 5,954 CG-grounded 1-hop QA items for Diabetes.  The training set is used to expose the model to all basic facts in the KG, while the validation set is held out and used during the adaptive triple repair stage. Specifically, after SFT, the validation 1-hop set is used to identify which triples the model still fails to answer correctly, enabling us to detect remaining knowledge gaps and target them for repair.



For RL training, our goal was to generate around 5,000 QA items per setting using a mixture of 1-hop and 2-hop paths. In preliminary experiments, using only 2-hop QA items for RL degraded performance because the model began to forget 1-hop factual knowledge. Therefore, we used a mixed 1-hop/2-hop QA dataset for RL training. The curation of the 1-hop and 2-hop paths, together with the final number of QA items used for RL training, is described in Section~\ref{sec:rl_setup}. To evaluate whether lower-hop training generalizes to harder reasoning, we generated 8,000 QA items each for 3-hop, 4-hop, and 5-hop evaluation.



Each QA item is stored as a JSON record. For both KG-grounded and CG-grounded datasets, the record contains the source path or triple, represented as \texttt{paths = [\{start, relation, end\}]}. For 1-hop items, this field contains one triple, while for multi-hop items it contains a connected multi-edge path. Each record also includes the \texttt{source\_concept}, \texttt{target\_concept}, \texttt{k\_hops}, and \texttt{path\_idx}. 
For CG-grounded QA items, the record additionally includes \texttt{context\_triples}, a list of supporting triples of the form \texttt{[head, relation, tail]}. The final training text is stored in a \texttt{question\_and\_explanation} field with tagged sections: \texttt{<Question>} for the clinical vignette, \texttt{<Options>} for the four answer choices, \texttt{<Explanation>} for the thinking trace, and \texttt{<Answer>} for the keyed correct option.


\subsection{The Supervised Fine-Tuning Setup}
\label{sec:sft_setup}

Before applying RL, we first fine-tune the base model to learn the 1-hop knowledge contained in each KG. Our hypothesis is that RL is more effective when initialized from an SFT model that has learned all 1-hop knowledge triples it is capable of reliably learning. 
Since multi-hop reasoning is built from combinations of lower-hop facts, errors in basic 1-hop knowledge can propagate during RL and negatively affect higher-hop reasoning. 
We expect a model with higher 1-hop accuracy before RL to produce fewer downstream errors after RL and to generalize better to 3-hop, 4-hop, and 5-hop QA.

We perform SFT using two types of QA supervision, as indicated by the SFT model variants in the box. 

\begin{tcolorbox}[
    colback=cyan!2,
    colframe=cyan!57!black,
    title={SFT Model Variants},
    fonttitle=\bfseries,
    arc=1.5mm,
    boxrule=0.5pt
]
\textbf{KGModel:} A Qwen3-14B model fine-tuned using KG-grounded QA items generated only from the target KG triple or multi-hop KG path.

\vspace{1mm}

\textbf{CGModel:} A Qwen3-14B model fine-tuned using CG-grounded QA items generated from the target primary triple or path together with supporting context triples from the same source text chunk.
\end{tcolorbox}

For validation of the approach, we train separate KGModel and CGModel variants for each disease setting, resulting in four SFT models: Gastroparesis-KGModel, Gastroparesis-CGModel, Diabetes-KGModel, and Diabetes-CGModel. For a fair comparison, all four models are initialized from the same base model and trained using the same fine-tuning configuration, LoRA setup, optimizer, sequence length, batch size, number of epochs, and learning-rate schedule. The only difference between KGModel and CGModel is the type of QA supervision used during training.

We fine-tune the model using parameter-efficient LoRA adapters under the same configuration for all KGModel and CGModel variants. Table~\ref{tab:sft_hyperparameters} summarizes the hyperparameters used for all SFT experiments.

\begin{table}[h]
\centering
\caption{Hyperparameters used for SFT.}
\label{tab:sft_hyperparameters}
\small
\begin{tabular}{p{0.38\columnwidth}p{0.52\columnwidth}}
\toprule
\textbf{Hyperparameter} & \textbf{Value} \\
\midrule
Base model & \texttt{Qwen3-14B} \\
Fine-tuning method & LoRA \\
LoRA rank / $\alpha$ / dropout & 16 / 16 / 0.05 \\
Target modules & \texttt{q\_proj, k\_proj, v\_proj, o\_proj, gate\_proj, up\_proj, down\_proj} \\
Training epochs & 8 \\
Per-device micro-batch size & 1 \\
Gradient accumulation steps & 32 \\
Effective batch size & 32 \\
Learning rate & $2\times10^{-4}$ \\
Learning-rate schedule & Cosine decay \\
Warmup ratio & 0.1 \\
Weight decay & 0.01 \\
Optimizer & AdamW, \texttt{adamw\_torch\_fused} \\
AdamW parameters & $\beta_1=0.9$, $\beta_2=0.99$, $\epsilon=1\times10^{-8}$ \\
Maximum gradient norm & 1.0 \\
Maximum sequence length & 4096 tokens \\
Precision & \texttt{bf16} \\
Hardware & 1$\times$ GPU; NVIDIA A100 80GB\\
\bottomrule
\end{tabular}
\end{table}


\subsection{LLM-Judged and History-Aware Adaptive Triple Repair}
\label{sec:LLM_Judge_SFT_Repair}

As discussed in Section~\ref{sec:bg_targeted_repair}, 
He et al.~\cite{STAT} proposed the STAT strategy, which uses a stronger teacher model to identify missing skills and synthesize targeted training
examples. Building on this concept, we design an adaptive repair pipeline for triple-level KG and CG supervision.

After the initial SFT stage, the model is evaluated on the full 1-hop validation set. Each validation example is associated with a source triple $\tau=(h,r,t)$ and a generated QA item. If the model answers a validation question incorrectly, we treat the corresponding triple-question pair as a candidate failure case. Importantly, an incorrect prediction does not necessarily imply that the model lacks the corresponding knowledge. Even though we use a QA generation and filtering pipeline, some validation failures may still arise from residual noise in the data-generation process. In particular, a failed validation case can have several possible causes:
\begin{itemize}
    \item \textbf{Noisy or unsupported triple:} the source triple itself may be incorrect, weakly supported, or too ambiguous to serve as reliable supervision.
    \item \textbf{Misaligned validation question:} the generated question, answer choices, or expected answer may not faithfully test the intended triple, even if the triple itself is valid.  The model may partially understand the relevant fact but still choose the wrong option due to confusing distractors in the question.
    \item \textbf{True missing knowledge:} the triple and validation question may both be valid, but the SFT model may still fail to answer correctly because the corresponding knowledge or reasoning pattern was not learned.
\end{itemize}

\begin{figure}[h]
    \centering
    \includegraphics[width=1\columnwidth]{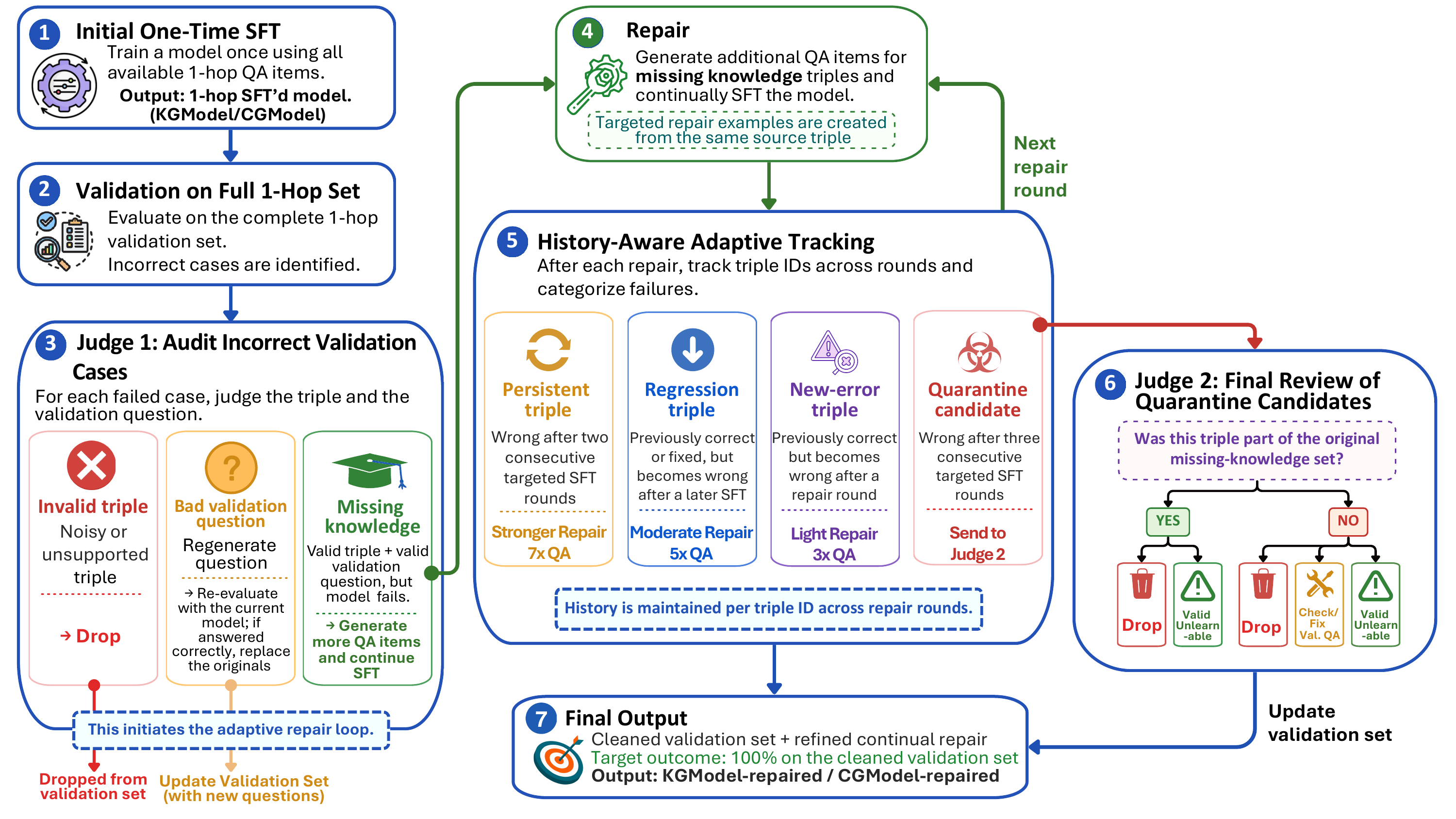}
    \caption{Overview of the LLM-judged and history-aware adaptive triple repair pipeline. After initial SFT, incorrect validation cases are audited by Judge 1 to separate invalid triples, bad validation questions, and valid missing-knowledge cases. Missing-knowledge cases are repaired through targeted QA generation and continual SFT, while triple-level error history is tracked across repair rounds to identify fixed, new-error, regression, persistent, and quarantine-candidate triples. Judge 2 is applied only to repeatedly failing quarantine candidates before they are removed or labeled as valid but unlearnable.}
    \label{fig:repair_pipeline}
\end{figure}

Therefore, instead of blindly re-training on new QA items generated from incorrectly learned triples, we introduce an LLM-judged and history-aware adaptive repair pipeline that first diagnoses the source of each failure and then applies targeted repair only when the failure corresponds to valid missing knowledge. The overall pipeline is summarized in Fig.~\ref{fig:repair_pipeline}.

We define a \emph{missing knowledge item} as a triple, or a reasoning pattern induced by a triple, for which both the triple and the corresponding validation question are judged to be valid, yet the SFT model fails to produce the correct answer. For each missing knowledge item, we generate additional targeted QA examples from the same source triple. These examples provide multiple alternative phrasings and reasoning traces for the same underlying knowledge. The model is then continually fine-tuned on the targeted examples so that the repair step directly addresses the validated failures observed during post-SFT evaluation.

The repair process is adaptive because the set of training examples for each round is determined by the error history of individual triple IDs (\texttt{path\_idx}). After every repair round, we re-evaluate the model on the validation set and update the status of each triple. Table~\ref{tab:repair_buckets} summarizes the main triple categories used in our pipeline to bin each triple.

\begin{table}[h]
\centering
\small
\caption{History-aware error categories used during adaptive triple repair. The repair weight increases as a triple repeatedly fails targeted SFT, while newly introduced errors receive lighter repair to avoid overfitting and further forgetting.}
\begin{tabular}{p{0.20\linewidth} p{0.50\linewidth} p{0.22\linewidth}}
\hline
\textbf{Category} & \textbf{Definition} & \textbf{Repair action} \\
\hline
Fixed triple &
A previously incorrect triple-question pair that becomes correct after a repair round. &
Track as fixed; no targeted repair in the next round. \\
\hline
Missing-skill triple &
A Judge-1-confirmed valid triple and valid validation question that the model answers incorrectly after initial SFT. &
Generate targeted QA examples and include in the first repair round. \\
\hline
New-error triple &
A triple that was previously answered correctly but becomes incorrect after a repair round, with no prior repair or fix history. &
Light repair, e.g., $3\times$ additional QA examples. \\
\hline
Regression triple &
A triple that was previously fixed or previously correct, but becomes wrong again after a later repair round. This also includes new-error triples that remain wrong after being targeted once. &
Moderate repair, e.g., $5\times$ additional QA examples. \\
\hline
Persistent triple &
A triple that remains wrong after two consecutive targeted SFT repair rounds. &
Stronger repair, e.g., $7\times$ additional QA examples. \\
\hline
Quarantine candidate triple &
A triple that remains wrong after three targeted SFT repair attempts. &
Send to Judge 2 before removal or final labeling. \\
\hline
Quarantined triple &
A triple removed from the cleaned validation set after repeated failure and Judge-2 review. &
Exclude from the retained validation set. \\
\hline
\end{tabular}

\label{tab:repair_buckets}
\end{table}

The repair strength is controlled by the number of newly generated QA items assigned to each error category. Since different errors reflect different levels of difficulty, we do not use the same number of repair examples for all failed triples. New-error triples are usually recent repair-induced failures and are, therefore, given a light repair. Regression triples indicate forgetting of previously correct or previously fixed knowledge and receive a moderate repair. Persistent triples have failed across repeated targeted SFT rounds and, therefore, receive the strongest repair. Based on preliminary experiments, we use a $7/5/3$ weighting scheme: persistent triples receive $7$ additional QA examples, regression triples receive $5$, and new-error triples receive $3$. We also experimented with more aggressive and lighter schedules, such as $10/7/5$ and $5/3/1$, but these alternatives either did not improve validation accuracy consistently or required additional repair rounds to reach the same performance. 

\paragraph{Judge 1: initial failure audit.}
The first LLM judge is employed immediately after the initial SFT validation step. Given an incorrect validation case, Judge 1 determines whether the failure is due to an invalid triple ({\footnotesize\texttt{INVALID\_TRIPLE}}), a bad validation question ({\footnotesize\texttt{BAD\_VAL}}), or a true missing knowledge ({\footnotesize\texttt{MISSING\_KNOWLEDGE}}). We used \texttt{Qwen3.6-27B} as Judge 1, a recent state-of-the-art Qwen-family model with strong reasoning and long-context capabilities, making it suitable for the initial failure diagnosis~\cite{qwen36_27b}.

Invalid triples are removed from the validation set, because repairing the model on unsupported or noisy facts would introduce incorrect supervision. For unlearned triples, validation questions are regenerated and then re-evaluated using the current model. If the regenerated question is aligned with the source triple and the model answers it correctly, it replaces the original bad validation question in the validation set. Only cases judged as valid triples with valid questions are treated as missing knowledge and passed to the first targeted repair round. The prompts used for Judge 1 are provided in Appendix~\ref{app:adaptive_repair_details}.

This first judging stage is used to reduce manual inspection and prevent unnecessary repair iterations. Without Judge 1, the repair procedure may repeatedly train on noisy triples or flawed questions, which can create misleading improvements or cause additional regressions. Judge 1, therefore, separates data-quality issues from genuine model-side missing knowledge before continual SFT begins.

\paragraph{History-aware adaptive repair.}
After Judge 1, the repair loop proceeds automatically using the history of triple-level outcomes. In the first repair round, the model is trained on additional QA examples generated from Judge-1-confirmed missing-knowledge triples. After re-evaluation, fixed triples are recorded, still-wrong missing knowledge is carried forward, and newly introduced errors are assigned to the new-error bin. In later rounds, triples are promoted between bins based on their failure history. For example, a new-error triple that remains wrong after targeted repair becomes a regression triple, while a regression triple that remains wrong after another targeted repair becomes persistent. Persistent triples receive stronger repair. If a triple continues to fail after three targeted SFT attempts, it is not immediately removed; instead, it is passed to Judge 2 for final review.

\paragraph{Judge 2: Quarantine review.}
The second LLM judge is directed at only quarantine candidates. These are triples that have repeatedly resisted targeted repair despite multiple rounds of additional QA generation and continual SFT. Judge 2 is not targeted at every new error because new errors are usually evidence of model-side forgetting rather than data corruption. 
Instead, new errors are first repaired through the history-aware mechanism and Judge 2 is reserved for cases that persist after repeated repair.

Judge 2 has a narrower and more history-aware role than Judge 1. While Judge 1 performs the initial broad failure audit over all incorrect validation cases, Judge 2 only reviews a small set of repeatedly failing quarantine candidates after their repair history is known. We used \texttt{Qwen3-32B}~\cite{qwen3} as Judge 2 to provide an independent second review within the same Qwen model family, rather than relying on the same judge for both the initial diagnosis and the final quarantine decision. This reduces the risk that model-specific judgment errors from Judge 1 are repeated during the final review stage. Since Judge 2 operates on a much smaller and more constrained set of cases, its role is not to repeat the full initial audit, but to decide whether a repeatedly failing case should be dropped, regenerated, or retained as valid but currently unlearnable.

If the triple was already judged as valid missing knowledge by Judge 1, Judge 2 decides whether it should be dropped as invalid/noisy under closer inspection or retained as valid but currently unlearnable (Case A). If the triple was not part of the original Judge-1 missing-knowledge set, Judge 2 may additionally check whether the validation question itself should be regenerated (Case B). In practice, triples labeled as invalid or unsupported are dropped, while valid-but-unlearnable triples are reported separately as cases that remain valid but are not reliably learned by the current model and repair strategy.

\paragraph{Quarantining noisy or unlearnable triples.}
In this work, \emph{quarantining} means removing a small number of problematic triples from the retained validation set after repeated targeted repair and Judge-2 review. Quarantining is not used to hide ordinary model errors. It is used only after a triple has repeatedly failed repair and has been reviewed by the judge. This step is justified because KGs extracted from text can contain ambiguous, underspecified, or weakly supported triples, and it may occasionally be difficult to align generated QA items perfectly with the intended fact. Retaining such cases as ordinary validation examples can prevent a meaningful estimate of whether the model has learned the validated portion of the graph.

The final reported perfect accuracy is, therefore, measured on the \emph{retained cleaned validation set}: invalid triples, bad validation questions, and Judge-2-quarantined cases are excluded or corrected, while valid repairable missing knowledge remains in the evaluation. \textit{This distinction is important:} the goal of the pipeline is not simply to force the model to memorize every noisy example, but to separate valid missing knowledge from data-quality artifacts, repair the valid failures through targeted continual SFT, and explicitly document any triples that are removed or labeled as valid but unlearnable.

Algorithm~\ref{alg:adaptive_repair} provides a concise procedural summary of the repair pipeline. Additional implementation details, including the step-by-step repair procedure and the prompts used for Judge 1 and Judge 2, are provided in Appendix~\ref{app:adaptive_repair_details}.

\refstepcounter{algorithm}
\label{alg:adaptive_repair}

\begin{tcolorbox}[
    breakable,
    enhanced,
    colback=white,
    colframe=black,
    boxrule=0.6pt,
    arc=0mm,
    left=1mm,
    right=1mm,
    top=1mm,
    bottom=1mm,
    title={Algorithm~\thealgorithm: LLM-Judged and History-Aware Adaptive Triple Repair},
    fonttitle=\bfseries,
    coltitle=black,
    colbacktitle=white,
    attach boxed title to top left={xshift=2mm,yshift=-2mm},
    boxed title style={
        colback=white,
        colframe=white,
        boxrule=0pt
    }
]

\small
\begin{algorithmic}[1]
\Require Initial SFT model $M$, one-hop validation set $\mathcal{V}$, QA generator $G$, Judge 1, Judge 2
\Ensure Repaired model $M^\star$ and retained cleaned validation set $\mathcal{V}_{clean}$

\State Evaluate $M$ on $\mathcal{V}$ and collect incorrect triple-question pairs $\mathcal{E}$
\State Initialize $\mathcal{V}_{clean} \leftarrow \mathcal{V}$
\State Initialize repair history $\mathcal{H}$ for each triple ID

\For{each failed example $e \in \mathcal{E}$}
    \State $y \leftarrow \textsc{Judge1}(e)$
    \If{$y = \texttt{INVALID\_TRIPLE}$}
        \State Remove the corresponding triple from $\mathcal{V}_{clean}$
    \ElsIf{$y = \texttt{BAD\_VAL}$}
        \State Regenerate the validation question using $G$
        \State Replace the original validation question if the regenerated item is valid
    \ElsIf{$y = \texttt{MISSING\_KNOWLEDGE}$}
        \State Add the triple ID to the missing-knowledge bin
    \EndIf
\EndFor

\For{each repair round}
    \State Generate targeted QA examples for current repair bins
    \State Apply repair weights: persistent $7\times$, regression $5\times$, new-error $3\times$
    \State Continually fine-tune $M$ on the targeted QA examples
    \State Evaluate updated $M$ on $\mathcal{V}_{clean}$

    \For{each triple ID in $\mathcal{V}_{clean}$}
        \State Update status as fixed, new-error, regression, persistent, or quarantine candidate
    \EndFor

    \For{each quarantine candidate}
        \State $z \leftarrow \textsc{Judge2}(\text{candidate})$
        \If{$z = \texttt{INVALID\_TRIPLE}$}
            \State Remove the triple from $\mathcal{V}_{clean}$
        \ElsIf{$z = \texttt{BAD\_VAL}$}
            \State Regenerate and replace the validation question
        \ElsIf{$z = \texttt{VALID\_UNLEARNABLE}$}
            \State Remove from $\mathcal{V}_{clean}$ and report separately
        \EndIf
    \EndFor

    \State Update repair history $\mathcal{H}$

    \If{no incorrect examples remain in $\mathcal{V}_{clean}$}
        \State \textbf{break}
    \EndIf
\EndFor

\State \Return $M^\star \leftarrow M$ and $\mathcal{V}_{clean}$
\end{algorithmic}
\normalsize
\end{tcolorbox}


\subsection{Reinforcement Learning Setup}
\label{sec:rl_setup}

After SFT and adaptive repair, we use RL to improve the model's multi-hop reasoning ability. Following the intuition that 1-hop SFT helps the model acquire basic factual associations, while RL can encourage generalization beyond memorized facts~\cite{SFTMemorizesRLGeneralizes}, our hypothesis is that a model with fewer unresolved 1-hop errors before RL will learn more stable multi-hop reasoning. 

Our training pipeline, therefore, consists of two stages. First, the base \texttt{Qwen3-14B} model is fine-tuned using full-coverage 1-hop QA items, producing KGModel and CGModel variants for both Gastroparesis and Diabetes. These models are then further improved through adaptive repair until the remaining 1-hop validation failures are removed or quarantined. Second, the resulting SFT LoRA adapters are merged into the base model, and full-model GRPO is applied using a mixed RL training corpus containing both 1-hop and 2-hop QA items.

We evaluate two RL initialization settings for each disease and grounding regime (see box below). In the first setting, RL starts directly from the initial SFT models, denoted as KGModel and CGModel. In the second setting, RL starts from the repaired SFT models, denoted as KGModel-Repaired and CGModel-Repaired, where adaptive repair has improved 1-hop validation accuracy to 100\%, excluding quarantined triples when applicable. The resulting RL models are denoted as KGModel-RL and CGModel-RL when initialized from the initial SFT models, and KGModel-Repaired-RL and CGModel-Repaired-RL when initialized from the repaired SFT models. This comparison enables us to test whether RL benefits from a repaired lower-hop initialization.

\begin{tcolorbox}[
    colback=cyan!2,
    colframe=cyan!57!black,
    title={RL Model Variants},
    fonttitle=\bfseries,
    arc=1.5mm,
    boxrule=0.5pt
]
\textbf{KGModel-RL/CGModel-RL:} RL models initialized from KGModel or CGModel after the original 1-hop SFT stage, before adaptive repair reaches full 1-hop validation accuracy.

\vspace{1mm}

\textbf{KGModel-Repaired-RL/CGModel-Repaired-RL:} RL models initialized from KGModel-Repaired or CGModel-Repaired after adaptive repair, where the model has learned all reliably learnable 1-hop validation items.
\end{tcolorbox}

\paragraph{RL training corpus construction.}
The RL training corpus is constructed from a mixture of 1-hop and 2-hop QA items. Although each KG contains many possible 2-hop paths, we do not use all available paths for RL training. RL training is computationally expensive. Prior work by Kansal et al.~\cite{YuvalPaper} shows that using more RL examples does not necessarily lead to better performance in KG reasoning settings. Therefore, rather than maximizing the number of RL examples, we construct a carefully curated path manifest before generating the RL QA items. This enables the RL corpus to emphasize informative 1-hop and 2-hop paths while keeping the training size practical.
The target manifest contains 6,000 paths: 1,500 1-hop paths and 4,500 2-hop paths. For the 1-hop portion, we prioritize triples that are useful for RL by considering three criteria: repaired or hard triples, high-connectivity triples, and relation-proportional sampling. Repaired or hard triples are included because they correspond to knowledge that was difficult for the model to learn during SFT. High-connectivity triples are included because they involve concepts that participate in many graph paths and are, therefore, likely to support multi-hop reasoning. Relation-proportional sampling is used to preserve the relation distribution of the KG.

For the 2-hop portion, we first ensure broad coverage of all reachable unique concept nodes. The remaining 2-hop paths are then selected to improve rare relation-pair diversity. Specifically, relation-pair sampling is biased toward rare relation pairs using an inverse-frequency weighting proportional to $1/\mathrm{freq}(r_1,r_2)$, where $(r_1,r_2)$ denotes the relation pair along the 2-hop path. This prevents the RL corpus from being dominated by frequent relation patterns and encourages the model to learn more diverse multi-hop reasoning structures.

Unlike the full-coverage 1-hop SFT dataset, the RL dataset does not require every selected path to survive QA generation. In the SFT stage, each 1-hop triple is repeatedly sent through the QA generation pipeline until a valid QA item is obtained, because the goal is complete 1-hop coverage. In the RL stage, however, there are many possible 2-hop paths, and complete coverage is not required. Therefore, we generate QA items from the selected path manifest and retain only those that pass the quality filtering and correctness verification stages. Although the target manifest contains 6,000 paths, filtering produces slightly fewer final QA items, with approximately 5,500 retained examples per setting. The final RL training counts are shown in Table~\ref{tab:rl_training_counts}.

\begin{table}[h]
\centering
\caption{Final RL training corpus sizes after QA generation, quality filtering, and correctness verification.}
\label{tab:rl_training_counts}
\begin{tabular}{lccc}
\toprule
\textbf{Setting} & \textbf{Final} & \textbf{1-hop} & \textbf{2-hop} \\
\midrule
Diabetes KG & 5,706 & 1,458 & 4,248 \\
Diabetes CG & 5,370 & 1,430 & 3,940 \\
Gastroparesis KG & 5,705 & 1,435 & 4,270 \\
Gastroparesis CG & 5,544 & 1,423 & 4,121 \\
\bottomrule
\end{tabular}
\end{table}

\paragraph{GRPO training objective.}
We optimize the policy $\pi_{\theta}$ using GRPO~\cite{GRPO1,GRPO2}. For each QA prompt $q$, the model generates a group of $N$ outputs,
\begin{equation}
\mathcal{O}(q)=\{O_1,O_2,\ldots,O_N\}.
\end{equation}
Each output $O_i$ consists of an internal reasoning trace enclosed in \texttt{<think>} tags, followed by a final multiple-choice answer enclosed in \texttt{<answer>} tags. In our experiments, we generate $N=2$ completions per prompt.

\paragraph{Reward design.}
Building on prior work on reward design for KG-based reasoning, we use a reward function composed of
answer correctness and gated path alignment. Kansal et al.~\cite{YuvalPaper} studied multiple reward formulations for KG
reasoning and found that binary correctness and path-alignment rewards are particularly effective. Stephen et al.~\cite{JakePaper} further showed that gating path-alignment reward by answer correctness improves training stability. Motivated by this finding, our path-alignment reward is activated only for correct final answers. This gating follows the motivation that a model should not receive process-level
credit for mentioning path-relevant biomedical concepts if it ultimately selects the wrong answer.

Let $q$ denote a QA prompt generated from a KG path,
$p = (\tau_1,\tau_2,\ldots,\tau_L)$,
where each triple is represented as $\tau_i=\langle h_i,r_i,t_i\rangle$ and $L$ is the hop length. For each prompt $q$, the policy generates an output $O_i$ consisting of a reasoning trace $r_i$ and a final predicted answer $\hat{a}_i$. The ground-truth answer is denoted by $a^{*}$. The total reward for output $O_i$ is defined as
\begin{equation}
R_{\mathrm{total}}(O_i)
=
R_{\mathrm{bin}}(\hat{a}_i,a^{*})
+
R_{\mathrm{path}}(r_i,p,\hat{a}_i,a^{*}).
\label{eq:total_reward}
\end{equation}

\paragraph{Binary correctness reward.}
The binary correctness reward provides the primary outcome-level supervision signal. It assigns a small positive reward for the correct final answer and a larger negative reward for an incorrect answer:
\begin{equation}
R_{\mathrm{bin}}(\hat{a}_i,a^{*})
=
\begin{cases}
\alpha, & \text{if } \hat{a}_i=a^{*},\\
-\beta, & \text{otherwise},
\end{cases}
\label{eq:binary_reward}
\end{equation}
where $\alpha,\beta>0$ and $\beta>\alpha$. In our experiments, we use $\alpha=0.1$ and $\beta=1.0$. This asymmetric design strongly penalizes incorrect answers while still giving a positive signal for correct generations.

\paragraph{Gated path-alignment reward.}
The path-alignment reward provides KG-grounded process supervision by measuring whether the model's reasoning trace aligns with the concepts in the ground-truth path. Let $T(r_i)$ denote the set of normalized tokens extracted from the reasoning trace $r_i$ and let $T(p)$ denote the set of normalized entity tokens appearing on path $p$. For a path composed of triples $\langle h_i,r_i,t_i\rangle$, $T(p)$ is derived from the path entities $\{h_i,t_i\}_{i=1}^{L}$.

The path coverage score is defined as
\begin{equation}
\mathrm{coverage}(r_i,p)
=
\frac{|T(r_i)\cap T(p)|}{|T(p)|}.
\label{eq:path_coverage}
\end{equation}

Path alignment is rewarded only when the final answer is correct:
\begin{equation}
R_{\mathrm{path}}(r_i,p,\hat{a}_i,a^{*})=0
\quad
\text{if } \hat{a}_i \neq a^{*}.
\label{eq:path_reward_gate}
\end{equation}
This prevents the model from receiving partial credit for mentioning path-relevant concepts while arriving at an incorrect conclusion.

When $\hat{a}_i=a^{*}$, the path-alignment reward is computed as
\begin{equation}
R_{\mathrm{path}}(r_i,p,\hat{a}_i,a^{*})
=
\min
\left(
\left[
\gamma_1 \cdot \mathrm{coverage}(r_i,p)
+
\gamma_2 \cdot
\mathbb{I}\left(|T(r_i)\cap T(p)| \geq 2\right)
\right]
\cdot \phi_{\mathrm{rep}}(r_i),
\;
R_{\max}
\right),
\label{eq:path_alignment_reward}
\end{equation}
where $\mathbb{I}(\cdot)$ is an indicator function that rewards traces mentioning at least two distinct path entities and $\phi_{\mathrm{rep}}(r_i)\in[0,1]$ is a repetition-penalty factor that down-weights degenerate traces with excessive token repetition. In our experiments, we set $\gamma_1=1.2$, $\gamma_2=0.3$, and $R_{\max}=1.5$, following \cite{YuvalPaper}.

This reward design encourages the model to produce the correct final answer while also aligning its reasoning trace with the underlying KG path. Since the path-alignment component is grounded in the true KG structure and gated by answer correctness, the reward promotes compositional reasoning without giving credit for superficially mentioning relevant concepts in incorrect responses.

\paragraph{GRPO implementation details.}
For each RL run, the corresponding SFT LoRA adapter is first merged into the base \texttt{Qwen3-14B} model. We then perform full-model GRPO using the Huggingface TRL implementation with DeepSpeed ZeRO-3. All RL runs use the same hyperparameter configuration to ensure a fair comparison across diseases, grounding regimes, and initialization settings. The shared GRPO configuration is summarized in Table~\ref{tab:grpo_hyperparameters}.

\begin{table}[h]
\centering
\caption{GRPO hyperparameters used for RL.}
\label{tab:grpo_hyperparameters}
\small
\renewcommand{\arraystretch}{0.9}
\begin{tabular}{p{0.42\columnwidth}p{0.48\columnwidth}}
\toprule
\textbf{Hyperparameter} & \textbf{Value} \\
\midrule
Algorithm & GRPO, TRL \\
Base model & \texttt{Qwen3-14B} \\
Initialization & Merged SFT LoRA adapter \\
Update type & Full-model GRPO \\
Parallelism & DeepSpeed ZeRO-3 \\
Hardware & 4$\times$ NVIDIA H200, 141GB each \\
Training epochs & 2 \\
Generations per prompt & 2 \\
Temperature & 0.6 \\
Top-$p$ & 0.9 \\
Repetition penalty & 1.15 \\
Maximum prompt length & 512 tokens \\
Maximum completion length & 2048 tokens \\
Per-device batch size & 1 \\
Gradient accumulation steps & 16 \\
Effective batch size & 64 \\
Learning rate & $6\times10^{-6}$ \\
KL coefficient $\beta$ & 0.08 \\
Optimizer & AdamW, \texttt{adamw\_torch} \\
Precision & \texttt{bf16} \\
Reward components & Binary correctness $(+0.1/-1.0)$ + gated path alignment \\

\bottomrule
\end{tabular}
\renewcommand{\arraystretch}{1.0}
\end{table}

Algorithm~\ref{alg:sft_rl_pipeline} summarizes the complete SFT--RL pipeline, including RL initialized from both the initial SFT models and the repaired SFT models.

\refstepcounter{algorithm}
\label{alg:sft_rl_pipeline}

\begin{tcolorbox}[
    breakable,
    enhanced,
    colback=white,
    colframe=black,
    boxrule=0.6pt,
    arc=0mm,
    left=1mm,
    right=1mm,
    top=1mm,
    bottom=1mm,
    title={Algorithm~\thealgorithm: Complete SFT-RL Training Pipeline},
    fonttitle=\bfseries,
    coltitle=black,
    colbacktitle=white,
    attach boxed title to top left={xshift=2mm,yshift=-2mm},
    boxed title style={
        colback=white,
        colframe=white,
        boxrule=0pt
    }
]

\small
\begin{algorithmic}[1]
\Require Base model $M_0$, 1-hop SFT set $\mathcal{D}_{1}$, 1-hop validation set $\mathcal{V}_{1}$, RL set $\mathcal{D}_{RL}=\mathcal{D}_{1h}^{RL}\cup\mathcal{D}_{2h}^{RL}$
\Ensure RL models initialized from both initial and repaired SFT models

\State Fine-tune $M_0$ on the full-coverage 1-hop QA set $\mathcal{D}_{1}$
\State Obtain KGModel or CGModel, depending on the grounding regime

\State Evaluate KGModel/CGModel on $\mathcal{V}_{1}$
\State Apply adaptive repair to unresolved valid 1-hop validation failures
\State Obtain KGModel-Repaired or CGModel-Repaired

\Statex

\State \textbf{RL from initial SFT model}
\State Merge the LoRA adapter of KGModel/CGModel into the base model
\State Train with GRPO on $\mathcal{D}_{RL}$ using binary correctness and gated path-alignment rewards
\State Obtain KGModel-RL or CGModel-RL

\Statex

\State \textbf{RL from repaired SFT model}
\State Merge the LoRA adapter of KGModel-Repaired/CGModel-Repaired into the base model
\State Train with GRPO on $\mathcal{D}_{RL}$ using binary correctness and gated path-alignment rewards
\State Obtain KGModel-Repaired-RL or CGModel-Repaired-RL

\State \Return KGModel-RL/CGModel-RL and KGModel-Repaired-RL/CGModel-Repaired-RL
\end{algorithmic}
\normalsize
\end{tcolorbox}

        

\subsection{Evaluation Setup}
\label{sec:evaluation_setup}

We evaluate all models using multiple-choice biomedical QA. The main evaluation metric is accuracy, defined as the percentage of questions for which the model selects the correct answer option:
\begin{equation}
\mathrm{Accuracy}
=
\frac{1}{|\mathcal{D}_{test}|}
\sum_{i=1}^{|\mathcal{D}_{test}|}
\mathbb{I}(\hat{a}_i = a_i^{*}),
\end{equation}
where $\hat{a}_i$ is the model-predicted answer choice and $a_i^{*}$ is the ground-truth answer choice for test item $i$.

Each test item consists of a clinical vignette and four answer choices labeled A-D. During inference, the model receives an MCQ and is required to output one final answer choice from A, B, C, or D. The model completion consists of a reasoning trace followed by a final answer. The exact evaluation prompt used for inference is provided in Appendix~\ref{app:Evaluation}.

The primary goal of evaluation is to measure higher-hop generalization. Although SFT is performed using full-coverage 1-hop QA and RL is performed using a mixture of 1-hop and 2-hop QA, testing is conducted on 3-hop, 4-hop, and 5-hop QA items. This enables us to evaluate whether lower-hop training leads to compositional reasoning that transfers to longer biomedical reasoning chains.

We evaluate models on both KG-grounded and CG-grounded higher-hop QA sets for Gastroparesis and Diabetes. For each disease, we compare nine model variants, summarized in Table~\ref{tab:evaluation_models}.  


\begin{table}[h]
\centering
\caption{Model variants evaluated in the higher-hop QA experiments.}
\label{tab:evaluation_models}
\small
\renewcommand{\arraystretch}{1.2}
\begin{tabular}{ll}
\toprule
\textbf{Model name} & \textbf{Description} \\
\midrule
Base & Original \texttt{Qwen3-14B} model \\
KGModel & SFT model trained using KG-grounded 1-hop QA \\
CGModel & SFT model trained using CG-grounded 1-hop QA \\
KGModel-Repaired & KGModel after adaptive repair \\
CGModel-Repaired & CGModel after adaptive repair \\
KGModel-RL & RL initialized from KGModel \\
CGModel-RL & RL initialized from CGModel \\
KGModel-Repaired-RL & RL initialized from KGModel-Repaired \\
CGModel-Repaired-RL & RL initialized from CGModel-Repaired \\
\bottomrule
\end{tabular}
\renewcommand{\arraystretch}{1.1}
\end{table}

This evaluation design allows us to isolate the contribution of each stage in the proposed pipeline:

\begin{itemize}
    \item \textbf{Effect of context-augmentation:} Comparing KGModel with CGModel measures whether CG-grounded QA supervision improves higher-hop reasoning compared with KG-grounded QA supervision during SFT.

    \item \textbf{Effect of adaptive repair:} Comparing KGModel with KGModel-Repaired, and CGModel with CGModel-Repaired, measures whether repairing unresolved 1-hop knowledge gaps improves downstream performance.

    \item \textbf{Effect of repaired RL initialization:} Comparing KGModel-RL and CGModel-RL with KGModel-Repaired-RL and CGModel-Repaired-RL tests whether RL benefits more from a repaired lower-hop initialization.

    \item \textbf{Higher-hop generalization:} Evaluating all models on 3-hop, 4-hop, and 5-hop QA measures whether improvements from SFT, repair, and RL transfer to increasingly longer reasoning chains.
\end{itemize}

\paragraph{Robustness Analysis.}
In addition, we evaluate robustness to answer-option order. LLMs may sometimes exploit superficial patterns in MCQs, such as the ordering or position of answer options, instead of relying on the underlying clinical evidence and reasoning path\cite{OptionsShufflePaper}. To assess whether our models are robust to this type of positional bias, we use Stress-Test 3 (option shuffling) following Gu et al.~\cite{OptionsShufflePaper}. For each 3-hop, 4-hop, and 5-hop multiple-choice evaluation set, we rebuild the questions by randomly permuting all four answer-option texts while keeping the question stem unchanged. The correct answer text is preserved, but its option label is updated according to its new position after shuffling. We then re-evaluate every model variant on the shuffled evaluation sets using the same inference and answer-parsing protocol as in the original evaluation.

For each model, we report both the original accuracy and the option-shuffled accuracy. We also compute the accuracy change:
\begin{equation}
\Delta_{\mathrm{shuffle}}
=
\mathrm{Acc}_{\mathrm{shuffle}}
-
\mathrm{Acc}_{\mathrm{original}},
\end{equation}
where $\mathrm{Acc}_{\mathrm{original}}$ is the accuracy on the original test set and $\mathrm{Acc}_{\mathrm{shuffle}}$ is the accuracy after distractor shuffling. A small value of $|\Delta_{\mathrm{shuffle}}|$ indicates that the model is relatively robust to distractor ordering, while a large negative value suggests sensitivity to answer-option position or distractor arrangement.

Overall, this evaluation setup measures both higher-hop QA accuracy and prediction stability under controlled perturbations of the answer choices.


\section{Experimental Results}
\label{sec:results}

This section evaluates the proposed context-augmented QA framework across both Gastroparesis and Diabetes. We organize the results around the following questions:

\begin{itemize}
    \item \textbf{Context augmentation:} Does CG-grounded supervision improve higher-hop QA compared with KG-grounded supervision?

    \item \textbf{Adaptive repair:} Can the repair stage identify noisy or unreliable validation failures, quarantine them when necessary, and improve the model on the remaining valid 1-hop knowledge items?

    \item \textbf{Repair before RL:} Does RL benefit from starting from a repaired SFT model rather than an unrepaired SFT model?

    \item \textbf{Higher-hop generalization:} Do improvements from SFT, adaptive repair, and RL transfer to 3-hop, 4-hop, and 5-hop QA?

    \item \textbf{Robustness:} Are the resulting models stable under answer-option shuffling?
\end{itemize}

We report results for both diseases under KG-grounded and CG-grounded evaluation settings, enabling us to analyze context augmentation, adaptive repair, RL, and robustness within a unified experimental framework.

\subsection{Context-Augmented SFT Improves Multi-Hop QA}
\label{sec:results_context_aug}

We first evaluate whether context-augmented QA supervision improves higher-hop generalization after SFT. For this comparison, we use the initial SFT models. KGModel is trained using KG-grounded 1-hop QA, while CGModel is trained using CG-grounded 1-hop QA. Both models are evaluated on 3-hop, 4-hop, and 5-hop QA under two evaluation settings: KG-grounded QA and CG-grounded QA.

Table~\ref{tab:context_aug_sft_results} shows the results for Gastroparesis and Diabetes. Across both diseases and both evaluation settings, CGModel consistently outperforms KGModel. Importantly, the improvement is observed not only on CG-grounded QA, where the test format includes context triples, but also on KG-grounded QA, where only the KG path is provided. This suggests that context-augmented SFT improves the model's internalization of biomedical relations and its ability to compose them across longer reasoning chains, rather than merely improving performance through train-test format matching.

\begin{table*}[h]
\centering
\caption{Higher-hop QA accuracy of Base, KGModel, and CGModel before adaptive repair and RL. Results are shown for both diseases under KG-grounded and CG-grounded evaluation settings. Bold values indicate the highest accuracy for each disease, evaluation setting, and hop length.}
\label{tab:context_aug_sft_results}
\small
\renewcommand{\arraystretch}{1.2}
\begin{tabular}{llccccccccc}
\toprule
\textbf{Disease} & \textbf{Eval. set} 
& \multicolumn{3}{c}{\textbf{Base}} 
& \multicolumn{3}{c}{\textbf{KGModel}} 
& \multicolumn{3}{c}{\textbf{CGModel}} \\
\cmidrule(lr){3-5}
\cmidrule(lr){6-8}
\cmidrule(lr){9-11}
 & & \textbf{3-hop} & \textbf{4-hop} & \textbf{5-hop}
   & \textbf{3-hop} & \textbf{4-hop} & \textbf{5-hop}
   & \textbf{3-hop} & \textbf{4-hop} & \textbf{5-hop} \\
\midrule
Gastroparesis & KG-grounded & 91.01 & 89.10 & 87.04 & 95.30 & 91.38 & 89.81 & \textbf{97.28} & \textbf{93.58} & \textbf{91.96} \\
Gastroparesis & CG-grounded & 90.00 & 87.00 & 83.50 & 93.63 & 90.34 & 86.39 & \textbf{96.59} & \textbf{93.65} & \textbf{90.18} \\
Diabetes & KG-grounded & 88.98 & 86.10 & 81.21 & 95.55 & 93.04 & 89.83 & \textbf{97.14} & \textbf{95.70} & \textbf{91.85} \\
Diabetes & CG-grounded & 89.76 & 86.79 & 81.86 & 94.35 & 92.35 & 87.30 & \textbf{95.85} & \textbf{93.92} & \textbf{89.96} \\
\bottomrule
\end{tabular}
\renewcommand{\arraystretch}{1.0}
\end{table*}

The strongest trend is that CGModel improves over KGModel in every disease, evaluation setting, and hop length. For Gastroparesis, CGModel improves over KGModel by 1.98, 2.20, and 2.15 percentage points on KG-grounded 3-hop, 4-hop, and 5-hop QA, respectively. On CG-grounded QA, the gains are even larger: 2.96, 3.31, and 3.79 percentage points. A similar pattern is observed for Diabetes, where CGModel improves over KGModel by 1.59, 2.66, and 2.02 percentage points on KG-grounded QA, and by 1.50, 1.57, and 2.66 percentage points on CG-grounded QA. These results indicate that training with supporting context triples improves higher-hop QA performance beyond the original 1-hop supervision.

\subsection{LLM-Judged Adaptive Repair Produces 100\% Accuracy on Cleaned One-Hop Validation Sets}
\label{sec:results_repair}

After the initial SFT stage, we apply the LLM-judged and history-aware adaptive triple repair pipeline to the 1-hop validation sets. The purpose of this stage is not simply to force the model to memorize every failed validation item. Instead, the repair pipeline separates validation failures into three categories: invalid or noisy triples, misaligned validation questions, and valid missing-knowledge items. Invalid triples are removed, bad validation questions are regenerated, and valid missing-knowledge items are used to generate targeted repair QA examples for continual fine-tuning.

Table~\ref{tab:repair_summary_main} summarizes the repair outcomes for both diseases and grounding regimes. Across all four settings, adaptive repair produces 100\% accuracy on the cleaned retained validation set. The number of repair rounds ranges from 9 to 10. The final cleaned validation sets retain between 5861 and 5943 examples, depending on the number of triples removed or quarantined during Judge-1 and Judge-2 review. A full round-by-round repair trace for CGModel-Gastro, including fixed, persistent, regression, new-error, quarantine, and best-epoch counts, is provided in Appendix~\ref{app:repair_cg_gastro}.

\begin{table*}[h]
\centering
\caption{Summary of LLM-judged adaptive repair on one-hop validation sets. Final accuracy is reported on the cleaned retained validation set after invalid triples, bad validation questions, and valid-unlearnable cases are handled.}
\label{tab:repair_summary_main}
\small
\renewcommand{\arraystretch}{1.2}
\resizebox{\textwidth}{!}{%
\begin{tabular}{lcccccc}
\toprule
\textbf{Setting} 
& \textbf{Initial SFT Acc.} 
& \textbf{Initial Wrong} 
& \textbf{Judge-1 Invalid} 
& \textbf{Judge-1 Missing} 
& \textbf{Repair Rounds} 
& \textbf{Final Clean Acc.} \\
\midrule
KGModel-Gastro & 97.26\% $(5853/6018)$ & 165 & 10 & 92  & 10 & 100.00\% $(5943/5943)$ \\
CGModel-Gastro & 95.96\% $(5775/6018)$ & 243 & 16 & 140 & 9  & 100.00\% $(5922/5922)$ \\
KGModel-Diab   & 97.65\% $(5814/5954)$ & 140 & 34 & 84  & 9  & 100.00\% $(5893/5893)$ \\
CGModel-Diab   & 96.91\% $(5770/5954)$ & 184 & 36 & 101 & 9  & 100.00\% $(5861/5861)$ \\
\bottomrule
\end{tabular}%
}
\renewcommand{\arraystretch}{1.1}
\end{table*}

The initial SFT models already achieve high 1-hop validation accuracy, ranging from 95.96\% to 97.65\%. However, each model still leaves a non-trivial number of validation failures, ranging from 140 to 243 incorrect cases. Judge-1 review shows that these failures are not all genuine missing knowledge. For example, some are invalid triples and others are bad validation questions. This supports the need for an LLM-judged repair stage rather than blindly retraining on every incorrect validation item.

After Judge-1 filtering and multiple targeted repair rounds, the remaining difficult cases are reviewed by Judge 2. Cases judged to be invalid or valid but currently unlearnable are removed from the retained validation denominator, while bad validation questions are regenerated and kept when the underlying triple is valid. Therefore, the final 100\% accuracy should be interpreted as 100\% accuracy on the cleaned retained validation set, not as evidence that every originally generated validation item was valid or learnable.

\begin{table*}[h]
\centering
\caption{Judge-2 quarantine review summary. Invalid triples and valid-unlearnable cases are removed from the retained validation denominator, while bad validation questions are regenerated and kept.}
\label{tab:judge2_summary_main}
\small
\renewcommand{\arraystretch}{0.95}
\begin{tabular}{lcccc}
\toprule
\textbf{Setting} 
& \textbf{Judge-2 Invalid} 
& \textbf{Valid-Unlearnable} 
& \textbf{Bad Val. Regenerated} 
& \textbf{Judge-2 Removed} \\
\midrule
KGModel-Gastro & 35 & 30 & 3 & 65 \\
CGModel-Gastro & 33 & 47 & 2 & 80 \\
KGModel-Diab   & 7  & 20 & 0 & 27 \\
CGModel-Diab   & 25 & 32 & 2 & 57 \\
\bottomrule
\end{tabular}
\renewcommand{\arraystretch}{1.0}
\end{table*}

Table~\ref{tab:judge2_summary_main} further summarizes the Judge-2 quarantine review. The number of removed Judge-2 cases ranges from 27 for KGModel-Diab to 80 for CGModel-Gastro. These removals reflect triples that were judged to be invalid/noisy or valid but currently unlearnable under the repair setup. This cleaning step prevents unreliable validation items from being repeatedly used as repair targets and produces a cleaner one-hop validation set for selecting repaired SFT checkpoints.
These repaired checkpoints, denoted as KGModel-Repaired and CGModel-Repaired, are then used as stronger lower-hop initializations for the RL experiments in the next section.

\subsection{Repair Before RL Improves Higher-Hop Generalization}
\label{sec:results_repair_rl}
We next evaluate whether RL benefits from starting from a repaired SFT checkpoint. For each disease and grounding regime, we compare two RL initialization settings. In the first setting, RL is initialized from the original SFT model, yielding KGModel-RL or CGModel-RL. In the second setting, RL is initialized from the repaired SFT model, yielding KGModel-Repaired-RL or CGModel-Repaired-RL. This comparison directly tests whether resolving valid 1-hop knowledge gaps before RL leads to stronger generalization on 3-hop, 4-hop, and 5-hop QA.

\begin{figure}[H]
    \centering
    \includegraphics[width=1\columnwidth]{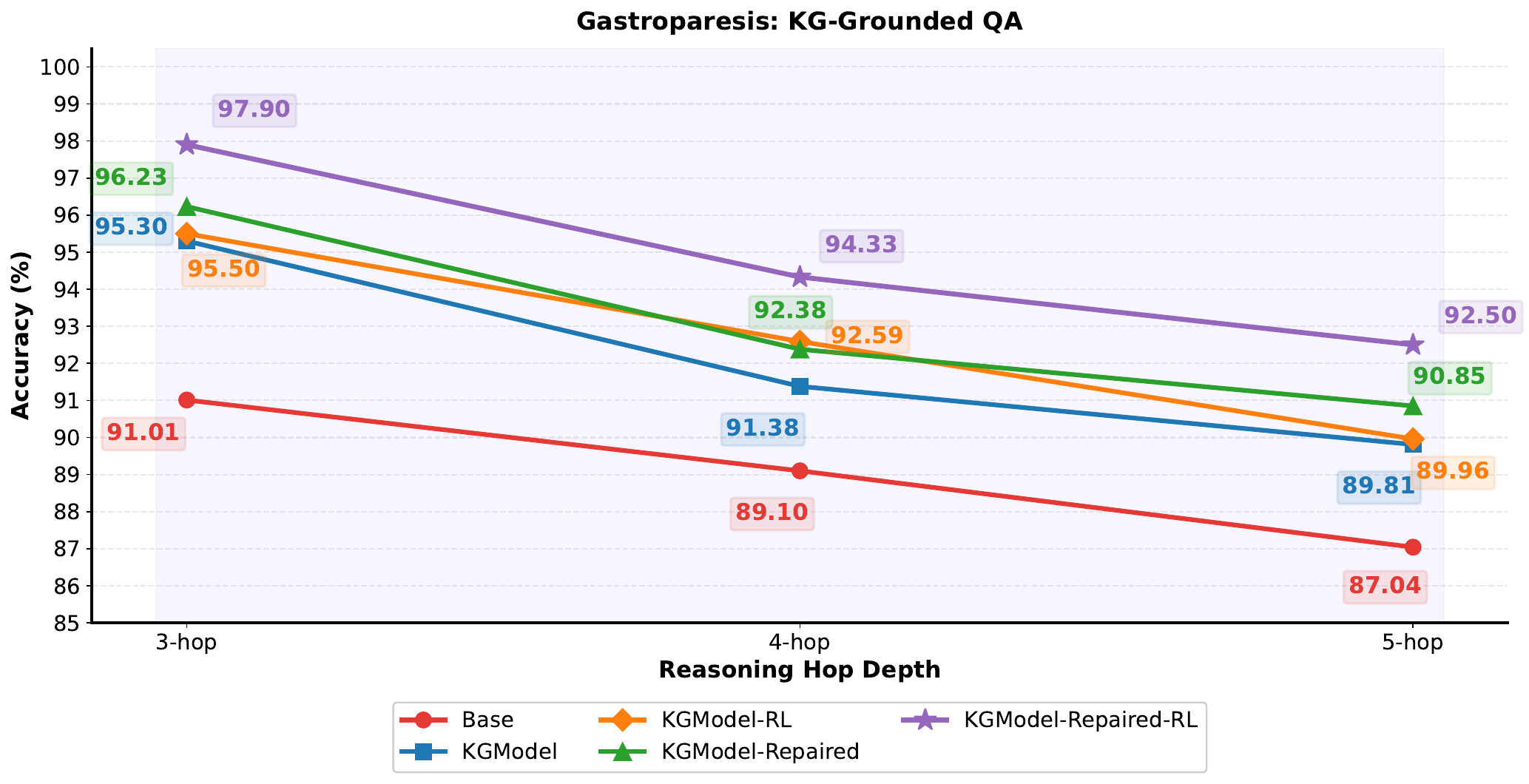}
    \caption{Comparison of RL initialization settings on Gastroparesis KG-grounded QA. The figure compares the Base model, KGModel, KGModel-RL, KGModel-Repaired, and KGModel-Repaired-RL across 3-hop, 4-hop, and 5-hop reasoning. RL initialized from the repaired SFT model consistently achieves higher accuracy than RL initialized from the original SFT model.}
    \label{fig:gastro_kg_rl_init}
\end{figure}

Before presenting the aggregate comparison, Fig.~\ref{fig:gastro_kg_rl_init} shows a representative example for the Gastroparesis KG-grounded setting. The figure illustrates how performance evolves from KGModel to KGModel-RL and from KGModel-Repaired to KGModel-Repaired-RL across 3-hop, 4-hop, and 5-hop QA, providing an intuitive view of the benefit of repaired RL initialization. 

Table~\ref{tab:repair_before_rl} shows that repaired RL initialization improves both final accuracy and RL gain across all disease and grounding settings. Compared with KGModel-RL and CGModel-RL, the repaired variants consistently achieve higher 3-hop, 4-hop, and 5-hop accuracy, indicating that adaptive repair provides a stronger initialization for RL-based multi-hop generalization.

\begin{table*}[h]
\centering
\caption{Higher-hop QA accuracy comparing RL initialized from original SFT models and repaired SFT models. $\Delta_{\mathrm{RL}}$ denotes the gain from RL over the corresponding SFT checkpoint.}
\label{tab:repair_before_rl}
\small
\renewcommand{\arraystretch}{1.1}
\resizebox{\textwidth}{!}{%
\begin{tabular}{llcccccc}
\toprule
\textbf{Setting} 
& \textbf{Model}
& \textbf{3-hop} & \textbf{$\Delta_{\mathrm{RL}}$}
& \textbf{4-hop} & \textbf{$\Delta_{\mathrm{RL}}$}
& \textbf{5-hop} & \textbf{$\Delta_{\mathrm{RL}}$} \\
\midrule
Gastro KG-grounded 
& KGModel-RL 
& 95.50 & +0.20 \gainlow
& 92.59 & +1.21 \gainmid
& 89.96 & +0.15 \gainlow \\

Gastro KG-grounded 
& KGModel-Repaired-RL 
& \textbf{97.90} & +1.67 \gainmid
& \textbf{94.33} & +1.95 \gainmid
& \textbf{92.50} & +1.65 \gainmid \\

\midrule
Gastro CG-grounded 
& CGModel-RL 
& 96.75 & +0.16 \gainlow
& 93.94 & +0.29 \gainlow
& 90.59 & +0.41 \gainlow \\

Gastro CG-grounded 
& CGModel-Repaired-RL 
& \textbf{98.70} & +1.91 \gainmid
& \textbf{96.15} & +2.37 \gainmid
& \textbf{93.39} & +2.78 \gainmid \\

\midrule
Diab KG-grounded 
& KGModel-RL 
& 95.65 & +0.10 \gainlow
& 93.56 & +0.52 \gainlow
& 90.62 & +0.79 \gainlow \\

Diab KG-grounded 
& KGModel-Repaired-RL 
& \textbf{98.91} & +2.17 \gainmid
& \textbf{96.51} & +2.41 \gainmid
& \textbf{92.80} & +1.31 \gainmid \\

\midrule
Diab CG-grounded 
& CGModel-RL 
& 95.99 & +0.14 \gainlow
& 94.35 & +0.43 \gainlow
& 90.21 & +0.25 \gainlow \\

Diab CG-grounded 
& CGModel-Repaired-RL 
& \textbf{98.79} & +1.90 \gainmid
& \textbf{97.51} & +3.26 \gainhigh
& \textbf{93.95} & +2.14 \gainmid \\
\bottomrule
\end{tabular}%
}
\renewcommand{\arraystretch}{1.1}
\end{table*}

\begin{figure}[h]
    \centering
    \includegraphics[width=1\columnwidth]{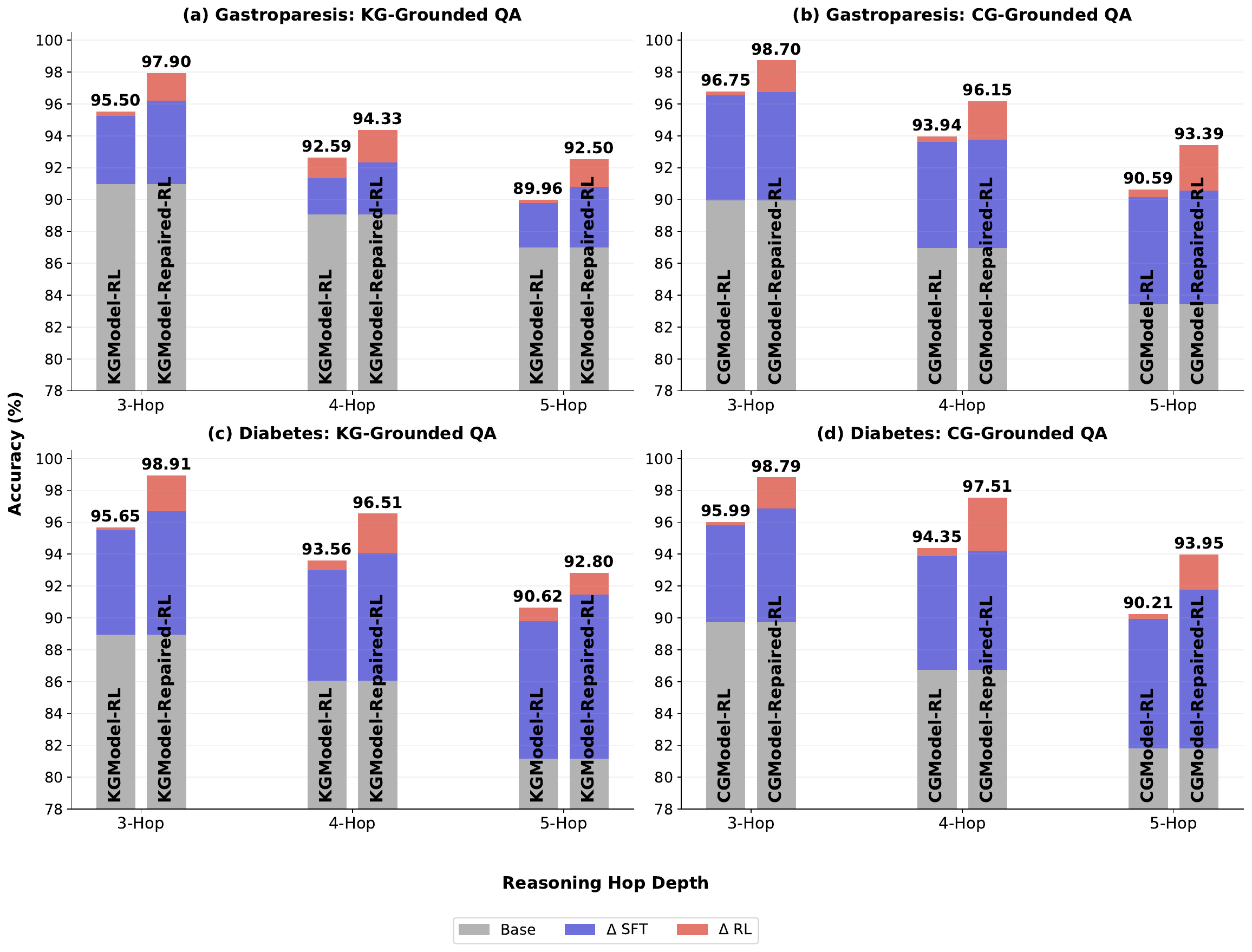}
    \caption{Effect of repaired SFT initialization on RL performance across both diseases and evaluation settings. Each panel compares RL initialized from the original SFT model against RL initialized from the repaired SFT model. For KG-grounded QA, the comparison is between KGModel-RL and KGModel-Repaired-RL; for CG-grounded QA, the comparison is between CGModel-RL and CGModel-Repaired-RL. Gray bars show Base accuracy, blue segments show the gain from SFT, and red segments show the additional gain from RL.}
    \label{fig:repair_before_rl_2x2}
\end{figure}

Fig.~\ref{fig:repair_before_rl_2x2} summarizes the effect of repaired RL initialization across both diseases and both evaluation settings. In all four settings, RL initialized from the repaired SFT model consistently outperforms RL initialized from the original SFT model. This trend is observed across 3-hop, 4-hop, and 5-hop QA. The gains are especially pronounced for the repaired CG-based models, which achieve the strongest overall higher-hop performance.

These results support our hypothesis that unresolved lower-hop errors can limit the effectiveness of RL, whereas starting from a cleaner and more reliable SFT checkpoint provides a stronger foundation for learning compositional reasoning. 

\subsection{Full Comparison Across Model Variants}
\label{sec:results_full_comparison}

We next compare all model variants to summarize the cumulative effect of context augmentation, adaptive repair, and RL. This comparison includes the Base model, the initial SFT models, the repaired SFT models, and the RL models initialized from both unrepaired and repaired checkpoints. Results are reported separately for Gastroparesis and Diabetes, with each model evaluated on both KG-grounded and CG-grounded 3-hop, 4-hop, and 5-hop QA.

\begin{figure}[h]
    \centering
    \includegraphics[width=1\columnwidth]{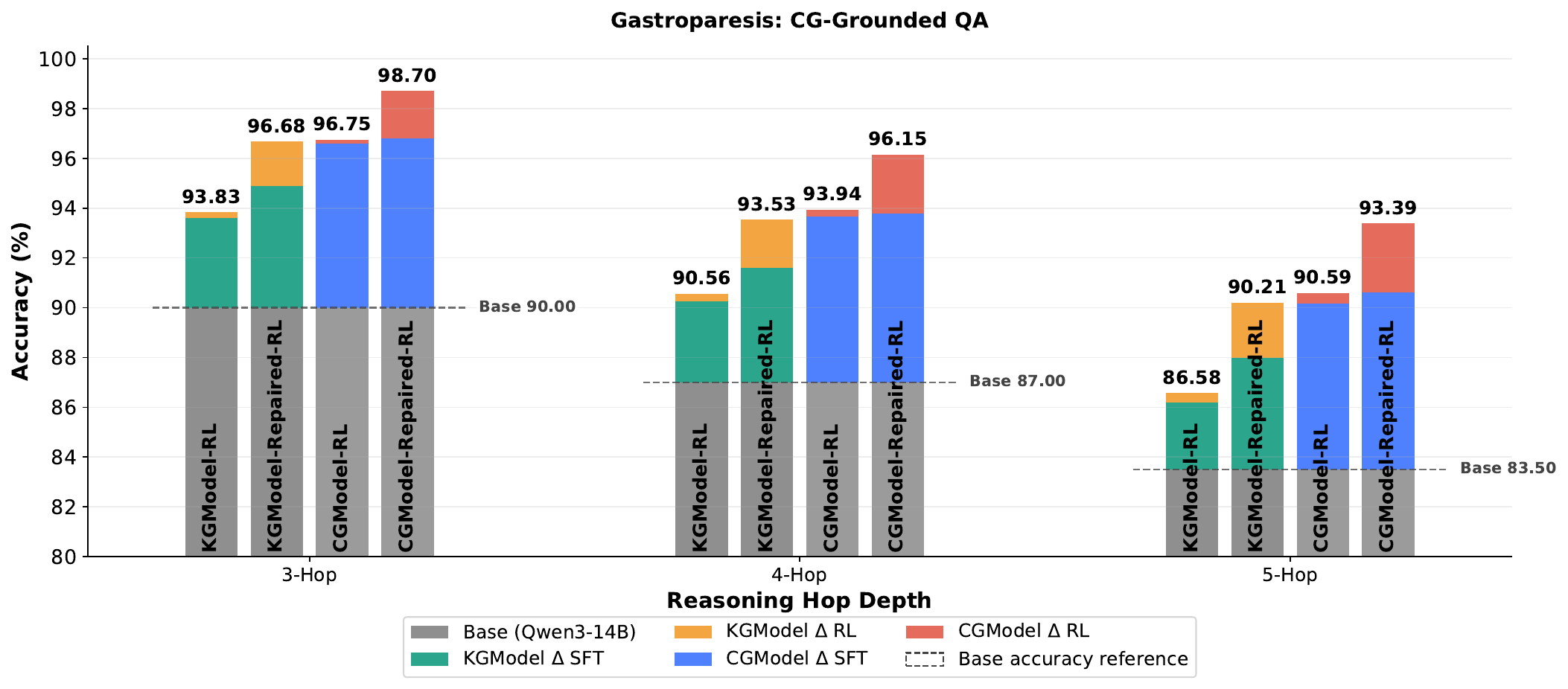}
    \caption{Representative comparison of all nine model variants on Gastroparesis KG-grounded QA across 3-hop, 4-hop, and 5-hop reasoning. The figure illustrates the cumulative effect of SFT, adaptive repair, and RL under both KGModel and CGModel branches, using the Base model as the reference.}
    \label{fig:gastro_kg_all_models}
\end{figure}

Fig.~\ref{fig:gastro_kg_all_models} provides a representative summary of the nine-model progression for the Gastroparesis CG-grounded evaluation setting. The figure highlights how performance changes across the full pipeline, from the Base model to the SFT, repaired SFT, and RL variants under both KGModel and CGModel branches. 

Table~\ref{tab:gastro_full_comparison} presents the full comparison for gastroparesis. The corresponding full results for diabetes are provided in Appendix~\ref{app:diab_results}.
For each RL row, $\Delta_{\mathrm{RL}}$ denotes the per-hop gain over the corresponding non-RL checkpoint. For example, KGModel-RL is compared against KGModel, while KGModel-Repaired-RL is compared against KGModel-Repaired.

Across both diseases and evaluation settings, the per-hop $\Delta_{\mathrm{RL}}$ columns show that RL initialized from unrepaired SFT checkpoints gives small or inconsistent gains, whereas RL initialized from repaired checkpoints produces larger and more consistent improvements. This trend is especially clear for the CGModel branch, where CGModel-Repaired-RL produces positive gains for every hop across both diseases and evaluation settings. Overall, CGModel-Repaired-RL achieves the best accuracy for every 3-hop, 4-hop, and 5-hop task, indicating that context-augmented SFT, adaptive repair, and repaired RL initialization provide complementary benefits.

\begin{table}[H]
\centering
\caption{Gastroparesis higher-hop QA accuracy across all model variants. $\Delta_{\mathrm{RL}}$ denotes the per-hop gain from RL over the corresponding non-RL checkpoint. Bold accuracy values indicate the highest accuracy for each evaluation setting and hop depth, and underlined values indicate the second-highest accuracy.}
\label{tab:gastro_full_comparison}
\footnotesize
\renewcommand{\arraystretch}{1.2}
\setlength{\tabcolsep}{2.2pt}
\begin{tabular*}{\textwidth}{@{\extracolsep{\fill}}lcccccccccccc}
\toprule
\textbf{Model}
& \multicolumn{6}{c}{\textbf{KG-grounded QA}}
& \multicolumn{6}{c}{\textbf{CG-grounded QA}} \\
\cmidrule(lr){2-7}
\cmidrule(lr){8-13}
& \textbf{3-hop} & \textbf{$\Delta_3$}
& \textbf{4-hop} & \textbf{$\Delta_4$}
& \textbf{5-hop} & \textbf{$\Delta_5$}
& \textbf{3-hop} & \textbf{$\Delta_3$}
& \textbf{4-hop} & \textbf{$\Delta_4$}
& \textbf{5-hop} & \textbf{$\Delta_5$} \\
\midrule
Base 
& 91.01 &  & 89.10 &  & 87.04 &  
& 90.00 &  & 87.00 &  & 83.50 &  \\
\midrule

\rowcolor{kgplain}
KGModel 
& 95.30 &  & 91.38 &  & 89.81 &  
& 93.59 &  & 90.26 &  & 86.20 &  \\

\rowcolor{kgplain}
KGModel-RL 
& 95.50 & +0.20 & 92.59 & +1.21 & 89.96 & +0.15 
& 93.83 & +0.24 & 90.56 & +0.30 & 86.58 & +0.38 \\

\rowcolor{kgrepair}
KGModel-Repaired 
& 96.23 &  & 92.38 &  & 90.85 &  
& 94.88 &  & 91.61 &  & 87.98 &  \\

\rowcolor{kgrepair}
KGModel-Repaired-RL 
& \underline{97.90} & \textbf{+1.67} & 94.33 & \textbf{+1.95} & \underline{92.50} & \textbf{+1.65}
& 96.68 & \textbf{+1.80} & 93.53 & \textbf{+1.92} & 90.21 & \textbf{+2.23} \\

\midrule

\rowcolor{cgplain}
CGModel 
& 97.28 &  & 93.58 &  & 91.96 &  
& 96.59 &  & 93.65 &  & 90.18 &  \\

\rowcolor{cgplain}
CGModel-RL 
& 97.78 & +0.50 & \underline{94.36} & +0.78 & 92.11 & +0.15
& 96.75 & +0.16 & \underline{93.94} & +0.29 & 90.59 & +0.41 \\

\rowcolor{cgrepair}
CGModel-Repaired 
& 97.39 &  & 93.67 &  & 92.08 &  
& \underline{96.79} &  & 93.78 &  & \underline{90.61} &  \\

\rowcolor{cgrepair}
CGModel-Repaired-RL 
& \textbf{98.88} & \textbf{+1.49} & \textbf{95.56} & \textbf{+1.89} & \textbf{94.19} & \textbf{+2.11}
& \textbf{98.70} & \textbf{+1.91} & \textbf{96.15} & \textbf{+2.37} & \textbf{93.39} & \textbf{+2.78} \\
\bottomrule
\end{tabular*}
\renewcommand{\arraystretch}{1.1}
\setlength{\tabcolsep}{6pt}
\normalsize
\end{table}

\subsection{Robustness to Option Shuffling}
\label{sec:results_shuffle}

Finally, we evaluate whether the observed improvements are robust to answer-option ordering. We apply an option-shuffling stress test in which all four answer-option texts are randomly reassigned to option labels A-D under a fixed seed, while the question stem remains unchanged. We then compare each model's accuracy before and after option shuffling on 3-hop, 4-hop, and 5-hop QA. Table~\ref{tab:shuffle_gastro_cg_full} reports the original accuracy, shuffled accuracy, and accuracy change for the Gastroparesis CG-grounded evaluation setting. The remaining disease/evaluation settings are reported in Appendix~\ref{app:option_shuffling_results}.

Overall, the option-shuffling results show that most models remain relatively stable under answer-option reordering. Across all settings, the average absolute change is below one percentage point for every model,
indicating that the evaluated models are not strongly dependent on fixed answer-option positions.

\FloatBarrier
\begin{table}[H]
\centering
\caption{Option-shuffling robustness on Gastroparesis CG-grounded QA. $\Delta$ denotes shuffled accuracy minus original accuracy in percentage points.}
\label{tab:shuffle_gastro_cg_full}
\small
\renewcommand{\arraystretch}{1.1}
\setlength{\tabcolsep}{4.2pt}
\begin{tabular*}{\textwidth}{@{\extracolsep{\fill}}lcccccccccc}
\toprule
\textbf{Model} 
& \multicolumn{3}{c}{\textbf{3-hop}}
& \multicolumn{3}{c}{\textbf{4-hop}}
& \multicolumn{3}{c}{\textbf{5-hop}}
& \textbf{Avg.} \\
\cmidrule(lr){2-4}
\cmidrule(lr){5-7}
\cmidrule(lr){8-10}
& \textbf{Orig.} & \textbf{Shuf.} & \textbf{$\Delta$}
& \textbf{Orig.} & \textbf{Shuf.} & \textbf{$\Delta$}
& \textbf{Orig.} & \textbf{Shuf.} & \textbf{$\Delta$}
& \textbf{$\Delta$} \\
\midrule
Base & 90.00 & 90.44 & +0.44 & 87.00 & 87.25 & +0.25 & 83.50 & 84.39 & +0.89 & +0.53 \\
KGModel & 93.59 & 94.15 & +0.56 & 90.26 & 91.03 & +0.77 & 86.20 & 87.06 & +0.86 & +0.73 \\
KGModel-RL & 93.83 & 94.04 & +0.21 & 90.56 & 89.86 & -0.70 & 86.58 & 86.44 & -0.14 & -0.21 \\
KGModel-Repaired & 94.88 & 95.16 & +0.29 & 91.61 & 92.51 & +0.90 & 87.98 & 87.22 & -0.76 & +0.14 \\
KGModel-Repaired-RL & 96.68 & 95.89 & -0.79 & 93.53 & 92.97 & -0.64 & 90.21 & 90.24 & +0.03 & -0.47 \\
CGModel & 96.59 & 95.88 & -0.71 & 93.65 & 92.62 & -1.03 & 90.18 & 89.66 & -0.52 & -0.88 \\
CGModel-RL & 96.75 & 95.22 & -0.53 & 93.94 & 92.89 & -1.05 & 90.59 & 89.67 & -0.92 & -0.83 \\
CGModel-Repaired & 96.79 & 96.04 & -0.75 & 93.78 & 93.49 & -0.29 & 90.60 & 90.69 & +0.09 & -0.32 \\
CGModel-Repaired-RL & 98.70 & 98.11 & -0.59 & 96.25 & 95.14 & -1.11 & 93.39 & 92.53 & -0.86 & -0.85 \\
\bottomrule
\end{tabular*}
\renewcommand{\arraystretch}{1.0}
\setlength{\tabcolsep}{6pt}
\end{table}



\section{Discussion}
\label{sec:discussion}


This section discusses the main findings of the study, focusing on the effects of context-augmented supervision, adaptive repair, and repaired RL initialization on multi-hop QA performance.

\subsection{Context-Augmented Supervision Helps}



One important finding is that CGModel outperforms KGModel across both diseases and evaluation settings, suggesting that context augmentation provides stronger training supervision. This benefit likely comes from presenting each target fact with richer and more clinically meaningful local context. We represent this context using supporting triples because they are structured, traceable, and directly available from the GraphMERT extraction pipeline. Thus, the key insight is that treating triples as part of a local evidence neighborhood, rather than as isolated facts, can improve how the model learns and composes relations.

\subsection{Adaptive Repair Is Necessary}

The adaptive repair results show that post-SFT validation failures should not be treated uniformly. An incorrect answer can arise from several different sources: the source triple may be noisy, the validation question may be misaligned, or there may be a genuine gap in the model's knowledge. Therefore, blindly fine-tuning on every incorrect validation item risks reinforcing noisy or unsupported supervision. The LLM-judged repair pipeline addresses this problem by first diagnosing the source of each failure and then applying targeted repair only to valid missing-knowledge cases.

\subsection{Repair Before RL Improves Multi-Hop Reasoning}

A key finding is that RL is substantially more effective when initialized from repaired SFT checkpoints, supporting the hypothesis that multi-hop reasoning depends on reliable lower-hop knowledge.
Since a 3-hop, 4-hop, or 5-hop reasoning path is composed of multiple lower-hop factual links, unresolved 1-hop errors can propagate when the model is later trained to reason over longer paths. In this setting, RL does not operate on a clean factual foundation; instead, it may reinforce incomplete or unstable reasoning patterns that originate from unresolved lower-hop knowledge gaps. This is consistent with recent findings that RL-based reasoning improvements can be strongly shaped by the SFT initialization and may remain limited by the solution space already induced before RL~\cite{HavrillaRLReasoning}.

The per-hop $\Delta_{\mathrm{RL}}$ columns in Tables~\ref{tab:gastro_full_comparison} and~\ref{tab:diabetes_full_comparison} make this effect explicit. Across both diseases and evaluation settings, RL initialized from unrepaired SFT checkpoints gives small or inconsistent gains, whereas RL initialized from repaired checkpoints produces larger and more stable improvements. For example, in the Gastroparesis CG-grounded setting, CGModel-RL improves over CGModel by an average of only $+0.29$ percentage points across 3-hop, 4-hop, and 5-hop QA, whereas CGModel-Repaired-RL improves over CGModel-Repaired by an average of $+2.35$ points. This indicates that repair does not merely improve the SFT checkpoint; it also makes RL more effective for higher-hop reasoning.

These results suggest that adaptive repair changes what RL is optimizing. Before repair, the reward signal is applied to a model with unresolved factual gaps. Hence, improvement can be limited by errors inherited from the SFT stage. After repair, the remaining RL objective is more focused: it can reward the composition of already-learned relations rather than compensating for missing lower-hop knowledge.


\subsection{Combined Effect of Context, Repair, and RL}

The full model comparison shows that CGModel-Repaired-RL is consistently the strongest model, indicating that the best performance comes from the ordered combination of all three components rather than from any single stage alone. Context augmentation improves the intial SFT stage, adaptive repair produces a cleaner lower-hop checkpoint, and RL further improves higher-hop generalization when initialized from that repaired checkpoint. Thus, the complete pipeline is best understood as a staged learning process: first learn basic facts, then repair unresolved reliable facts, and finally optimize for compositional reasoning.

\subsection{Robustness to Option Shuffling}

The option-shuffling analysis suggests that the observed gains are not primarily caused by fixed answer-position artifacts. Although some repaired RL models show small negative shifts after option shuffling, the average absolute change remains below one percentage point across models. This indicates that the models are reasonably stable under distractor reordering and that the main performance improvements are likely due to improved reasoning rather than option-position bias.

\section{Limitations and Future Work}
\label{sec:limitations}

Although our work shows that context-augmented KG supervision, adaptive repair, and repaired RL initialization can improve multi-hop QA, several constraints remain and motivate future extensions. First, the QA datasets used for SFT, repair, RL, and evaluation are synthetically generated. We reduce the risk of noisy supervision through a multi-stage generation and verification pipeline, including MCQ format checking, quality filtering, thinking-trace generation, and triple-grounded correctness filtering. The adaptive repair stage further audits failed validation cases before using them for targeted repair. Nevertheless, because the datasets are synthetic, some residual ambiguity or generation noise may remain. Future work could incorporate expert review for a subset of generated QA items, especially for difficult repaired or quarantined triples.

Second, the adaptive repair pipeline relies on LLM judges. In this work, Judge 1 uses \texttt{Qwen3.6-27B} for the initial failure audit, while Judge 2 uses \texttt{Qwen3-32B} for the final quarantine review. This design provides scalable failure diagnosis and reduces the need for repeated manual inspection, but judge decisions may still reflect model-specific biases. A natural extension is to use stronger judge models, ensembles of judges, or judges specifically fine-tuned for biomedical KG validation and QA-quality assessment. Prior work, such as JudgeLM, has shown that LLMs can be fine-tuned as scalable judges, suggesting that task-specific judge training may further improve the reliability of automated validation in future versions of this pipeline~\cite{JudgeLMTuned}.

Third, although we evaluate the framework on two disease settings, Gastroparesis and Diabetes, broader validation is still needed. Gastroparesis serves as the main disease-specific case study and Diabetes is used as an additional setting to test whether the observed gains are specific to a single KG or disease domain. The fact that the main trends hold across both diseases strengthens the empirical evidence, but future work should evaluate the framework on additional KGs and more diverse relation sets.

Finally, context augmentation in this work is limited to supporting triples extracted from the same source text chunk. This choice makes the context structured, traceable, and compatible with KG-grounded QA generation. However, future extensions could incorporate richer forms of evidence, such as source-text passages, clinical notes, electronic health records, laboratory findings, imaging reports, treatment histories, biomedical figures, or clinical guideline snippets. As long as the added context is reliable and relevant to the target fact, richer context may provide stronger supervision and further improve multi-hop reasoning.


\section{Conclusion}
\label{sec:conclusion}

This work presents a context-augmented and repair-aware adaptive training framework for multi-hop question-answering. Starting with KGs, we construct CGs by attaching supporting triples from the same source text chunks to each primary triple. These context-augmented triples are then used to generate structured QA supervision for SFT, adaptive repair, and RL.

Our results show three main findings. First, context-augmented SFT improves higher-hop QA performance compared with SFT using isolated KG triples. This indicates that training with supporting context helps the model learn relations in a more meaningful local evidence neighborhood. Second, LLM-judged adaptive repair provides a cleaner lower-hop foundation by separating valid missing-knowledge cases from noisy triples and misaligned validation questions. Third, RL is most effective when initialized from repaired SFT checkpoints. Across both Gastroparesis and Diabetes, the strongest performance is consistently achieved by CGModel-Repaired-RL, showing that context augmentation, adaptive repair, and repaired RL initialization provide complementary benefits.

Overall, these findings suggest that improving multi-hop reasoning requires more than simply applying RL after SFT. The model first needs reliable lower-hop knowledge, and that knowledge can be strengthened through context-aware supervision and targeted repair. Future work can extend this framework to additional KGs, stronger judge models, and richer forms of context beyond supporting triples.

\section{Acknowledgments}
The experiments reported in this
paper were performed using the Princeton Research Computing resources at Princeton University. Princeton Research Computing is a consortium of groups including the Princeton Institute for Computational Science and Engineering (PICSciE) and Research Computing at Princeton University.
\bibliographystyle{IEEEtran}
\bibliography{references}

\section{Appendices}

\subsection{GraphMERT-based KG Extraction}
\label{app:graphmert_details}





\begin{figure}[h]
    \centering
    \includegraphics[width=0.95\linewidth]{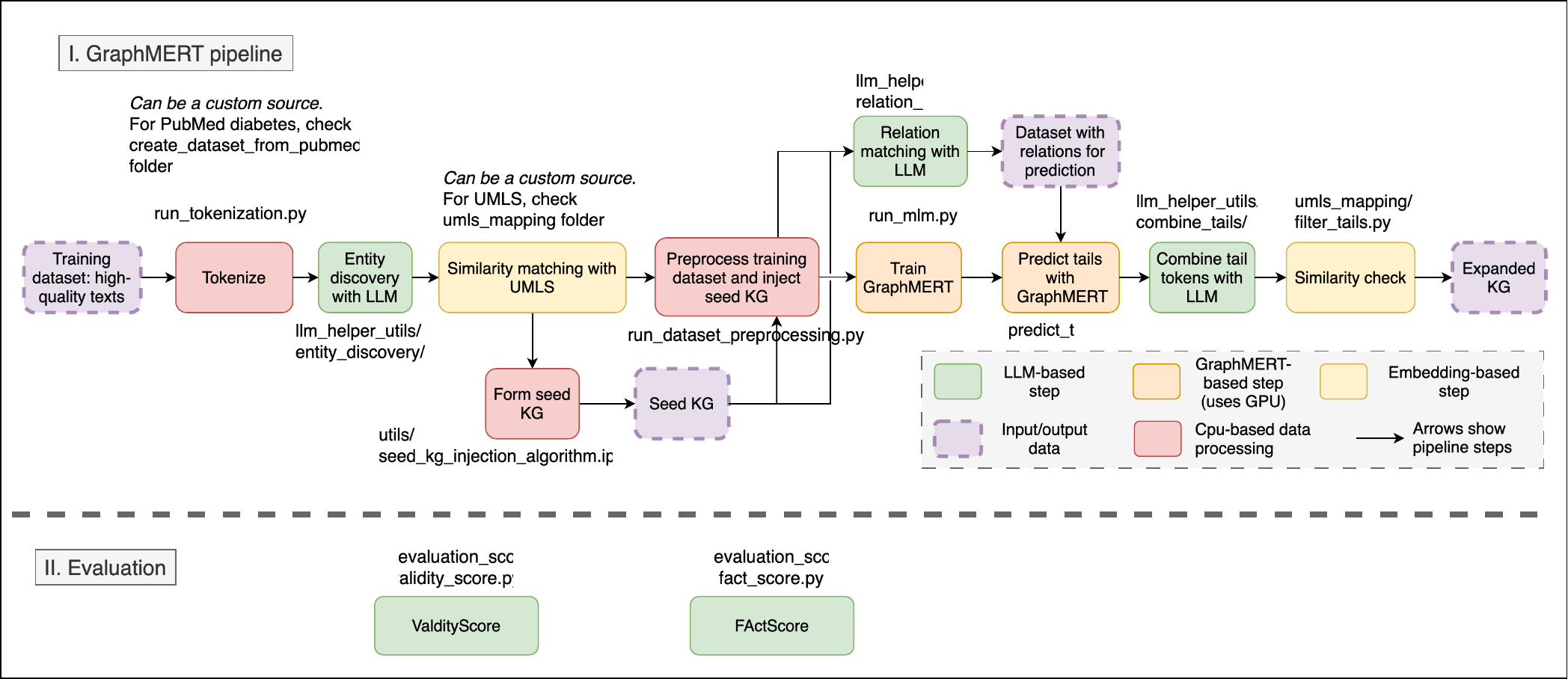}
    \caption{Original GraphMERT pipeline extracted from Belova et al.~\cite{margarita}.}
    \label{fig:appendix_graphmert_pipeline}
\end{figure}

This appendix provides implementation details for the GraphMERT-based KG construction pipeline used in this work. The overall procedure follows the GraphMERT framework~\cite{margarita}. Fig.~\ref{fig:appendix_graphmert_pipeline} shows the original GraphMERT pipeline extracted from Belova et al.~\cite{margarita}. 

\subsubsection{Paper Retrieval from PubMed Central}

Gastroparesis-related articles were retrieved from PubMed Central using the following query:

\begin{tcolorbox}[
    breakable,
    colback=green!4,
    colframe=black!50,
    title={PubMed Central Search Query},
    fonttitle=\bfseries,
    arc=1.5mm,
    boxrule=0.5pt
]
\small
\ttfamily
(gastroparesis[tiab] OR gastroparesis[body] OR ``Gastroparesis''[mh] \\
OR ``gastric stasis''[tiab] OR ``gastric stasis''[body] \\
OR ``delayed gastric emptying''[tiab] OR ``delayed gastric emptying''[body] \\
OR ``gastric emptying''[tiab] OR ``gastric emptying''[body]) \\
NOT (covid-19 OR sars-cov-2 in tiab, body, or MeSH)
\normalfont
\end{tcolorbox}

COVID-19 and SARS-CoV-2 articles were excluded to avoid introducing virus-related gastrointestinal associations that could confound the disease-specific gastroparesis KG.

Table~\ref{tab:pmc_retrieval_summary} summarizes the PubMed Central (PMC) retrieval process used to construct the gastroparesis corpus. The PMC search returned 41,230 PMCIDs. Querying the PMC open-access (OA) service partitioned these records into 27,102 PMCIDs with a downloadable OA package, 13,936 PMCIDs without an OA file, and 192 unresolved PMCIDs due to application programming interface (API) or identifier errors. 

\begin{table}[h]
\small
\centering
\caption{Summary of PMC retrieval and filtering for the gastroparesis corpus.}
\label{tab:pmc_retrieval_summary}
\begin{tabular}{llr}
\hline
Stage & Category & Count \\
\hline
PMC search & Returned PMCIDs & 41,230 \\
\hline
OA availability check & Downloadable OA package found & 27,102 \\
OA availability check & No OA file available & 13,936 \\
OA availability check & Unresolved due to API/ID error & 192 \\
\hline
OA package processing & Downloadable OA packages & 27,102 \\
OA package processing & Failed archive extraction & 133 \\
OA package processing & Successfully extracted folders & 26,969 \\
\hline
Final filtering & Removed non-English or malformed records & 186 \\
Final filtering & Final article folders used & 26,783 \\
\hline
\end{tabular}
\end{table}

\subsubsection{Cleaning and Abstract Dataset Construction}

Only abstracts were used for GraphMERT training. Articles were removed if the abstract was shorter than 200 characters, non-English, or associated with a missing or broken \texttt{.nxml} file. Language detection was performed using \texttt{langdetect}. This produced a non-English or bad-file list of 319 PMCIDs. Table~\ref{tab:appendix_abstract_processing} summarizes the settings used.

\begin{table}[h]
\centering
\caption{Abstract cleaning and dataset construction settings.}
\label{tab:appendix_abstract_processing}
\small
\begin{tabular}{lp{0.62\linewidth}}
\toprule
\textbf{Setting} & \textbf{Value} \\
\midrule
Input text & Abstracts only \\
Text normalization & \texttt{BertNormalizer} with cleaning, Chinese-character handling, and lowercasing \\
Lead phrase removal & Regex removal of phrases such as Background, Abstract, and Purpose \\
Fallback abstract extraction & JATS \texttt{<abstract>} $\rightarrow$ \texttt{sec-type=abstract} $\rightarrow$ titled Abstract $\rightarrow$ first 1--3 body paragraphs \\
Shuffle seed & 1331 \\
Train / eval abstracts & 21,426 / 5,357 \\
\bottomrule
\end{tabular}
\end{table}

\subsubsection{Tokenization}

Abstracts were tokenized using PubMedBERT with a maximum sequence length of 128 tokens. Long abstracts were split into multiple token sequences. Tokenization statistics are given in Table~\ref{tab:appendix_tokenization}.
\begin{table}[h]
\centering
\caption{Tokenization statistics for the gastroparesis corpus.}
\label{tab:appendix_tokenization}
\small
\begin{tabular}{lcc}
\toprule
\textbf{Split} & \textbf{Abstracts} & \textbf{Token sequences} \\
\midrule
Train & 21,425 & 55,715 \\
Eval  & 5,357  & 13,782 \\
\bottomrule
\end{tabular}
\end{table}

\subsubsection{Entity Discovery and Seed KG Construction}

Head entities were discovered using \texttt{Qwen3-32B-FP8} through vLLM. The helper LLM was used for entity discovery, relation assignment, and tail-token combination, while GraphMERT was used for KG completion.

Tables~\ref{tab:appendix_entity_discovery}-\ref{tab:appendix_seed_kg_stats} summarize the helper-LLM configuration used for entity discovery, the entity-linking and seed-KG construction settings, and the resulting seed-KG statistics used for GraphMERT training.

\begin{table}[H]
\centering
\caption{Helper LLM configuration for entity discovery.}
\label{tab:appendix_entity_discovery}
\small
\begin{tabular}{lc}
\toprule
\textbf{Parameter} & \textbf{Value} \\
\midrule
Model & \texttt{Qwen3-32B-FP8} \\
Inference engine & vLLM \\
Temperature & 0.6 \\
Top-$p$ & 0.95 \\
Top-$k$ & 20 \\
Maximum tokens & 8192 \\
Maximum model length & 8192 \\
\bottomrule
\end{tabular}
\end{table}

\begin{table}[H]
\centering
\caption{Entity linking and seed KG construction settings.}
\label{tab:appendix_seed_kg_settings}
\small
\begin{tabular}{lc}
\toprule
\textbf{Component} & \textbf{Setting} \\
\midrule
Concept embedder & \texttt{SapBERT-from-PubMedBERT-fulltext} \\
FAISS candidates & Top 50, reduced to final 40 \\
SapBERT similarity threshold & 0.85 \\
Character 3-gram Jaccard threshold & 0.5 \\
Embedding ranker & \texttt{gemini-embedding-001} \\
Document/triple cosine threshold & 0.7 \\
Seed KG cosine cutoff $\alpha$ & 0.55 \\
Maximum triples per head & 4, followed by diversity bucketing \\
\bottomrule
\end{tabular}
\end{table}

\begin{table}[H]
\centering
\caption{Seed KG statistics used for GraphMERT training.}
\label{tab:appendix_seed_kg_stats}
\small
\begin{tabular}{lc}
\toprule
\textbf{Artifact} & \textbf{Count} \\
\midrule
Seed KG unique triples & 4,751 \\
Relation types & 20 \\
Relation labels including padding & 21 \\
Train injections & 5,093 \\
Eval injections & 957 \\
\bottomrule
\end{tabular}
\end{table}

\subsubsection{GraphMERT Training}

GraphMERT was trained from scratch on the gastroparesis corpus using the injected seed KG. The model and training hyperparameters are summarized in Table~\ref{tab:appendix_graphmert_hparams}.

\begin{table}[h]
\centering
\caption{GraphMERT model and training hyperparameters.}
\label{tab:appendix_graphmert_hparams}
\small
\begin{tabular}{lc}
\toprule
\textbf{Hyperparameter} & \textbf{Value} \\
\midrule
Layers/heads/hidden size & 6/8/512 \\
Intermediate size & 2048 \\
Vocabulary size & 30,522 \\
Maximum nodes & 1024 \\
Root nodes & 128 \\
Leaf count & 7 \\
Graph types & root-undirected, leaf-undirected, leaf-connected-undirected \\
MLM + SBO & Enabled \\
\texttt{exp\_mask\_base} & 0.6 \\
\texttt{mlm\_on\_leaves\_probability} & 0.20 \\
Relation embedding dropout & 0.35 \\
Learning rate & $5\times10^{-5}$ \\
LR schedule & Cosine with minimum LR $1\times10^{-6}$ \\
Epochs & 25 \\
Training steps & 21,775 \\
Per-device batch size & 32 \\
Gradient accumulation & 2 \\
Warmup steps & 500 \\
Weight decay & 0.01 \\
Precision & bf16 \\
Random seed & 0 \\
\bottomrule
\end{tabular}
\end{table}

\subsubsection{Prediction-Time Triple Generation and Filtering}

After GraphMERT training, candidate relations were first added using \texttt{Qwen3-32B-FP8}, restricted to the 20 seed relation types. This produced 11,199 cleaned sequences with valid heads and relations. GraphMERT then predicted candidate tail tokens for each head-relation pair. The helper LLM combined predicted tail tokens into coherent biomedical tail entities. Finally, \texttt{gemini-embedding-001} was used to score similarity between the source text and the generated triple, and triples with similarity score greater than $\beta=0.65$ were retained. A summary is provided in Table~\ref{tab:appendix_prediction_filtering}.

\begin{table}[H]
\centering
\caption{Prediction-time triple generation and filtering summary.}
\label{tab:appendix_prediction_filtering}
\small
\begin{tabular}{lc}
\toprule
\textbf{Stage} & \textbf{Output} \\
\midrule
Relation adding model & \texttt{Qwen3-32B-FP8} \\
Allowed relation set & 20 seed relations \\
Cleaned valid head--relation sequences & 11,199 \\
Tail prediction model & GraphMERT \\
Tail combination model & \texttt{Qwen3-32B-FP8} \\
Final triple filter & \texttt{gemini-embedding-001} cosine similarity \\
Filtering threshold $\beta$ & 0.65 \\
Final generated triple rows & 6,900 \\
Final unique triples after deduplication & 6,018 \\
\bottomrule
\end{tabular}
\end{table}

\subsubsection{Models Used in the Gastroparesis KG Construction Run}

Table~\ref{tab:appendix_graphmert_models} summarizes the models used at each stage of the gastroparesis KG construction run.

\begin{table}[h]
\centering
\caption{Models used in the gastroparesis GraphMERT KG construction run.}
\label{tab:appendix_graphmert_models}
\small
\renewcommand{\arraystretch}{1.1}
\begin{tabularx}{\linewidth}{p{0.34\linewidth}X}
\toprule
\textbf{Role} & \textbf{Model} \\
\midrule
Tokenizer / text encoder base 
& PubMedBERT \\

Entity linking 
& SapBERT \\

Embedding ranking and final filtering 
& \texttt{gemini-embedding-001} \\

Helper LLM for heads, relations, and tail combination 
& \texttt{Qwen3-32B-FP8} through vLLM \\

KG completion model 
& GraphMERT trained from scratch on the gastroparesis corpus \\
\bottomrule
\end{tabularx}
\renewcommand{\arraystretch}{1.0}
\end{table}

\subsubsection{Diabetes KG Subgraph Selection}

For the diabetes setting, we use an adapted subgraph from the original GraphMERT diabetes KG~\cite{margarita}. Since the original diabetes KG is larger than the gastroparesis KG, we construct a size-matched diabetes subset to make the two disease settings comparable. The subset is not sampled uniformly at random; instead, it is selected using a relation-aware strategy that preserves relation diversity while maintaining graph connectivity.


The relation-aware selection strategy is shown below. It is used because uniform random sampling can remove rare relation types and produce a less connected subgraph. In contrast, the proposed strategy keeps rare relations represented while also encouraging selected triples to reconnect to the existing node set. The final diabetes subset contains 6,000 sampled triple rows, corresponding to 5,954 unique triples after deduplication.

\begin{tcolorbox}[
    colback=green!4,
    colframe=black!50,
    title={Diabetes KG Subgraph Selection Strategy},
    fonttitle=\bfseries,
    arc=1.5mm,
    boxrule=0.5pt
]
\small
\begin{enumerate}
    \item \textbf{Stratify by relation:} assign each relation a quota using a minimum floor of 20 triples, or all available triples if fewer than 20 exist.
    \item \textbf{Allocate remaining budget:} distribute the remaining sampling budget proportionately based on relation frequency.
    \item \textbf{Preserve connectivity:} within each relation, sample triples in the following order: both endpoints already selected, one endpoint already selected, and neither endpoint selected.
    \item \textbf{Fill quotas:} process larger-quota relations first and fill each quota tier-by-tier using random seed 42.
\end{enumerate}
\normalsize
\end{tcolorbox}

\subsubsection{Relation Frequencies}

Table~\ref{tab:relation_freq_side_by_side} summarizes the relation-frequency distributions of the final disease-specific KGs. The gastroparesis KG has a broader spread across relation types, whereas the diabetes KG is dominated by the \texttt{associated\_with} relation, consistent with the original GraphMERT diabetes KG~\cite{margarita}.

\begin{table*}[h]
\centering
\caption{Relation-frequency distributions of the final gastroparesis and diabetes KGs.}
\label{tab:relation_freq_side_by_side}
\scriptsize
\resizebox{\textwidth}{!}{%
\begin{tabular}{lcc|lcc}
\toprule
\multicolumn{3}{c|}{\textbf{Gastroparesis KG}} 
& \multicolumn{3}{c}{\textbf{Diabetes KG}} \\
\cmidrule(lr){1-3}
\cmidrule(lr){4-6}
\textbf{Relation} & \textbf{Count} & \textbf{\%} 
& \textbf{Relation} & \textbf{Count} & \textbf{\%} \\
\midrule
plays\_role & 1342 & 22.30 
& associated\_with & 3747 & 62.93 \\

cause\_of & 762 & 12.66 
& plays\_role & 383 & 6.43 \\

has\_finding\_site & 565 & 9.39 
& cause\_of & 307 & 5.16 \\

has\_disposition & 474 & 7.88 
& isa & 165 & 2.77 \\

is\_interpreted\_by & 426 & 7.08 
& has\_component & 133 & 2.23 \\

inverse\_isa & 426 & 7.08 
& focus\_of & 131 & 2.20 \\

associated\_finding\_of & 368 & 6.11 
& due\_to & 115 & 1.93 \\

has\_method & 261 & 4.34 
& has\_disposition & 113 & 1.90 \\

has\_modification & 230 & 3.82 
& has\_associated\_morphology & 102 & 1.71 \\

has\_direct\_procedure\_site & 228 & 3.79 
& part\_of & 86 & 1.44 \\

isa & 184 & 3.06 
& interprets & 72 & 1.21 \\

focus\_of & 182 & 3.02 
& has\_finding\_site & 70 & 1.18 \\

interprets & 180 & 2.99 
& has\_part & 63 & 1.06 \\

finding\_site\_of & 113 & 1.88 
& inverse\_isa & 56 & 0.94 \\

has\_associated\_morphology & 109 & 1.81 
& is\_interpreted\_by & 44 & 0.74 \\

direct\_procedure\_site\_of & 48 & 0.80 
& causative\_agent\_of & 42 & 0.71 \\

possibly\_equivalent\_to & 39 & 0.65 
& has\_method & 39 & 0.66 \\

causative\_agent\_of & 38 & 0.63 
& associated\_morphology\_of & 37 & 0.62 \\

associated\_morphology\_of & 37 & 0.61 
& has\_causative\_agent & 37 & 0.62 \\

disposition\_of & 6 & 0.10 
& has\_modification & 30 & 0.50 \\

 &  &  
& has\_focus & 28 & 0.47 \\

 &  &  
& associated\_finding\_of & 27 & 0.45 \\

 &  &  
& same\_as & 23 & 0.39 \\

 &  &  
& occurs\_in & 22 & 0.37 \\

&  &  
& finding\_site\_of & 22 & 0.37 \\
 &  &  
& has\_clinical\_course & 21 & 0.35 \\

&  &  
& possibly\_equivalent\_to & 20 & 0.34 \\

 &  &  
& occurs\_before & 19 & 0.32 \\

\bottomrule
\end{tabular}%
}
\end{table*}


\subsection{Context Augmentation}
\label{app:context_graph_details}



Each candidate context triple is encoded with SapBERT and compared against the primary triple using cosine similarity. Candidate triples with similarity greater than or equal to a threshold $\theta$ are retained as supporting CG context. 
We performed a preliminary threshold-selection experiment using the existing CG-grounded 1-hop QA validation dataset and the Base model. For each candidate threshold, we retrieved the corresponding context triples and modified only the inference prompt so that the model could use the retrieved context when answering.





Table~\ref{tab:appendix_context_threshold_selection} reports the preliminary threshold-selection sweep used to choose the SapBERT similarity threshold $\theta$ for each disease setting. Table~\ref{tab:appendix_context_summary} summarizes the final CG context-selection statistics after applying the selected thresholds.
\begin{table}[H]
\centering
\caption{Preliminary threshold-selection results for CG context retrieval. For each threshold $\theta$, context triples are retrieved using SapBERT similarity to the primary triple. The Base model is evaluated on the existing CG-grounded 1-hop QA dataset using a context-aware inference prompt.}
\label{tab:appendix_context_threshold_selection}
\small
\resizebox{\textwidth}{!}{%
\begin{tabular}{ccc|ccc}
\toprule
\multicolumn{3}{c|}{\textbf{Gastroparesis CG}} 
& \multicolumn{3}{c}{\textbf{Diabetes CG}} \\
\cmidrule(lr){1-3}
\cmidrule(lr){4-6}
\textbf{$\theta$} 
& \textbf{Mean context triples / row} 
& \textbf{1-hop accuracy (\%)} 
& \textbf{$\theta$} 
& \textbf{Mean context triples / row} 
& \textbf{1-hop accuracy (\%)} \\
\midrule
0.40 & 3.32 & 85.8 & 0.40 & 3.77 & 88.8 \\
0.45 & 3.21 & 85.4 & 0.45 & 3.15 & 87.2 \\
0.50 & 2.89 & 86.7 & 0.50 & 3.02 & 88.6 \\
0.55 & 2.51 & 88.1 & 0.55 & 2.97 & 90.5 \\
0.60 & 2.24 & 88.6 & \textbf{0.60} & \textbf{2.88} & \textbf{93.2} \\
\textbf{0.65} & \textbf{1.97} & \textbf{92.2} & 0.65 & 2.34 & 90.2 \\
0.70 & 1.81 & 90.3 & 0.70 & 2.19 & 90.3 \\
0.75 & 1.60 & 87.5 & 0.75 & 2.08 & 89.0 \\
0.80 & 1.25 & 87.5 & 0.80 & 1.83 & 87.6 \\
\bottomrule
\end{tabular}%
}
\end{table}

Based on this preliminary sweep, we select $\theta=0.65$ for Gastroparesis and $\theta=0.60$ for Diabetes. These thresholds give the highest 1-hop accuracy in the corresponding CG-grounded setting while retaining a moderate number of context triples per primary triple. 

\begin{table}[h]
\centering
\caption{Final CG context-selection summary for the two disease settings.}
\label{tab:appendix_context_summary}
\small
\begin{tabular}{lccccc}
\toprule
\textbf{Setting} 
& \textbf{Primaries} 
& \textbf{$\theta$} 
& \textbf{Mean ctx. before filter} 
& \textbf{Mean ctx. after filter} 
& \textbf{Empty rows} \\
\midrule
Gastroparesis CG & 6,018 & 0.65 & 3.61 & 1.97 & 677 \\
Diabetes CG      & 5,954 & 0.60 & 3.44 & 2.88 & 968 \\
\bottomrule
\end{tabular}
\end{table}

A few representative examples illustrating how primary triples and their supporting context triples are selected from the same source text chunk after applying the similarity threshold $\theta$ are shown below.

\begin{tcolorbox}[
breakable,
colback=green!3,
colframe=green!45!black,
title={Example of Primary and Context Triple Construction: Gastroparesis CG (Source ID: 8596)},
fonttitle=\bfseries,
arc=2mm,
boxrule=0.6pt
]

\textbf{Source text chunk:}

\begin{tcolorbox}[
    colback=blue!4,
    colframe=blue!40,
    arc=1.5mm,
    boxrule=0.4pt
]
High HbA1c, increased fasting blood glucose, \colorbox{yellow!25}{polyneuropathy}, cigarette smoking, and history of comorbid conditions were observed. \colorbox{yellow!25}{Obesity} and female gender were predictors of at least one cardinal \colorbox{yellow!25}{gastroparesis} symptom. Conclusions: gastric emptying is significant in the pathogenesis of gastroparesis-related symptoms. Disease duration of more than 10 years, poor glycemic control with \colorbox{yellow!25}{hyperglycemia}, high HbA1c, \colorbox{yellow!25}{polyneuropathy}, and cigarette smoking must be considered as predictors for early detection and risk factors for the advancement of \colorbox{yellow!25}{gastroparesis} in T2DM.
\end{tcolorbox}

\vspace{1mm}
\textbf{Primary triple:}

\begin{tcolorbox}[
    colback=red!4,
    colframe=red!50!black,
    arc=1.5mm,
    boxrule=0.4pt
]
\[
\tau_p = \triple{hyperglycemia}{cause\_of}{gastroparesis}
\]
\end{tcolorbox}

\vspace{1mm}
\textbf{Context triples extracted from the same source chunk after $\theta=0.65$ filtering:}

\begin{tcolorbox}[
    colback=green!4,
    colframe=green!45!black,
    arc=1.5mm,
    boxrule=0.4pt
]
\[
\mathcal{C}(\tau_p) =
\left\{
\begin{array}{l}
\triple{obesity}{cause\_of}{gastroparesis},\\[2pt]
\triple{polyneuropathy}{cause\_of}{gastroparesis}
\end{array}
\right\}
\]
\end{tcolorbox}

\end{tcolorbox}

\begin{tcolorbox}[
breakable,
colback=green!3,
colframe=green!45!black,
title={Example of Primary and Context Triple Construction: Diabetes CG (Source ID: 34399)},
fonttitle=\bfseries,
arc=2mm,
boxrule=0.6pt
]

\textbf{Source text chunk:}

\begin{tcolorbox}[
    colback=blue!4,
    colframe=blue!40,
    arc=1.5mm,
    boxrule=0.4pt
]
There is a two-way relationship between \colorbox{yellow!25}{diabetes mellitus} and 
\colorbox{yellow!25}{periodontitis}. \colorbox{yellow!25}{Diabetes mellitus} represents an established risk factor for 
\colorbox{yellow!25}{chronic periodontitis}. Conversely, \colorbox{yellow!25}{chronic periodontitis} adversely modulates serum glucose levels in 
\colorbox{yellow!25}{diabetic patients}. Activated immune and inflammatory responses are noted during 
\colorbox{yellow!25}{diabetes} and \colorbox{yellow!25}{periodontitis}, under the modulation of similar biological mediators. These activated responses result in increased activity of certain immune-inflammatory mediators, including adipokines and microRNAs, in diabetic patients with 
\colorbox{yellow!25}{periodontal disease}. Notably, certain 
\colorbox{yellow!25}{microbes in the oral cavity} were identified to be involved in the occurrence of diabetes and periodontitis.
\end{tcolorbox}

\vspace{1mm}
\textbf{Primary triple:}

\begin{tcolorbox}[
    colback=red!4,
    colframe=red!50!black,
    arc=1.5mm,
    boxrule=0.4pt
]
\[
\tau_p = \triple{periodontitis}{associated\_with}{diabetes mellitus}
\]
\end{tcolorbox}

\vspace{1mm}
\textbf{Context triples extracted from the same source chunk after $\theta=0.60$ filtering:}

\begin{tcolorbox}[
    colback=green!4,
    colframe=green!45!black,
    arc=1.5mm,
    boxrule=0.4pt
]
\[
\mathcal{C}(\tau_p) =
\left\{
\begin{array}{l}
\triple{diabetes mellitus}{cause\_of}{periodontitis},\\[2pt]
\triple{periodontitis}{associated\_with}{periodontal disease},\\[2pt]
\triple{microbes in the oral cavity}{causative\_agent\_of}{periodontitis}
\end{array}
\right\}
\]
\end{tcolorbox}

\end{tcolorbox}

\paragraph{Context Graph Storage and Access:}
\label{app:cg_storage_access}

The CG is stored as a JSON array of triple records. Each record corresponds to one primary KG triple and contains the source identifier, source text chunk, head concept, relation, tail concept, supporting context triples, and their similarity scores. Thus, the primary triple defines the main KG edge, and the supporting triples store the local evidence associated with that edge. An example record is shown below:

\begin{verbatim}
{
  "id": "PMC125_chunk_004",
  "text": "Gastroparesis is associated with delayed gastric emptying ...",
  "head": "gastroparesis",
  "relation": "associated_with",
  "tail": "delayed gastric emptying",
  "context_triples": [
    ["gastroparesis", "associated_with", "nausea"],
    ["gastroparesis", "associated_with", "vomiting"],
    ["gastroparesis", "associated_with", "early satiety"]
  ],
  "context_similarities": [0.86, 0.82, 0.78]
}
\end{verbatim}

This JSON representation is used directly during CG-grounded QA generation. For each primary triple or reasoning path, the QA generator retrieves the corresponding supporting triples from the \texttt{context\_triples} field and uses them as local evidence. Therefore, training and evaluation stages that only require generated QA items can operate directly from the JSON exports.

When graph traversal is required, such as for multi-hop path sampling, path enumeration, or curriculum generation, the JSON representation is materialized as a NetworkX \texttt{MultiDiGraph}. Each unique concept is assigned an integer node ID, each unique relation is assigned an integer relation ID, and each primary triple is added as a directed edge from the head concept to the tail concept. The relation ID is stored as an edge attribute. The materialized graph is saved with companion mapping files: \texttt{vocab.txt} maps node IDs to concept names, \texttt{relations.json} maps relation IDs to relation names, and \texttt{custom\_cg.graph} stores the pickled NetworkX graph.

At runtime, the graph is loaded from \texttt{custom\_cg.graph}, and node and relation names are resolved using \texttt{vocab.txt} and \texttt{relations.json}. The graph is then traversed to sample $k$-hop paths. After a path is selected, the corresponding JSON records are used to recover the supporting context triples for CG-grounded QA generation.


\subsection{QA Item Generation Prompts and Filtering Details}
\label{app:QA_items_generation}

This section provides the implementation details of the QA item generation pipeline used in Section~\ref{sec:qa_generation}. All four settings, Gastroparesis-KG, Gastroparesis-CG, Diabetes-KG, and Diabetes-CG, follow the same four-stage pipeline summarized in Table~\ref{tab:appendix_qa_pipeline}.

\begin{table}[h]
\centering
\caption{Four-stage QA item generation pipeline used for all disease and grounding settings.}
\label{tab:appendix_qa_pipeline}
\small
\begin{tabular}{p{0.08\linewidth}p{0.22\linewidth}p{0.22\linewidth}p{0.38\linewidth}}
\toprule
\textbf{Step} & \textbf{Stage} & \textbf{Model} & \textbf{Purpose} \\
\midrule
1 & MCQ generation & Gemini-2.5-Flash & Generate a clinical vignette, four answer options, and the correct option letter. KG and CG differ at this step. \\
2 & Quality filtering & Qwen3-1.7B-FP8 & Reject questions with near-duplicate options or formatting problems. \\
3 & Explanation generation & Gemini-2.5-Pro & Generate a detailed explanation grounded in the primary triple or path. \\
4 & Correctness verification & Qwen3-32B/Qwen2.5-72B & Verify that the question, answer, and explanation are correct and supported by the primary source triple or path. \\
\bottomrule
\end{tabular}
\end{table}

\paragraph{Final QA item format.}
After all four stages, each accepted QA item is stored using the following format:

\begin{tcolorbox}[
    breakable,
    colback=cyan!4,
    colframe=black!50,
    title={Final QA Item Format},
    fonttitle=\bfseries,
    arc=1.5mm,
    boxrule=0.5pt
]
\small
\begin{verbatim}
<Question>
[Clinical vignette]
</Question>

<Options>
A. [Option]
B. [Option]
C. [Option]
D. [Option]
</Options>

<Explanation>
[Generated explanation/thinking trace]
</Explanation>

<Answer>
[Correct option letter]
</Answer>
\end{verbatim}
\normalsize
\end{tcolorbox}

\subsubsection{Step 1: MCQ Generation}

The first stage generates the initial multiple-choice clinical QA item. The KG-grounded prompt receives only the target triple or multi-hop path. The CG-grounded prompt receives the same target triple or path together with supporting context triples extracted from the same source text chunk. In both settings, the keyed answer must be supported by the primary triple or path.

\paragraph{KG-grounded one-hop prompt.}

\begin{tcolorbox}[
    breakable,
    colback=blue!3,
    colframe=blue!50!black,
    title={KG-grounded One-Hop MCQ Generation Prompt},
    fonttitle=\bfseries,
    arc=1.5mm,
    boxrule=0.5pt
]
\small
You are a biomedical question writer with expertise in the \texttt{\{disease\}} domain.

Create a medical examination question, similar to those found in medical board exams, for advanced medical students. The question should test the relationship between \texttt{\{source\_concept\}} and \texttt{\{target\_concept\}}.

The relationship is:

\texttt{\{paths\_str\}}

The question should:
\begin{enumerate}
    \item Be in multiple-choice format with four options.
    \item Require clinical reasoning along the given relationship.
    \item Include a brief clinical vignette.
    \item Not directly mention or quote the relation string in the question stem.
    \item Have one clearly correct answer.
\end{enumerate}

Use the following format:

\begin{verbatim}
[Clinical Vignette]

A. [Option]
B. [Option]
C. [Option]
D. [Option]

[Correct Option Letter]
\end{verbatim}

\normalsize
\end{tcolorbox}

\paragraph{Example fill.}
For a one-hop KG-grounded QA item, the prompt variables may be filled as follows:
\begin{itemize}
    \item \texttt{disease = Diabetes}
    \item \texttt{source\_concept = diabetes mellitus}
    \item \texttt{target\_concept = gastroparesis}
    \item \texttt{paths\_str = (diabetes mellitus , cause\_of , gastroparesis)}
\end{itemize}

\paragraph{CG-grounded one-hop prompt.}
\vspace*{1mm}
\begin{tcolorbox}[
    breakable,
    colback=green!3,
    colframe=green!45!black,
    title={CG-grounded One-Hop MCQ Generation Prompt},
    fonttitle=\bfseries,
    arc=1.5mm,
    boxrule=0.5pt
]
\small
You are a biomedical question writer with expertise in the \texttt{\{disease\}} domain.

Create a medical examination question, similar to those found in medical board exams, for advanced medical students.

Primary relationship to assess:

\texttt{\{primary\}}

The keyed answer must follow the primary relationship above.

Additional triples from the same literature chunk:

\texttt{\{ctx\}}

These additional triples may be used to build a richer clinical vignette or more plausible distractors. However, the correct answer must still be justified by the primary relationship.

The question should:
\begin{enumerate}
    \item Be in multiple-choice format with four options.
    \item Require clinical reasoning related to the primary relationship.
    \item Include a brief clinical vignette.
    \item Not directly mention or quote the primary relation string in the question stem.
    \item Have one clearly correct answer.
\end{enumerate}

Use the following format:

\begin{verbatim}
[Clinical Vignette]

A. [Option]
B. [Option]
C. [Option]
D. [Option]

[Correct Option Letter]
\end{verbatim}

\normalsize
\end{tcolorbox}

\paragraph{Example fill.}
For a one-hop CG-grounded QA item, the prompt variables may be filled as follows:
\begin{itemize}
    \item \texttt{disease = Gastroparesis}
    \item \texttt{primary = (vomiting , associated\_finding\_of , delayed gastric emptying)}
    \item \texttt{ctx =}
    \begin{itemize}
        \item \texttt{(type 2 diabetes , associated\_with , gastroparesis)}
        \item \texttt{(gastroparesis , associated\_finding\_of , nausea)}
        \item \texttt{(gastroparesis , associated\_finding\_of , vomiting)}
        \item \texttt{(delayed gastric emptying , affects\_site , stomach)}
    \end{itemize}
\end{itemize}

\paragraph{KG-grounded multi-hop prompt.}
\vspace*{1mm}
\begin{tcolorbox}[
    breakable,
    colback=blue!3,
    colframe=blue!50!black,
    title={KG-grounded Multi-Hop MCQ Generation Prompt},
    fonttitle=\bfseries,
    arc=1.5mm,
    boxrule=0.5pt
]
\small
You are a biomedical question writer with expertise in the \texttt{\{disease\}} domain.

Create a medical examination question, similar to those found in medical board exams, for advanced medical students.

Multi-hop path to assess:

\texttt{\{path\}}

The keyed answer must follow the full chain above.

The question should:
\begin{enumerate}
    \item Be in multiple-choice format with four options.
    \item Require clinical reasoning along all relations in the path.
    \item Require the intermediate concept or concepts to arrive at the correct answer.
    \item Not be answerable using only a direct or single-hop relationship between the source and target concepts.
    \item Include a brief clinical vignette.
    \item Not directly mention or quote the relation strings in the question stem.
    \item Have one clearly correct answer.
\end{enumerate}

Use the following format:

\begin{verbatim}
[Clinical Vignette]

A. [Option]
B. [Option]
C. [Option]
D. [Option]

[Correct Option Letter]
\end{verbatim}

\normalsize
\end{tcolorbox}

\paragraph{Example fill.}
For a multi-hop KG-grounded QA item, the prompt variables may be filled as follows:
\begin{itemize}
    \item \texttt{disease = Gastroparesis}
    \item \texttt{path =}
    \begin{itemize}
        \item \texttt{(type 2 diabetes , associated\_with , gastroparesis)}
        \item \texttt{(gastroparesis , associated\_finding\_of , vomiting)}
        \item \texttt{(vomiting , associated\_finding\_of , delayed gastric emptying)}
    \end{itemize}
\end{itemize}

\paragraph{CG-grounded multi-hop prompt.}
\vspace*{1mm}
\begin{tcolorbox}[
    breakable,
    colback=green!3,
    colframe=green!45!black,
    title={CG-grounded Multi-Hop MCQ Generation Prompt},
    fonttitle=\bfseries,
    arc=1.5mm,
    boxrule=0.5pt
]
\small
You are a biomedical question writer with expertise in the \texttt{\{disease\}} domain.

Create a medical examination question, similar to those found in medical board exams, for advanced medical students.

Multi-hop path to assess:

\texttt{\{path\}}

The keyed answer must follow the full chain above.

Additional triples from the same literature chunks:

\texttt{\{ctx\}}

These additional triples may be used to build a richer clinical vignette or more plausible distractors. However, the correct answer must still be justified by the primary multi-hop path.

The question should:
\begin{enumerate}
    \item Be in multiple-choice format with four options.
    \item Require clinical reasoning along all relations in the path.
    \item Require the intermediate concept or concepts to arrive at the correct answer.
    \item Not be answerable using only a direct or single-hop relationship between the source and target concepts.
    \item Include a brief clinical vignette.
    \item Not directly mention or quote the relation strings in the question stem.
    \item Have one clearly correct answer.
\end{enumerate}

Use the following format:

\begin{verbatim}
[Clinical Vignette]

A. [Option]
B. [Option]
C. [Option]
D. [Option]

[Correct Option Letter]
\end{verbatim}

\normalsize
\end{tcolorbox}

\paragraph{Example fill.}
For a multi-hop CG-grounded QA item, the prompt variables may be filled as follows:
\begin{itemize}
    \item \texttt{disease = Gastroparesis}
    \item \texttt{path =}
    \begin{itemize}
        \item \texttt{(type 2 diabetes , associated\_with , gastroparesis)}
        \item \texttt{(gastroparesis , associated\_finding\_of , vomiting)}
        \item \texttt{(vomiting , associated\_finding\_of , delayed gastric emptying)}
    \end{itemize}
    \item \texttt{ctx =}
    \begin{itemize}
        \item \texttt{(gastroparesis , associated\_finding\_of , nausea)}
        \item \texttt{(delayed gastric emptying , affects\_site , stomach)}
    \end{itemize}
\end{itemize}

\subsubsection{Step 2: Quality Filtering and Answer-Option Uniformization}

After MCQ generation, each question is checked for option quality. The goal of this step is to reject questions whose answer options are near-duplicates of each other.

\begin{tcolorbox}[
    breakable,
    colback=orange!6,
    colframe=orange!70!black,
    title={Option Quality Filtering Prompt},
    fonttitle=\bfseries,
    arc=1.5mm,
    boxrule=0.5pt
]
\small
You will be given a question. Check whether the answer options are near-duplicates of each other.

Only respond with \texttt{Yes} or \texttt{No}.

Respond \texttt{Yes} if the options are not near-duplicates of each other. Respond \texttt{No} otherwise.

Question:

\texttt{\{question\}}
\normalsize
\end{tcolorbox}

Questions that pass this quality filter are then answer-option uniformized before being passed to the remaining stages of the pipeline. Specifically, the correct answer positions are approximately balanced across A, B, C, and D so that the model cannot exploit answer-position frequency as a shortcut. Although the target distribution is 25\% for each option after uniformization, later filtering stages may remove some items. Therefore, the final accepted datasets remain close to, but not always exactly at, 25\% per option.

\subsubsection{Step 3: Explanation Generation}

For each QA item that passes the option-quality filter, we generate an explanation using the primary triple or primary multi-hop path as the answer-bearing source. In CG-grounded settings, the supporting context triples are also provided so that the explanation can use the local biomedical evidence around the primary path. However, the final answer must still be justified by the primary triple or path.

\begin{tcolorbox}[
    breakable,
    colback=purple!4,
    colframe=purple!60!black,
    title={Explanation Generation Prompt},
    fonttitle=\bfseries,
    arc=1.5mm,
    boxrule=0.5pt
]
\small
Generate a detailed explanation for the answer to the following question:

\texttt{\{question\}}

Primary source triple or path:

\texttt{\{paths\_str\}}

Supporting context triples, if provided:

\texttt{\{ctx\}}

The explanation should:
\begin{enumerate}
    \item Include the steps leading to the correct answer.
    \item Use the primary source triple or path to justify the final answer.
    \item If supporting context triples are provided, use them to enrich the reasoning and connect the relevant biomedical concepts.
    \item Not mention that a source, path, or context was provided.
    \item Sound like a medical student explaining the answer to a peer.
\end{enumerate}
\normalsize
\end{tcolorbox}

For KG-grounded QA items, \texttt{\{ctx\}} is set to \texttt{None provided.}; for CG-grounded QA items, \texttt{\{ctx\}} contains the supporting context triples attached to the primary triple or path.
\subsubsection{Step 4: Triple-Grounded Correctness Verification}

The final stage verifies whether the generated question, answer, and explanation are medically correct and supported by the primary source triple or path. This step is applied after explanation generation. The correctness verifier receives the complete QA item and the primary source triple or path. 
\begin{tcolorbox}[
    breakable,
    colback=red!3,
    colframe=red!55!black,
    title={Triple-Grounded Correctness Verification Prompt},
    fonttitle=\bfseries,
    arc=1.5mm,
    boxrule=0.5pt
]
\small
You are a medical examiner. You are given a medical question, an explanation, and an answer. The question and answer are formatted as follows:

\begin{verbatim}
<Question>
[Clinical Vignette]

A. [Option]
B. [Option]
C. [Option]
D. [Option]
</Question>

<Explanation>
[Explanation]
</Explanation>

<Answer>
[Correct Option Letter]
</Answer>
\end{verbatim}

Judge whether the question and answer are logically correct, medically accurate, and supported by the source. If an explanation is provided, also judge whether the explanation is correct and supported by the source.

Respond using only the following format:

\begin{verbatim}
Correct: [Yes/No]
\end{verbatim}

Question and Answer:

\texttt{\{question\_answer\_explanation\}}

Source:

\texttt{\{paths\_str\}}
\normalsize
\end{tcolorbox}

\paragraph{Representative full generated QA item.}
The generated QA items include long thinking traces; hence, we show one complete representative example from the Gastroparesis CG-grounded setting. The full generated QA records for all settings are provided as JSON files in the project code repository. 

\begin{tcolorbox}[
breakable,
colback=green!3,
colframe=green!45!black,
title={Representative Full QA Item: Gastroparesis CG-grounded},
fonttitle=\bfseries,
arc=1.5mm,
boxrule=0.5pt,
left=1mm,
right=1mm
]

\begin{lstlisting}[style=qaexample]
Primary:
(delayed gastric emptying , associated_finding_of , gastroparesis)

CG context:
1. (delayed gastric emptying , has_finding_site , stomach)
2. (gastroparesis , has_finding_site , stomach)
3. (symptomatic patients , focus_of , gastroparesis)
4. (gastroparesis , has_associated_morphology , autonomic neuropathy)

<Question>
A 58-year-old male with a 20-year history of type 2 diabetes mellitus presents with chronic nausea, early satiety, and occasional postprandial vomiting of undigested food particles, which has worsened over the past six months. He reports significant weight loss. Endoscopy reveals no mechanical obstruction or mucosal abnormalities in the stomach or duodenum. His hemoglobin A1c is 9.2%. Based on patient's clinical presentation and initial workup, which of the following findings would be most critical in establishing the definitive diagnosis?
</Question>

<Options>
A. Markedly reduced gastric emptying demonstrated by scintigraphy
B. Evidence of vagal nerve damage on nerve conduction studies
C. Presence of chronic intractable nausea and vomiting
D. Normal findings on upper gastrointestinal endoscopy
</Options>

<Explanation>
### Step-by-step Thinking Proces

### Step 1: Deconstruct the Clinical Vignette

- Patient Profile: 58-year-old male.
- Key History: 20-year history of type 2 diabetes mellitus (T2DM). This is a very long duration, putting him at high risk for chronic complications.
- Key Labs: Hemoglobin A1c is 9.2%. This indicates very poorly controlled diabetes over the last 2-3 months, which accelerates and worsens diabetic complications, especially neuropathy.
- Presenting Symptoms: Chronic nausea, early satiety, postprandial vomiting of undigested food, and significant weight loss. These are the hallmark symptoms of a gastric motility disorder. The fact that the food is undigested suggests it is sitting in the stomach for a long time and not moving into the small intestine for digestion.
- Initial Workup: Upper endoscopy is normal. This is a crucial piece of information. It rules out a mechanical obstruction, such as a tumor, stricture, or severe peptic ulcer disease, as the cause of his symptoms. The problem is functional, not structural.
- The Question: What is the most critical finding to establish a definitive diagnosis? This means we are looking for the key diagnostic test that confirms the suspected condition.

### Step 2: Formulate a Differential Diagnosis and a Working Diagnosis

- Based on the symptoms, including nausea, vomiting, and early satiety, the long-standing poorly controlled diabetes, and the normal endoscopy that rules out obstruction, my leading diagnosis is diabetic gastroparesis.
- Gastroparesis literally means "stomach paralysis." It is a condition where the stomach cannot empty itself of food in a normal fashion due to damage to the nerves that control stomach motility, primarily the vagus nerve. The most common cause of gastroparesis is diabetes.
- So, the question is really asking, "How do you definitively diagnose gastroparesis?"

### Step 3: Analyze the Options in the Context of the Working Diagnosis

- Option A: Markedly reduced gastric emptying demonstrated by scintigraphy

  What is it? A gastric emptying scintigraphy study is the gold standard diagnostic test for gastroparesis. The patient eats a standardized meal containing a small amount of a radioactive tracer. A scanner then takes pictures over several hours to see how much of the meal is left in the stomach.

  Relevance: The very definition of gastroparesis is delayed gastric emptying in the absence of mechanical obstruction. This test directly and objectively measures that function. Finding that the stomach has retained too much food after a set amount of time confirms the pathophysiological basis of patient's symptoms. This is a very strong candidate for a definitive finding.

- Option B: Evidence of vagal nerve damage on nerve conduction studies

  What is it? Nerve conduction studies test how well electrical signals travel along a nerve.

  Relevance: In this patient, diabetic autonomic neuropathy affecting the vagus nerve is almost certainly the cause of his gastroparesis. However, diagnosing the underlying cause is not the same as diagnosing the condition itself. Furthermore, standard nerve conduction studies are not typically used to directly assess the vagus nerve's control of the stomach. While this finding would support the etiology, it does not directly measure gastric function and is not the primary diagnostic tool for gastroparesis. The diagnosis is based on the stomach's performance, not the nerve's condition.

- Option C: Presence of chronic intractable nausea and vomiting

  What is it? These are the patient's symptoms.

  Relevance: These symptoms are what prompted the investigation. They are required to suspect the diagnosis, but they are not definitive. Many other conditions can cause chronic nausea and vomiting, such as functional dyspepsia, cyclic vomiting syndrome, or CNS issues. Symptoms alone are subjective and not specific enough for a definitive diagnosis.

- Option D: Normal findings on upper gastrointestinal endoscopy

  What is it? This is the result of the initial workup.

  Relevance: This finding is critically important, but its role is to exclude other diagnoses. By showing there is no mechanical obstruction, it allows us to consider a motility disorder like gastroparesis. It tells us what the patient does not have, but it does not positively confirm what the patient does have. Therefore, it is a necessary step in the workup but not the definitive diagnostic finding itself.

### Step 4: Synthesize and Select the Best Answer

1. The patient's clinical picture strongly suggests gastroparesis.
2. The diagnosis of gastroparesis requires two key components: characteristic symptoms and objective confirmation of delayed gastric emptying after mechanical obstruction has been ruled out.
3. We already have the symptoms and have ruled out obstruction.
4. Therefore, the final and most critical step to make the diagnosis definitive is to prove that gastric emptying is delayed.
5. Option A, demonstrating delayed gastric emptying with scintigraphy, is the direct test for this. The core pathophysiological problem in gastroparesis is delayed gastric emptying, and this test confirms it.

### Final Conclusion

The most critical finding to establish the definitive diagnosis of gastroparesis is the objective demonstration of delayed gastric emptying. Gastric emptying scintigraphy is the standard method for this. Therefore, Option A is the correct answer.

### Explanation

Alright, let's walk through this case. We have a male patient with a long, poorly controlled history of type 2 diabetes, and he is presenting with classic symptoms: nausea, feeling full quickly, and vomiting food he ate hours ago. This clinical picture should immediately make you think of gastroparesis, especially in a diabetic patient.

First, the workup. They did an endoscopy and found nothing. That is a very important step because you have to rule out a physical blockage. If he had a tumor or a severe ulcer causing gastric outlet obstruction, the treatment would be completely different. So, the normal endoscopy tells us this is likely a functional problem, not a mechanical one. This makes our suspicion for gastroparesis even higher.

Now let's look at the options to see what would confirm the diagnosis.

- Option C, symptoms, and Option D, normal endoscopy, are essential parts of the puzzle, but they are not the final confirmation. The symptoms are what bring the patient in, and the normal endoscopy rules out other causes. They set the stage for the diagnosis but do not seal the deal.

- Option B, vagal nerve damage, is getting at the cause. In a patient with diabetic gastroparesis, the cause is autonomic neuropathy damaging the vagus nerve, which controls stomach motility. But we do not diagnose gastroparesis by testing the nerve directly; we diagnose it by looking at the effect on the stomach.

- This leaves us with Option A. The very definition of gastroparesis is delayed gastric emptying in the absence of a mechanical obstruction. To make a definitive diagnosis, you have to prove that the stomach is actually emptying too slowly. The gold standard for this is a gastric emptying scintigraphy study. This test objectively measures how long it takes for food to leave the stomach. If that study shows markedly reduced or delayed gastric emptying, that is the key finding that establishes the diagnosis of gastroparesis.

So, to sum it up: The symptoms make you suspect it, the normal endoscopy rules out other major causes, but the definitive proof, the most critical finding, is demonstrating that delayed gastric emptying is happening, which is exactly what scintigraphy does.
</Explanation>

<Answer>
A
</Answer>
\end{lstlisting}

\end{tcolorbox}

\subsubsection{Answer-Option Distribution After Filtering}

Table~\ref{tab:appendix_sft_option_distribution} shows the answer-option distribution for the final one-hop SFT training datasets after Step 4 filtering. The distributions remain close to 25\% for each option, indicating that answer-option uniformization is largely preserved after filtering.

\begin{table}[h]
\centering
\caption{Answer-option distribution for final one-hop SFT training datasets after Step 4 filtering.}
\label{tab:appendix_sft_option_distribution}
\small
\begin{tabular}{lccccc}
\toprule
\textbf{Setting} & \textbf{Count} & \textbf{A (\%)} & \textbf{B (\%)} & \textbf{C (\%)} & \textbf{D (\%)} \\
\midrule
Diabetes KG & 5,954 & 24.9 & 25.1 & 25.2 & 24.8 \\
Diabetes CG & 5,954 & 25.1 & 25.0 & 25.0 & 24.9 \\
Gastroparesis KG & 6,018 & 24.9 & 25.0 & 25.1 & 25.0 \\
Gastroparesis CG & 6,018 & 24.9 & 25.1 & 25.0 & 25.0 \\
\bottomrule
\end{tabular}
\end{table}

Table~\ref{tab:appendix_rl_option_distribution} reports the answer-option distribution for the mixed lower-hop RL datasets. Most settings remain close to uniform. 

\begin{table}[h]
\centering
\caption{Answer-option distribution for mixed lower-hop RL datasets.}
\label{tab:appendix_rl_option_distribution}
\small
\begin{tabular}{lccccc}
\toprule
\textbf{Setting} & \textbf{Count} & \textbf{A (\%)} & \textbf{B (\%)} & \textbf{C (\%)} & \textbf{D (\%)} \\
\midrule
Diabetes KG & 4,248 & 24.8 & 25.2 & 24.4 & 25.6 \\
Diabetes CG & 3,940 & 24.9 & 24.7 & 25.2 & 25.2 \\
Gastroparesis KG & 4,270 & 25.1 & 25.4 & 24. & 24.3 \\
Gastroparesis CG & 4,121 & 24.6 & 25.4 & 24.7 & 25.3 \\
\bottomrule
\end{tabular}
\end{table}

Table~\ref{tab:appendix_eval_option_distribution} shows the answer-option distribution for the higher-hop evaluation datasets. Each disease and grounding regime contains 8,000 QA items per hop depth. The distributions are close to uniform across A, B, C, and D.

\begin{table*}[h]
\centering
\caption{Answer-option distribution for higher-hop evaluation datasets. Each row contains 8,000 QA items. Percentages are rounded to one decimal place and sum to 100.0 within each row.}
\label{tab:appendix_eval_option_distribution}
\small
\begin{tabular}{llcccc}
\toprule
\textbf{Setting} & \textbf{Hop} & \textbf{A (\%)} & \textbf{B (\%)} & \textbf{C (\%)} & \textbf{D (\%)} \\
\midrule
Diabetes KG & 3-hop & 26.4 & 25.5 & 25.2 & 22.9 \\
Diabetes KG & 4-hop & 26.5 & 25.7 & 24.8 & 23.0 \\
Diabetes KG & 5-hop & 26.7 & 25.3 & 25.0 & 23.0 \\
\midrule
Diabetes CG & 3-hop & 26.3 & 24.5 & 25.1 & 24.1 \\
Diabetes CG & 4-hop & 26.7 & 24.4 & 25.3 & 23.6 \\
Diabetes CG & 5-hop & 26.6 & 24.3 & 25.1 & 24.0 \\
\midrule
Gastroparesis KG & 3-hop & 25.0 & 24.6 & 25.8 & 24.6 \\
Gastroparesis KG & 4-hop & 24.4 & 24.7 & 25.7 & 25.2 \\
Gastroparesis KG & 5-hop & 24.4 & 25.1 & 25.4 & 25.1 \\
\midrule
Gastroparesis CG & 3-hop & 25.6 & 25.1 & 25.3 & 24.0 \\
Gastroparesis CG & 4-hop & 25.1 & 25.0 & 25.4 & 24.5 \\
Gastroparesis CG & 5-hop & 25.0 & 24.7 & 25.3 & 25.0 \\
\bottomrule
\end{tabular}
\end{table*}

\subsubsection{Validation of Higher-Hop Evaluation QA Items}

The higher-hop evaluation sets also undergo an additional alignment check before being used for testing. This step is important because the 3-hop, 4-hop, and 5-hop QA items are used as the final evaluation benchmark. If an evaluation question is not actually aligned with its source triple/path, then an incorrect model prediction may reflect a flawed evaluation item rather than a true model error. Such cases are similar to the \texttt{BAD\_VAL} category used during adaptive repair.
Therefore, each generated evaluation QA item is checked using the same LLM-based validation logic used in the Judge-1 audit. 
This validation step is applied independently of the models being evaluated. The purpose is only to ensure that the 8,000 QA items used for each 3-hop, 4-hop, and 5-hop evaluation setting are valid, answerable, and aligned with the intended source path. This reduces the risk that reported evaluation errors are caused by misaligned or noisy QA items rather than by limitations of the model.

\paragraph{Higher-hop evaluation alignment prompt.}
We use \texttt{Qwen3.6-27B} as an LLM judge to validate the 3-hop, 4-hop, and 5-hop evaluation QA items before testing. 

\begin{tcolorbox}[
    breakable,
    colback=orange!5,
    colframe=orange!70!black,
    title={Higher-Hop Evaluation QA Alignment Prompt},
    fonttitle=\bfseries,
    arc=1.5mm,
    boxrule=0.5pt
]
\small
You are a medical examiner. You are given a biomedical multiple-choice question, its expected answer, explanation, and the source triple or multi-hop path used to generate it.

Judge whether the question is aligned with the source.

A question is aligned only if all of the following hold:
\begin{enumerate}
    \item Answering correctly requires knowing or applying the triple.
    \item The expected correct answer is medically correct given the stem and options.
    \item The expected answer is uniquely correct.
    \item Clinical vignettes are allowed if the correct answer logically depends on the triple.

\end{enumerate}

A question is not aligned if it can be answered without the source, if the labeled answer is wrong or debatable, if multiple options are equally defensible, or if the question tests a different relationship than the source states.

For CG-grounded questions, supporting context triples may enrich the vignette or distractors, but the keyed answer must still be supported by the primary source path.

Respond using exactly this format:

\begin{verbatim}
Aligned: [Yes/No]
\end{verbatim}

Question and Answer:

\texttt{\{question\_answer\_explanation\}}

Source:

\texttt{\{paths\_str\}}

Supporting Context, if available:

\texttt{\{ctx\}}
\normalsize
\end{tcolorbox}

\subsection{SFT}
\label{app:SFT}

\begin{figure}[H]
    \centering
    \includegraphics[width=1\columnwidth]{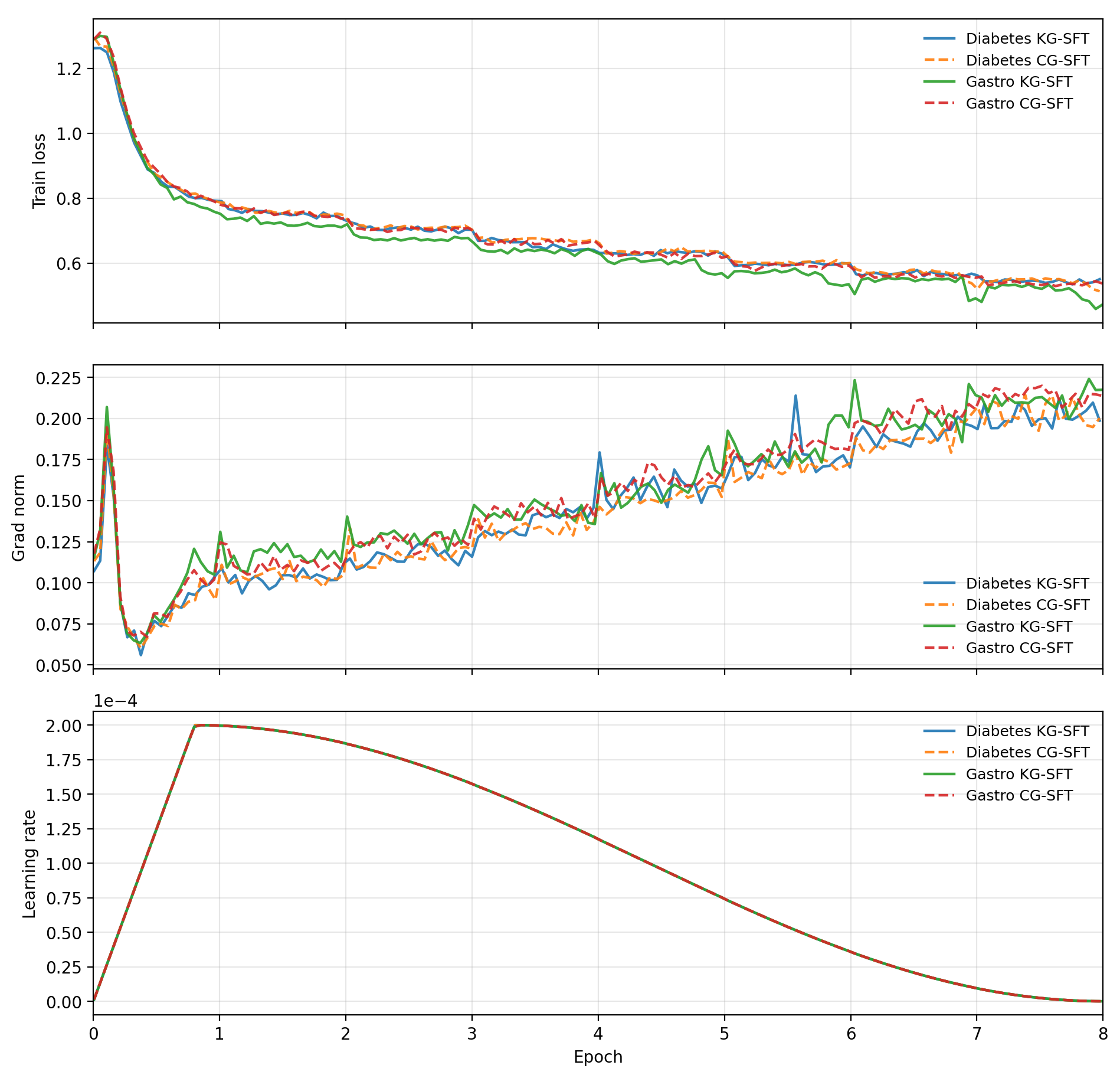}
    \caption{SFT dynamics across epochs. The plot shows the evolution of training loss, gradient norm, and learning rate during SFT.}
    \label{fig:sft_training_dynamics}
\end{figure}

Fig.~\ref{fig:sft_training_dynamics} illustrates the SFT dynamics across epochs by tracking training loss, gradient norm, and learning rate throughout the SFT process.

\FloatBarrier

We also computed the average token length of the final SFT training items using the Qwen3-14B tokenizer. Tokenization is performed on the full \texttt{question\_and\_explanation} field, which includes the question, answer options, thinking trace, and final answer. The token statistics are summarized in Table~\ref{tab:qa_token_lengths}. Overall, each QA item contains a relatively short question and option section, while most tokens are used up by the thinking trace. This reflects the fact that the training examples provide not only the final answer but also detailed relational reasoning supervision.
\begin{table}[h]
\centering
\caption{Average token length per SFT QA item using the Qwen3-14B tokenizer.}
\label{tab:qa_token_lengths}
\resizebox{\columnwidth}{!}{%
\begin{tabular}{lcccc}
\toprule
\textbf{Dataset} & \textbf{$n$} & \textbf{Mean tokens/item} & \textbf{Question+Options} & \textbf{Trace+Answer} \\
\midrule
Gastroparesis KG train & 6,018 & 1,923 & 212 & 1,710 \\
Gastroparesis CG train & 6,018 & 2,091 & 263 & 1,828 \\
Diabetes KG train & 5,954 & 2,066 & 227 & 1,839 \\
Diabetes CG train & 5,954 & 2,074 & 253 & 1,821 \\
\bottomrule
\end{tabular}%
}
\end{table}

\subsection{LLM-judged and History-Aware Adaptive Triple Repair}
\label{app:adaptive_repair_details}

Next, we discuss the repair methodology.

\subsubsection{Overview of the Repair Procedure}
\label{app:repair_overview}

Here, we provide implementation details for the adaptive triple repair procedure described in Section~\ref{sec:LLM_Judge_SFT_Repair}. The repair pipeline starts from a one-time SFT model trained on all available 1-hop QA items. The model is then evaluated on the full 1-hop validation set and every incorrect validation case is audited by Judge 1. Judge 1 separates failures into invalid triples (\texttt{INVALID\_TRIPLE}), bad validation questions (\texttt{BAD\_VAL}), and valid missing-knowledge cases (\texttt{MISSING\_KNOWLEDGE}). Invalid triples are removed, bad validation questions are regenerated and re-evaluated, and only valid missing-knowledge triples are passed to targeted repair.

After Judge 1, repair proceeds in rounds. In each round, additional QA examples are generated from the failed source triples and the model is continually fine-tuned on these targeted examples. After each repair round, the model is re-evaluated on the validation set and triple IDs (\texttt{path\_idx}) are tracked across rounds. Failed triples are assigned to history-aware categories, such as new-error, regression, persistent, or quarantine candidate. Triples that repeatedly fail repair are reviewed by Judge 2 before being removed from the retained validation set or marked as valid but unlearnable.

\subsubsection{Triple-Level Error Categories}
\label{app:repair_categories}

Each validation example is indexed by its source triple ID, denoted as \texttt{path\_idx}. We maintain the history of each \texttt{path\_idx} across repair rounds. This enables us to distinguish between triples that remain consistently wrong, triples that regress after being fixed, and triples that newly become incorrect after a repair round.

\begin{itemize}
    \item \textbf{Fixed:} previously wrong, now correct.
    \item \textbf{Missing knowledge:} Judge-1-confirmed valid triple and valid question, but the model fails.
    \item \textbf{New-error:} previously correct, becomes wrong after repair, with no prior repair/fix history.
    \item \textbf{Regression:} previously fixed or previously correct, but becomes wrong again after later SFT.
    \item \textbf{Persistent:} remains wrong after two consecutive targeted repair rounds.
    \item \textbf{Quarantine candidate:} remains wrong after three targeted repair attempts and is sent to Judge 2.
    \item \textbf{Quarantined:} removed from the retained validation set after repeated failure and Judge-2 review.
\end{itemize}

\subsubsection{Repair Example Allocation}
\label{app:repair_weights}

The number of generated repair examples depends on the failure category. We use a history-aware weighting scheme rather than assigning the same number of examples to every failed triple. Persistent triples receive the strongest repair, regression triples receive moderate repair, and new-error triples receive light repair. In all experiments, we use the following allocation:
\[
\text{persistent} = 7,\qquad
\text{regression} = 5,\qquad
\text{new-error} = 3.
\]

This weighting reflects the severity of the failure. Persistent triples have failed across repeated targeted repair rounds and, therefore, require stronger supervision. Regression triples indicate forgetting of previously correct or fixed knowledge and require moderate reinforcement. New-error triples are usually recent repair-induced failures and are, therefore, repaired lightly to avoid overfitting or creating additional regressions.

\subsubsection{Continual SFT Setup}
\label{app:continual_sft_hyperparams}

For continual SFT, we used a lower learning rate than in the initial one-time SFT stage, reducing it from $2\times10^{-4}$ to $5\times10^{-5}$. This was done because each repair round started from an already-trained LoRA adapter and fine-tuned it on a small targeted repair set. A gentler continual fine-tuning configuration helped repair missing-knowledge triples while reducing the risk of overwriting the full-coverage SFT adapter and introducing additional validation regressions.

Each continual repair round was trained for six epochs. Rather than automatically using the final epoch checkpoint, we ran inference with the checkpoint from each epoch on the currently available validation set and selected the best-performing epoch as the initialization LoRA adapter for the next repair round. This epoch-selection step was important because repair sets were small and targeted: Later epochs could continue improving the targeted triples, but they could also overfit the repair examples and introduce new validation regressions. In practice, lower epochs often gave the best trade-off in most cases. They fixed many targeted failures and produced fewer regression or new-error cases than later epochs.




\subsubsection{Models Repaired}
\label{app:models_repaired}

\begin{table}[H]
\centering
\small
\caption{Four one-hop SFT checkpoints repaired using the adaptive triple repair pipeline.}
\begin{tabular}{lll}
\hline
\textbf{Model} & \textbf{Disease graph} & \textbf{One-hop supervision type} \\
\hline
KGModel-Gastro & Gastroparesis & KG-grounded QA \\
KGModel-Diab & Diabetes & KG-grounded QA \\
CGModel-Gastro & Gastroparesis & CG-grounded QA \\
CGModel-Diab & Diabetes & CG-grounded QA \\
\hline
\end{tabular}

\label{tab:repair_models}
\end{table}

For each disease, we repair both a KG-grounded model and a CG-grounded model. 
Each model is repaired independently using the same Judge 1, history-aware repair, Judge 2, and continual SFT procedure. Table~\ref{tab:repair_models} shows the models we train continually.

\subsubsection{{Repair Trace for CGModel-Gastro}}
\label{app:repair_cg_gastro}

\paragraph{Initial Performance:}The initial base model achieved 91.81\% accuracy. After one-time SFT with CG-grounded 1-hop QA examples, the model achieved 95.96\% accuracy, leaving 243 incorrect validation cases out of 6018.

\paragraph{Judge 1 Audit:}Judge 1 categorized the 243 incorrect cases as follows:
\[
\texttt{INVALID\_TRIPLE}=16,\quad
\texttt{BAD\_VAL}=87,\quad
\texttt{MISSING\_KNOWLEDGE}=140.
\]
Invalid triples were removed from the validation set before the first repair round, 
reducing the validation set size from 6018 to 6002. Bad validation questions were regenerated and re-evaluated, and the 140 missing-knowledge triples were passed to the first targeted repair round.

\paragraph{Repair-round Table:}Table~\ref{tab:repair_trace_cg_gastro} summarizes the round-by-round repair trace.
\begin{table}[h]
\centering
\begin{threeparttable}
\small
\caption{Adaptive repair trace for CGModel-Gastro.}
\setlength{\tabcolsep}{2.5pt}
\renewcommand{\arraystretch}{1.08}
\begin{tabular}{c c c c c c c c c}
\hline
\textbf{Round} & 
\textbf{Accuracy} & 
\textbf{Fixed} & 
\makecell{\textbf{Still}\\\textbf{wrong}} & 
\textbf{Persistent} & 
\textbf{Regression} & 
\textbf{New-error} & 
\textbf{Quarantine} & 
\makecell{\textbf{Best}\\\textbf{epoch}} \\
\hline
Initial SFT & 95.96\% $(5775/6018)$ & -- & 243 & -- & -- & -- & -- & -- \\
Repair 1 & 98.03\% $(5884/6002)$ & 139/227 & 118 & -- & 88 & $30^\dagger$ & -- & 5 \\
Repair 2 & 98.12\% $(5889/6002)$ & 26/118 & 113 & 76 & 23 & 14 & -- & 1 \\
Repair 3 & 98.20\% $(5894/6002)$ & 37/113 & 108 & 8 & 17 & 18 & 65 & 2 \\
Repair 4 & 99.55\% $(5910/5937)$ & 32/43 & 27 & 4 & 15 & 5 & 3 & 3 \\
Repair 5 & 99.61\% $(5911/5934)$ & 11/24 & 23 & 9 & 6 & 5 & 3 & 1 \\
Repair 6 & 99.71\% $(5914/5931)$ & 10/20 & 17 & 1 & 9 & 0 & 7 & 1 \\
Repair 7 & 99.83\% $(5915/5925)$ & 4/11 & 10 & 6 & 2 & 1 & 1 & 1 \\
Repair 8 & 99.88\% $(5917/5924)$ & 5/9 & 7 & 0 & 3 & 1 & 3 & 1 \\
Repair 9 & 100.00\% $(5922/5922)$ & 5/5 & 0 & 0 & 0 & 0 & -- & 2 \\
\hline
\end{tabular}

\label{tab:repair_trace_cg_gastro}
\begin{tablenotes}
\footnotesize
\item[$\dagger$] At Repair 1, all 30 newly wrong triples are categorized as New-error because this is the first continual repair round and no prior repair history is available to define regression.

\end{tablenotes}
\end{threeparttable}
\end{table}

\paragraph{Final Cleaned Validation Set:}After the final repair round and Judge-2 quarantine review, the retained cleaned validation set contained 5922 validation examples. The final reported accuracy was 100.00\% on this retained cleaned validation set. This retained set was obtained from the original 6018 validation examples by removing 16 Judge-1 invalid triples and 80 Judge-2 invalid or valid-unlearnable cases.

\paragraph{Quarantine/Valid-unlearnable summary:} 

Judge 2 reviewed quarantine candidates accumulated across the repair rounds. 
\begin{table}[h]
\centering
\small
\caption{Judge 2 quarantine review summary for CGModel-Gastro.}
\begin{tabular}{l c l}
\hline
\textbf{Judge 2 label} & \textbf{Count} & \textbf{Action} \\
\hline
\texttt{INVALID\_TRIPLE} & 33 & Removed from the retained validation set \\
\texttt{VALID\_UNLEARNABLE} & 47 & Reported separately and removed from the retained validation set \\
\texttt{BAD\_VAL} & 2 & Regenerated in place and kept in the validation set \\
\hline
\end{tabular}

\label{tab:judge2_quarantine_cg_gastro}
\end{table}
Overall, as shown in Table \ref{tab:judge2_quarantine_cg_gastro}, $80$ cases were removed from the retained validation denominator: $33$ were labeled as invalid triples and $47$ were labeled as valid but currently unlearnable. The remaining two cases were labeled as bad validation questions; these were rewritten and re-evaluated, but not removed from the validation set.

Table~\ref{tab:cg_gastro_judge2_full_triples} lists the full Judge 2 quarantine outcomes.

{\small
\begin{longtable}{p{0.17\linewidth} p{0.08\linewidth} p{0.68\linewidth}}
\caption{Full Judge 2 quarantine cases for CGModel-Gastro.}
\label{tab:cg_gastro_judge2_full_triples}\\
\hline
\textbf{Judge 2 label} & \textbf{\texttt{path\_idx}} & \textbf{Triple} \\
\hline
\endfirsthead

\hline
\textbf{Judge 2 label} & \textbf{\texttt{path\_idx}} & \textbf{Triple} \\
\hline
\endhead

\hline
\endfoot

\texttt{INVALID\_TRIPLE} & 175 & (pdgfr$\alpha$ + cells) --[\texttt{isa}]$\rightarrow$ (smooth muscle cells) \\
\texttt{INVALID\_TRIPLE} & 267 & (promotility agents) --[\texttt{plays\_role}]$\rightarrow$ (endoscopic treatment) \\
\texttt{INVALID\_TRIPLE} & 600 & (glucose-dependent insulinotropic polypeptide (gip)) --[\texttt{has\_method}]$\rightarrow$ (plasma) \\
\texttt{INVALID\_TRIPLE} & 790 & (prucalopride) --[\texttt{has\_modification}]$\rightarrow$ (dose) \\
\texttt{INVALID\_TRIPLE} & 1730 & (phytobezoars) --[\texttt{has\_modification}]$\rightarrow$ (rare) \\
\texttt{INVALID\_TRIPLE} & 3943 & (azi) --[\texttt{possibly\_equivalent\_to}]$\rightarrow$ (ery) \\
\texttt{INVALID\_TRIPLE} & 3683 & (ascorbic acid) --[\texttt{plays\_role}]$\rightarrow$ (gastrointestinal functional) \\
\texttt{INVALID\_TRIPLE} & 5111 & (migration) --[\texttt{plays\_role}]$\rightarrow$ (gastric smooth muscle) \\
\texttt{INVALID\_TRIPLE} & 5180 & (cholinergic muscarinic stimulation) --[\texttt{plays\_role}]$\rightarrow$ (activation) \\
\texttt{INVALID\_TRIPLE} & 3445 & (interstitial cells of cajal) --[\texttt{has\_direct\_procedure\_site}]$\rightarrow$ (smooth tissue) \\
\texttt{INVALID\_TRIPLE} & 1410 & (full-thickness myotomy) --[\texttt{has\_finding\_site}]$\rightarrow$ (stomach wall) \\
\texttt{INVALID\_TRIPLE} & 4013 & (jejunal vip) --[\texttt{has\_associated\_morphology}]$\rightarrow$ (smooth muscle) \\
\texttt{INVALID\_TRIPLE} & 3771 & (icc decline) --[\texttt{has\_associated\_morphology}]$\rightarrow$ (muscle) \\
\texttt{INVALID\_TRIPLE} & 5865 & (iauc) --[\texttt{interprets}]$\rightarrow$ (plasma glucose / insulin ratio) \\
\texttt{INVALID\_TRIPLE} & 5638 & (glucagon) --[\texttt{is\_interpreted\_by}]$\rightarrow$ (plasma glucagon) \\
\texttt{INVALID\_TRIPLE} & 4851 & (abdominal bloating) --[\texttt{is\_interpreted\_by}]$\rightarrow$ (functional abdominal pain) \\
\texttt{INVALID\_TRIPLE} & 301 & (visceral pain unpleasantness) --[\texttt{is\_interpreted\_by}]$\rightarrow$ (gastric function) \\
\texttt{INVALID\_TRIPLE} & 1197 & (amyloid light-chain amyloidosis) --[\texttt{is\_interpreted\_by}]$\rightarrow$ (intestinal amyloid) \\
\texttt{INVALID\_TRIPLE} & 1461 & (sitagliptin) --[\texttt{cause\_of}]$\rightarrow$ (fasting plasma glucose) \\
\texttt{INVALID\_TRIPLE} & 1570 & (gastric electrical stimulation) --[\texttt{cause\_of}]$\rightarrow$ (gastric emptying) \\
\texttt{INVALID\_TRIPLE} & 264 & (postprandial hypotension (pph)) --[\texttt{cause\_of}]$\rightarrow$ (cardiovascular outcomes) \\
\texttt{INVALID\_TRIPLE} & 2664 & (trpv1) --[\texttt{cause\_of}]$\rightarrow$ (reduced mechanosensitivity) \\
\texttt{INVALID\_TRIPLE} & 3193 & (glucose (6\%)) --[\texttt{cause\_of}]$\rightarrow$ (absorption) \\
\texttt{INVALID\_TRIPLE} & 3543 & (diabetes duration) --[\texttt{cause\_of}]$\rightarrow$ (gastric disease) \\
\texttt{INVALID\_TRIPLE} & 4725 & (high-caloric / low-protein nutritional product) --[\texttt{cause\_of}]$\rightarrow$ (volume) \\
\texttt{INVALID\_TRIPLE} & 5576 & (mimbb23sg) --[\texttt{cause\_of}]$\rightarrow$ (increased expression) \\
\texttt{INVALID\_TRIPLE} & 3785 & (gastric dysrhythmias) --[\texttt{has\_method}]$\rightarrow$ (electrical activity) \\
\texttt{INVALID\_TRIPLE} & 5271 & (trpa1) --[\texttt{associated\_finding\_of}]$\rightarrow$ (nitric oxide synthase) \\
\texttt{INVALID\_TRIPLE} & 1155 & (vagal signaling) --[\texttt{associated\_finding\_of}]$\rightarrow$ (gut-brain axis) \\
\texttt{INVALID\_TRIPLE} & 1710 & (thyroxine (t4)) --[\texttt{has\_modification}]$\rightarrow$ (gastric acid) \\
\texttt{INVALID\_TRIPLE} & 1084 & (enteric purinergic inhibitory musculomotor neurotransmission) --[\texttt{focus\_of}]$\rightarrow$ (enteric nervous system function) \\
\texttt{INVALID\_TRIPLE} & 1475 & (accommodation testing) --[\texttt{focus\_of}]$\rightarrow$ (gastric function) \\
\texttt{INVALID\_TRIPLE} & 3793 & (electrical syncytium) --[\texttt{plays\_role}]$\rightarrow$ (gastrointestinal nervous system) \\

\hline
\texttt{VALID\_UNLEARNABLE} & 968 & (linagliptin) --[\texttt{isa}]$\rightarrow$ (dipeptidyl peptidase inhibitor) \\
\texttt{VALID\_UNLEARNABLE} & 2396 & (epigastric pain syndrome) --[\texttt{isa}]$\rightarrow$ (dyspepsia) \\
\texttt{VALID\_UNLEARNABLE} & 71 & (cholinergic neurons) --[\texttt{cause\_of}]$\rightarrow$ (gastric contractions) \\
\texttt{VALID\_UNLEARNABLE} & 98 & (clerodane diterpene 16-hydroxycleroda-3, 13-dien-15, 16-olide) --[\texttt{cause\_of}]$\rightarrow$ (dpp-4 inhibitory activity) \\
\texttt{VALID\_UNLEARNABLE} & 125 & (calcitonin gene-related peptide) --[\texttt{cause\_of}]$\rightarrow$ (activation of K ATP channels) \\
\texttt{VALID\_UNLEARNABLE} & 165 & (acute physical and emotional stress) --[\texttt{cause\_of}]$\rightarrow$ (intestinal effects) \\
\texttt{VALID\_UNLEARNABLE} & 617 & (gastric vagal afferent tension receptor mechanosensitivity) --[\texttt{cause\_of}]$\rightarrow$ (gastric emptying) \\
\texttt{VALID\_UNLEARNABLE} & 2113 & (postoperative refractory gastroparesis) --[\texttt{cause\_of}]$\rightarrow$ (nausea) \\
\texttt{VALID\_UNLEARNABLE} & 2350 & (glucagon like peptide-1) --[\texttt{cause\_of}]$\rightarrow$ (insulin secretion) \\
\texttt{VALID\_UNLEARNABLE} & 2889 & (smooth muscle length) --[\texttt{cause\_of}]$\rightarrow$ (gastric function) \\
\texttt{VALID\_UNLEARNABLE} & 4087 & (intestinal ffa release profile) --[\texttt{cause\_of}]$\rightarrow$ (hormone response) \\
\texttt{VALID\_UNLEARNABLE} & 4366 & (delayed gastric emptying) --[\texttt{cause\_of}]$\rightarrow$ (postoperative complications) \\
\texttt{VALID\_UNLEARNABLE} & 166 & (rikkunshito) --[\texttt{has\_disposition}]$\rightarrow$ (5-HT3 receptor antagonist) \\
\texttt{VALID\_UNLEARNABLE} & 2738 & (gip) --[\texttt{has\_disposition}]$\rightarrow$ (glucose-dependent) \\
\texttt{VALID\_UNLEARNABLE} & 3175 & (cyclopeptides) --[\texttt{has\_disposition}]$\rightarrow$ (inhibitor) \\
\texttt{VALID\_UNLEARNABLE} & 4640 & (cinitapride) --[\texttt{has\_disposition}]$\rightarrow$ (dopamine antagonist) \\
\texttt{VALID\_UNLEARNABLE} & 176 & (wirsung-pancreato-gastro-anastomosis (wpga)) --[\texttt{has\_modification}]$\rightarrow$ (laparoscopic) \\
\texttt{VALID\_UNLEARNABLE} & 4788 & (gastric volume and content) --[\texttt{has\_modification}]$\rightarrow$ (high) \\
\texttt{VALID\_UNLEARNABLE} & 360 & (neuronal nos (nnos)) --[\texttt{has\_finding\_site}]$\rightarrow$ (enteric nervous system) \\
\texttt{VALID\_UNLEARNABLE} & 583 & (acetycholinesterase (ache)) --[\texttt{has\_finding\_site}]$\rightarrow$ (gastric nerve) \\
\texttt{VALID\_UNLEARNABLE} & 1764 & (enlarged neurons) --[\texttt{has\_finding\_site}]$\rightarrow$ (myenteric plexus) \\
\texttt{VALID\_UNLEARNABLE} & 2211 & (gastric emptying t$\frac{1}{2}$) --[\texttt{has\_finding\_site}]$\rightarrow$ (stomach) \\
\texttt{VALID\_UNLEARNABLE} & 2294 & (vasoactive intestinal peptide-immunoreactive neurons) --[\texttt{has\_finding\_site}]$\rightarrow$ (mucosal) \\
\texttt{VALID\_UNLEARNABLE} & 3007 & (functional dyspepsia (fd)) --[\texttt{has\_finding\_site}]$\rightarrow$ (small intestine) \\
\texttt{VALID\_UNLEARNABLE} & 3573 & (slow colonic transit) --[\texttt{has\_finding\_site}]$\rightarrow$ (colon) \\
\texttt{VALID\_UNLEARNABLE} & 3984 & (dyspepsia) --[\texttt{has\_finding\_site}]$\rightarrow$ (stomach) \\
\texttt{VALID\_UNLEARNABLE} & 4424 & (heme oxygenase (ho) isoforms) --[\texttt{has\_finding\_site}]$\rightarrow$ (neurons) \\
\texttt{VALID\_UNLEARNABLE} & 4490 & (small intestinal microbial activity) --[\texttt{has\_finding\_site}]$\rightarrow$ (small intestine) \\
\texttt{VALID\_UNLEARNABLE} & 4493 & (inflammatory response in the muscularis) --[\texttt{has\_finding\_site}]$\rightarrow$ (intestine) \\
\texttt{VALID\_UNLEARNABLE} & 5681 & (muscle contractility) --[\texttt{has\_finding\_site}]$\rightarrow$ (gastric smooth muscle) \\
\texttt{VALID\_UNLEARNABLE} & 5927 & (peptide yy (pyy)) --[\texttt{has\_finding\_site}]$\rightarrow$ (intestinal cell) \\
\texttt{VALID\_UNLEARNABLE} & 2335 & (pancreatoduodenectomy (pd)) --[\texttt{has\_direct\_procedure\_site}]$\rightarrow$ (duodenum) \\
\texttt{VALID\_UNLEARNABLE} & 3205 & (jejunal interposition) --[\texttt{has\_direct\_procedure\_site}]$\rightarrow$ (small bowel) \\
\texttt{VALID\_UNLEARNABLE} & 3449 & (st36) --[\texttt{has\_direct\_procedure\_site}]$\rightarrow$ (lower extremity) \\
\texttt{VALID\_UNLEARNABLE} & 3786 & (gastric outlet obstruction) --[\texttt{has\_direct\_procedure\_site}]$\rightarrow$ (pyloric) \\
\texttt{VALID\_UNLEARNABLE} & 3891 & (jejunal manometry) --[\texttt{finding\_site\_of}]$\rightarrow$ (small bowel) \\
\texttt{VALID\_UNLEARNABLE} & 2568 & (gastric outlet obstruction) --[\texttt{possibly\_equivalent\_to}]$\rightarrow$ (duodenal obstruction) \\
\texttt{VALID\_UNLEARNABLE} & 2678 & (enterra therapy) --[\texttt{inverse\_isa}]$\rightarrow$ (gastric stimulation) \\
\texttt{VALID\_UNLEARNABLE} & 5703 & (refractory fd) --[\texttt{inverse\_isa}]$\rightarrow$ (functional dyspepsia) \\
\texttt{VALID\_UNLEARNABLE} & 199 & (ketones) --[\texttt{plays\_role}]$\rightarrow$ (insulin secretion) \\
\texttt{VALID\_UNLEARNABLE} & 307 & (microbially-derived neurotransmitters) --[\texttt{plays\_role}]$\rightarrow$ (gut-brain axis pathways) \\
\texttt{VALID\_UNLEARNABLE} & 1061 & (hereditary transthyretin amyloid (attr) amyloidosis) --[\texttt{plays\_role}]$\rightarrow$ (liver) \\
\texttt{VALID\_UNLEARNABLE} & 1162 & (granisetron transdermal system) --[\texttt{plays\_role}]$\rightarrow$ (gastrointestinal therapeutic) \\
\texttt{VALID\_UNLEARNABLE} & 4675 & (moxibustion) --[\texttt{plays\_role}]$\rightarrow$ (treatment) \\
\texttt{VALID\_UNLEARNABLE} & 4903 & (gastric decompression) --[\texttt{plays\_role}]$\rightarrow$ (abdominal decompression) \\
\texttt{VALID\_UNLEARNABLE} & 3295 & (point-of-care gastric ultrasound) --[\texttt{interprets}]$\rightarrow$ (gastric motility) \\
\texttt{VALID\_UNLEARNABLE} & 5113 & (enteral nutrition) --[\texttt{focus\_of}]$\rightarrow$ (postoperative management) \\
\texttt{VALID\_UNLEARNABLE} & 5209 & (gastric electrical stimulants) --[\texttt{focus\_of}]$\rightarrow$ (gastric emptying) \\

\end{longtable}
}

\subsubsection{LLM Judge Prompt Design}
\label{app:judge_prompt_design}

The adaptive repair pipeline uses two LLM judges with different roles. Judge 1 is applied immediately after the initial one-time SFT validation step and performs a broad failure audit over all incorrectly answered validation examples. Judge 2 is applied later, only to quarantine candidates that repeatedly fail after multiple targeted repair rounds. Thus, Judge 1 decides why an initial model failure occurred, whereas Judge 2 decides how to handle repeatedly failing cases after repair history is known.

\paragraph{Judge 1 decision tree.}
Judge 1 follows a three-step decision tree and stops at the first failing step:
\begin{itemize}
    \item \textbf{Triple validity check:} Judge 1 first checks whether the primary triple is factually valid and semantically precise. If the triple is unsupported, vague, ambiguous, or invalid, the example is labeled \texttt{INVALID\_TRIPLE}.

    \item \textbf{Validation-question alignment check:} If the triple is valid, Judge 1 checks whether the validation question is aligned with the primary triple. The keyed answer must depend on the primary triple, the ground-truth answer must be correct, and no other option should be equally defensible. If this check fails, the example is labeled \texttt{BAD\_VAL}.

    \item \textbf{Missing-knowledge assignment:} If both the triple and the validation question pass, the failure is labeled \texttt{MISSING\_KNOWLEDGE}. This means the model failure reflects a genuine knowledge gap suitable for targeted repair.
\end{itemize}

For KG-grounded examples, Judge 1 evaluates only the primary triple and the paired validation question. For CG-grounded examples, the same decision tree is used, but the prompt additionally includes supporting triples. These supporting triples are treated as background context only: the keyed answer must still hinge on the primary triple.
\paragraph{Judge 2 decision settings.}
Judge 2 is applied only to quarantine candidates. We use two Judge 2 settings because not all quarantine candidates have the same history.

\begin{itemize} 
    \item \emph{Case A} applies to triples that were already labeled as \texttt{MISSING\_KNOWLEDGE} by Judge 1. Since these examples have already passed the initial triple-validity and question-alignment checks, Judge 2 does not re-evaluate the question. It only reviews the primary triple and chooses between \texttt{INVALID\_TRIPLE} and \texttt{VALID\_UNLEARNABLE}. A triple is dropped if it is invalid, unsupported, vague, or semantically imprecise under closer inspection. Otherwise, it is retained as a valid but currently unlearnable case and excluded from further repair SFT.

    \item \emph{Case B} applies to triples that were not part of the original Judge-1-confirmed missing-knowledge set and became quarantine candidates through later repair dynamics, such as regressions or new errors. These cases were not previously audited by Judge 1; hence, Judge 2 performs a compact final triage over three possible outcomes: \texttt{INVALID\_TRIPLE}, \texttt{BAD\_VAL}, or \texttt{VALID\_UNLEARNABLE}. This differs from Judge 1 because the goal is not to identify ordinary missing knowledge at the start of training; instead, the goal is to decide what to do with a repeatedly failing quarantine candidate after multiple repair attempts. If the triple is invalid or too vague, it is dropped. If the triple is valid but the validation question is misaligned or ambiguous, the question is regenerated. If both the triple and question are acceptable, the case is labeled valid but currently unlearnable.
\end{itemize}

\subsubsection{Judge 1 Prompt Templates}
\label{app:judge1_prompt_templates}

\paragraph{KG-grounded Judge 1 prompt.}
For KG-grounded examples, Judge 1 receives the primary triple, the validation question, answer options, and the ground-truth answer.

\begin{tcolorbox}[title=KG-grounded Judge 1 prompt,breakable]
You are a medical knowledge-graph evaluator for disease-specific biomedical knowledge-graph QA.
The disease domain is \textbf{\{disease\}}.
Your job is to determine why a model failed a validation question item. 

You must follow the decision tree below IN ORDER. Stop at the first failing step.

\medskip
\textbf{Step 1: Is the triple factually valid and supported?}

Evaluate ONLY the triple \texttt{(head, relation, tail)}. Use general medical/scientific knowledge.

The triple passes Step 1 only if all of the following hold:
\begin{enumerate}
    \item \textbf{Factually valid:} the relation represents a defensible factual claim in medicine.
    \item \textbf{Semantically precise:} the triple is not vague, ambiguous, or underspecified.
    \item The head and tail form a clear, interpretable pair under the stated relation.
\end{enumerate}

Reject as \texttt{INVALID\_TRIPLE} if the relation is factually wrong, nonsensical, unsupported, contested, semantically vague, or too underspecified to support a well-defined factual claim.

Do not reject merely because the triple is obscure, narrow, unfamiliar, or because the model failed the question.

If Step 1 fails, output \texttt{INVALID\_TRIPLE} and stop.

\medskip
\textbf{Step 2: Is the test question aligned with the triple?}

Evaluate only if Step 1 passed.

A question is aligned if all of the following hold:
\begin{enumerate}
    \item Answering correctly requires knowing or applying the triple.
    \item The expected correct answer is medically correct given the stem and options.
    \item The expected answer is uniquely correct.
    \item Clinical vignettes are allowed if the correct answer logically depends on the triple.
\end{enumerate}

A question is not aligned if it can be answered without the triple, if the labeled answer is wrong or debatable, if multiple options are equally defensible, or if the question tests a different relationship than the triple states.

If Step 2 fails, output \texttt{BAD\_TEST\_QUESTION} and stop.

\medskip
\textbf{Step 3: Missing knowledge}

If Step 1 and Step 2 both pass, output \texttt{MISSING\_SKILL}.  

\medskip
\textbf{Important rules:}
\begin{itemize}
    \item Do not skip steps.
    \item Be conservative on \texttt{INVALID\_TRIPLE}; when uncertain about factual invalidity, prefer assuming the triple is valid.
    \item However, reject semantically vague triples even if they are not outright false.
    \item Be conservative on \texttt{BAD\_TEST\_QUESTION}; minor wording imperfections are acceptable if the item is fair and the ground truth is uniquely correct.
\end{itemize}

\medskip
\textbf{Test item to evaluate:}

\begin{verbatim}
path_idx: {path_idx}

triple:
  head: {head}
  relation: {relation}
  tail: {tail}

test_question:
{question_text}

answer_options:
{options_text}

expected_correct_answer: {ground_truth}

\end{verbatim}
\end{tcolorbox}

\paragraph{CG-grounded Judge 1 prompt.}
For CG-grounded examples, the Judge 1 prompt follows the same three-step decision tree, but it additionally receives supporting triples. The supporting triples are used only as background grounding for the vignette, options, and distractors. Step 1 judges only the primary triple, while Step 2 checks whether the keyed answer depends on the primary triple and remains valid given both the primary and supporting triples.

\begin{tcolorbox}[title=CG-grounded Judge 1 prompt,breakable]
You are a medical knowledge-graph evaluator for disease-specific biomedical knowledge-graph QA. The disease domain is \textbf{\{disease\}}.
Your job is to determine why a model failed a validation question item.

Validation questions are CONTEXT-GROUNDED. Each item is built from one primary triple \texttt{(head, relation, tail)}. Additional supporting triples describe the surrounding local graph neighborhood. They were provided to the question author as background to enrich the clinical vignette and distractors. The keyed correct answer must be driven by the primary triple.

You must follow the decision tree below IN ORDER. Stop at the first failing step.

\medskip
\textbf{Step 1: Is the primary triple factually valid and supported?}

Evaluate only the primary triple. The supporting triples are context only and should not be judged in this step.

The primary triple passes Step 1 only if all of the following hold:
\begin{enumerate}
    \item \textbf{Factually valid:} the relation represents a defensible factual claim in medicine.
    \item \textbf{Semantically precise:} the triple is not vague, ambiguous, or underspecified.
    \item The head and tail form a clear, interpretable pair under the stated relation.
\end{enumerate}

Reject as \texttt{INVALID\_TRIPLE} if the relation is factually wrong, nonsensical, unsupported, contested, semantically vague, or too underspecified to support a well-defined factual claim.

Do not reject merely because the triple is obscure, narrow, unfamiliar, or because the model failed the question.

If Step 1 fails, output \texttt{INVALID\_TRIPLE} and stop.

\medskip
\textbf{Step 2: Is the test question valid given the primary triple and supporting triples?}

Evaluate only if Step 1 passed. Use the primary triple as the fact under test and the supporting triples as background grounding.

A question is valid if:
\begin{enumerate}
    \item Answering correctly requires knowing or applying the primary triple.
    \item The expected correct answer is medically correct given the stem, options, primary triple, and supporting triples.
    \item The expected answer is uniquely correct.
    \item The supporting triples do not contradict the primary triple or keyed answer.
\end{enumerate}

A question is not aligned if it can be answered without the primary triple, if it is answerable from only the supporting triples or unrelated general knowledge, if the ground truth is wrong or debatable, if multiple options are equally defensible, or if the question tests a different relationship than the primary triple states.

If Step 2 fails, output \texttt{BAD\_TEST\_QUESTION} and stop.

\medskip
\textbf{Step 3: Missing knowledge}

If Step 1 and Step 2 both pass, output \texttt{MISSING\_SKILL}.

\medskip
\textbf{Important rules:}
\begin{itemize}
    \item Do not skip steps.
    \item Judge only the primary triple's validity in Step 1.
    \item Supporting triples are background context.
    \item Be conservative on \texttt{INVALID\_TRIPLE} and \texttt{BAD\_TEST\_QUESTION}.
\end{itemize}

\medskip

\medskip
\textbf{Test item to evaluate:}

\begin{verbatim}
path_idx: {path_idx}

primary_triple:
  head: {head}
  relation: {relation}
  tail: {tail}

supporting_triples:
{context_triples}

test_question:
{question_text}

answer_options:
{options_text}

expected_correct_answer: {ground_truth}

\end{verbatim}
\end{tcolorbox}

\subsubsection{Judge 2 Prompt Templates}
\label{app:judge2_prompt_templates}

\paragraph{Judge 2 Case A: previously audited missing-knowledge triples.}
Case A applies to triples that were already labeled as \texttt{MISSING\_KNOWLEDGE} by Judge 1. 

\begin{tcolorbox}[title=Judge 2 Case A prompt,breakable]
You are a medical knowledge-graph evaluator reviewing persistent model failures.

These triples were targeted across multiple rounds of repair SFT, yet the model still fails the paired test item. Your only job is to decide whether the primary triple itself should be dropped or accepted as valid but currently unlearnable.

The primary triple is from a literature chunk. Supporting triples, if provided, are from the same chunk and are given only as background to help interpret the primary triple. Judge only the primary triple.

\medskip
\textbf{Binary decision:}

Evaluate only the primary triple \texttt{(head, relation, tail)}.

Choose \texttt{INVALID\_TRIPLE} if the triple is factually invalid, unsupported, semantically vague, ambiguous, underspecified, directionally incorrect, or if the head and tail do not form a clear interpretable pair under the stated relation.

Choose \texttt{VALID\_UNLEARNABLE} if the triple is factually valid and semantically precise, but the model still fails after repeated targeted repair.

\medskip
\textbf{Candidate to evaluate:}

\begin{verbatim}
path_idx: {path_idx}

primary_triple:
  head: {head}
  relation: {relation}
  tail: {tail}

supporting_triples: {context_triples}
\end{verbatim}
\end{tcolorbox}

\paragraph{Judge 2 Case B: quarantine candidates not previously audited by Judge 1.}
Case B applies to triples that were not part of the initial Judge-1-confirmed missing-knowledge set.

\begin{tcolorbox}[title=Judge 2 Case B prompt,breakable]

You are a medical knowledge-graph evaluator reviewing repeatedly failing validation cases.
They emerged as errors during continual repair. You must perform a triage using both the primary triple and the paired validation question.
The primary triple is from a literature chunk. Supporting triples, if provided, are from the
same chunk and are given only as background to help interpret the primary triple. Judge only the
primary triple.

Your goal is to decide whether the case should be dropped because the triple is invalid, regenerated because the validation question is not aligned, or retained as valid but currently unlearnable.

Follow the decision tree in order. Stop at the first matching label.

\medskip
\textbf{Step 1: INVALID\_TRIPLE}

Evaluate the primary triple \texttt{(head, relation, tail)}. Choose \texttt{INVALID\_TRIPLE} if the triple is factually invalid, unsupported, semantically vague, ambiguous, underspecified, directionally incorrect, or if the head and tail do not form a clear interpretable pair under the stated relation.

If Step 1 matches, stop.

\medskip
\textbf{Step 2: BAD\_VAL}

Evaluate the paired validation question only if the triple is valid.

Choose \texttt{BAD\_VAL} if the question is not aligned with the triple, if the ground-truth answer is wrong or debatable, if multiple answer options are equally defensible, if the question can be answered without using the primary triple, or if the question tests a different relationship than the triple states.

If Step 2 matches, stop.

\medskip
\textbf{Step 3: VALID\_UNLEARNABLE}

If the triple is valid and the test question is fair, choose \texttt{VALID\_UNLEARNABLE}. This means the case remains valid but was not reliably learned by the current model and repair strategy.

Be conservative on \texttt{DROP\_TRIPLE} and \texttt{BAD\_TEST\_QUESTION}; when genuinely uncertain, prefer \texttt{VALID\_UNLEARNABLE}.

\medskip

\textbf{Candidate to evaluate:}
\begin{verbatim}
path_idx: {path_idx}

triple:
  head: {head}
  relation: {relation}
  tail: {tail}

test_question:
{question_text}

answer_options:
{options_text}

expected_correct_answer: {ground_truth}
supporting_triples: {context_triples}
\end{verbatim}
\end{tcolorbox}

\subsection{Evaluation}
\label{app:Evaluation}

The following prompt is used for MCQ inference during evaluation:

\begin{tcolorbox}[
    colback=green!4,
    colframe=green!45!black,
    title={Evaluation Inference Prompt},
    fonttitle=\bfseries,
    arc=1.5mm,
    boxrule=0.5pt
]
You are a medical expert in <gastroparesis/diabetes>. You are presented with a multiple-choice clinical question. First, think through the problem systematically using \texttt{<think>...</think>} tags. Then provide a brief explanation and the final answer. Your final answer must be one of A, B, C, or D, and must follow the format: \texttt{Final Answer: [A/B/C/D]}.

\vspace{1mm}

\texttt{<Question>}\\
\texttt{[Clinical Vignette]}\\
\texttt{</Question>}

\vspace{1mm}

\texttt{<Options>}\\
\texttt{A. [Option]}\\
\texttt{B. [Option]}\\
\texttt{C. [Option]}\\
\texttt{D. [Option]}\\
\texttt{</Options>}
\end{tcolorbox}

\subsection{Additional Experimental Results}
\label{app:results}

\subsubsection{Full Model Comparisons for Diabetes}
\label{app:diab_results}

Table~\ref{tab:diabetes_full_comparison} reports accuracy across all model variants for diabetes.

\FloatBarrier
\begin{table*}[h]
\centering
\caption{Diabetes higher-hop QA accuracy across all model variants. $\Delta_{\mathrm{RL}}$ denotes the per-hop gain from RL over the corresponding non-RL checkpoint. Bold accuracy values indicate the highest accuracy for each evaluation setting and hop depth, and underlined values indicate the second-highest accuracy.}
\label{tab:diabetes_full_comparison}
\footnotesize
\renewcommand{\arraystretch}{1.2}
\setlength{\tabcolsep}{2.2pt}
\begin{tabular*}{\textwidth}{@{\extracolsep{\fill}}lcccccccccccc}
\toprule
\textbf{Model}
& \multicolumn{6}{c}{\textbf{KG-grounded QA}}
& \multicolumn{6}{c}{\textbf{CG-grounded QA}} \\
\cmidrule(lr){2-7}
\cmidrule(lr){8-13}
& \textbf{3-hop} & \textbf{$\Delta_3$}
& \textbf{4-hop} & \textbf{$\Delta_4$}
& \textbf{5-hop} & \textbf{$\Delta_5$}
& \textbf{3-hop} & \textbf{$\Delta_3$}
& \textbf{4-hop} & \textbf{$\Delta_4$}
& \textbf{5-hop} & \textbf{$\Delta_5$} \\
\midrule
Base 
& 88.98 &  & 86.10 &  & 81.21 &  
& 89.76 &  & 86.79 &  & 81.86 &  \\
\midrule

\rowcolor{kgplain}
KGModel 
& 95.55 &  & 93.04 &  & 89.83 &  
& 94.35 &  & 92.35 &  & 87.30 &  \\

\rowcolor{kgplain}
KGModel-RL 
& 95.65 & +0.10 & 93.56 & +0.52 & 90.62 & +0.79
& 94.53 & +0.18 & 92.41 & +0.06 & 87.69 & +0.39 \\

\rowcolor{kgrepair}
KGModel-Repaired 
& 96.74 &  & 94.10 &  & 91.49 &  
& 95.58 &  & 93.05 &  & 89.26 &  \\

\rowcolor{kgrepair}
KGModel-Repaired-RL 
& \underline{98.91} & \textbf{+2.17} & \underline{96.51} & \textbf{+2.41} & 92.80 & \textbf{+1.31}
& 96.66 & \textbf{+1.08} & 93.90 & \textbf{+0.85} & 90.12 & \textbf{+0.86} \\
\midrule

\rowcolor{cgplain}
CGModel 
& 97.14 &  & 95.70 &  & 91.85 &  
& 95.85 &  & 93.92 &  & 89.96 &  \\

\rowcolor{cgplain}
CGModel-RL 
& 97.05 & -0.09 & 95.90 & +0.20 & 91.12 & -0.73
& 95.99 & +0.14 & \underline{94.35} & +0.43 & 90.21 & +0.25 \\

\rowcolor{cgrepair}
CGModel-Repaired 
& 97.43 &  & 95.91 &  & \underline{92.94} &  
& \underline{96.89} &  & 94.25 &  & \underline{91.81} &  \\

\rowcolor{cgrepair}
CGModel-Repaired-RL 
& \textbf{98.96} & \textbf{+1.53} & \textbf{97.54} & \textbf{+1.63} & \textbf{94.05} & \textbf{+1.11}
& \textbf{98.79} & \textbf{+1.90} & \textbf{97.51} & \textbf{+3.26} & \textbf{93.95} & \textbf{+2.14} \\
\bottomrule
\end{tabular*}
\renewcommand{\arraystretch}{1.1}
\setlength{\tabcolsep}{6pt}
\normalsize
\end{table*}

\subsubsection{Option-Shuffling Robustness Results}
\label{app:option_shuffling_results}

The remaining option-shuffling robustness results are reported in Tables~\ref{tab:shuffle_gastro_kg_full}--\ref{tab:shuffle_diab_cg_full}, covering the Gastroparesis KG-grounded, Diabetes KG-grounded, and Diabetes CG-grounded settings. Almost all the models remain relatively stable. The main exception is CGModel-RL on Diabetes KG-grounded QA, where the 3-hop shuffled accuracy drops by $2.07$ percentage points. However, this is an isolated case rather than a consistent trend across models or settings. 

\begin{table*}[h]
\centering
\caption{Option-shuffling robustness on Gastroparesis KG-grounded QA. $\Delta$ denotes shuffled accuracy minus original accuracy in percentage points.}
\label{tab:shuffle_gastro_kg_full}
\small
\renewcommand{\arraystretch}{1.1}
\setlength{\tabcolsep}{4.2pt}
\begin{tabular*}{\textwidth}{@{\extracolsep{\fill}}lcccccccccc}
\toprule
\textbf{Model} 
& \multicolumn{3}{c}{\textbf{3-hop}}
& \multicolumn{3}{c}{\textbf{4-hop}}
& \multicolumn{3}{c}{\textbf{5-hop}}
& \textbf{Avg.} \\
\cmidrule(lr){2-4}
\cmidrule(lr){5-7}
\cmidrule(lr){8-10}
& \textbf{Orig.} & \textbf{Shuf.} & \textbf{$\Delta$}
& \textbf{Orig.} & \textbf{Shuf.} & \textbf{$\Delta$}
& \textbf{Orig.} & \textbf{Shuf.} & \textbf{$\Delta$}
& \textbf{$\Delta$} \\
\midrule
Base & 91.01 & 91.19 & +0.18 & 89.10 & 89.22 & +0.12 & 87.04 & 86.95 & -0.09 & +0.07\\
KGModel & 95.30 & 95.72 & +0.42 & 91.38 & 92.01 & +0.63 & 89.81 & 90.60 & +0.79& +0.61
\\
KGModel-RL & 95.50 & 94.87 & -0.63 & 92.59 & 91.82 & -0.77 & 89.96 & 89.45 & -0.51 & -0.64\\
KGModel-Repaired & 96.23 & 96.56 & +0.33 & 92.38 & 93.26 & +0.89 & 90.85 & 91.32 & +0.47 & +0.56 \\
KGModel-Repaired-RL & 97.90 & 97.44 & -0.46 & 94.33 & 93.58 & -0.75 & 92.50 & 91.49 & -1.01 & -0.74 \\
CGModel & 97.28 & 96.67 & -0.60 & 93.58 & 93.47 & -0.10 & 91.96 & 92.05 & +0.09 & -0.20 \\
CGModel-RL & 97.78 & 96.69 & -1.09 & 94.36 & 93.38 & -0.99 & 92.11 & 91.59 & -0.53 & -0.87 \\
CGModel-Repaired & 97.39 & 97.11 & -0.28 & 93.67 & 93.94 & +0.26 & 92.08 & 92.86 & +0.79 & +0.26 \\
CGModel-Repaired-RL & 98.88 & 97.98 & -0.90 & 95.56 & 94.69 & -0.87 & 94.19 & 93.17 & -1.02 & -0.93 \\
\bottomrule
\end{tabular*}
\renewcommand{\arraystretch}{1.0}
\setlength{\tabcolsep}{6pt}
\end{table*}

\begin{table}[h]
\centering
\caption{Option-shuffling robustness on Diabetes KG-grounded QA. $\Delta$ denotes shuffled accuracy minus original accuracy in percentage points.}
\label{tab:shuffle_diab_kg_full}
\small
\renewcommand{\arraystretch}{1.1}
\setlength{\tabcolsep}{4.2pt}
\begin{tabular*}{\textwidth}{@{\extracolsep{\fill}}lcccccccccc}
\toprule
\textbf{Model} 
& \multicolumn{3}{c}{\textbf{3-hop}}
& \multicolumn{3}{c}{\textbf{4-hop}}
& \multicolumn{3}{c}{\textbf{5-hop}}
& \textbf{Avg.} \\
\cmidrule(lr){2-4}
\cmidrule(lr){5-7}
\cmidrule(lr){8-10}
& \textbf{Orig.} & \textbf{Shuf.} & \textbf{$\Delta$}
& \textbf{Orig.} & \textbf{Shuf.} & \textbf{$\Delta$}
& \textbf{Orig.} & \textbf{Shuf.} & \textbf{$\Delta$}
& \textbf{$\Delta$} \\
\midrule
Base & 88.98 & 89.35 & +0.37 & 86.10 & 86.08 & -0.02 & 81.21 & 81.97 & +0.76 & +0.21 \\
KGModel & 95.55 & 95.42 & -0.13 & 93.04 & 92.70 & -0.34 & 89.83 & 90.06 & +0.23 & -0.08 \\
KGModel-RL & 95.65 & 94.70 & -0.95 & 93.56 & 93.11 & -0.45 & 90.62 & 89.82 & -0.80 & -0.73 \\
KGModel-Repaired & 96.74 & 96.31 & -0.42 & 94.10 & 93.89 & -0.21 & 91.49 & 91.31 & -0.17 & -0.27 \\
KGModel-Repaired-RL & 98.91 & 99.03 & +0.12 & 96.51 & 96.01 & -0.50 & 92.80 & 91.86 & -0.94 & -0.44 \\
CGModel & 97.14 & 94.86 & -1.20 & 95.70 & 94.75 & -0.95 & 91.85 & 91.88 &+0.03 & -0.71 \\
CGModel-RL & 97.05 & 94.98 & -2.07 & 95.90 & 94.70 & -1.20 & 91.12 & 90.35 & -0.77 & -1.35 \\
CGModel-Repaired & 97.43 & 96.38 & -1.05 & 95.91 & 95.54 & -0.36 & 92.94 & 91.75 & -1.19 & -0.87\\
CGModel-Repaired-RL & 98.96 & 98.21 & -0.75 & 97.54 & 96.59 & -0.95 & 94.05 & 93.01 & -1.04 & -0.91 \\
\bottomrule
\end{tabular*}
\renewcommand{\arraystretch}{1.0}
\setlength{\tabcolsep}{6pt}
\end{table}

\FloatBarrier

\begin{table}[h]
\centering
\caption{Option-shuffling robustness on Diabetes CG-grounded QA. $\Delta$ denotes shuffled accuracy minus original accuracy in percentage points.}
\label{tab:shuffle_diab_cg_full}
\small
\renewcommand{\arraystretch}{1.1}
\setlength{\tabcolsep}{4.2pt}
\begin{tabular*}{\textwidth}{@{\extracolsep{\fill}}lcccccccccc}
\toprule
\textbf{Model} 
& \multicolumn{3}{c}{\textbf{3-hop}}
& \multicolumn{3}{c}{\textbf{4-hop}}
& \multicolumn{3}{c}{\textbf{5-hop}}
& \textbf{Avg.} \\
\cmidrule(lr){2-4}
\cmidrule(lr){5-7}
\cmidrule(lr){8-10}
& \textbf{Orig.} & \textbf{Shuf.} & \textbf{$\Delta$}
& \textbf{Orig.} & \textbf{Shuf.} & \textbf{$\Delta$}
& \textbf{Orig.} & \textbf{Shuf.} & \textbf{$\Delta$}
& \textbf{$\Delta$} \\
\midrule
Base & 89.76 & 89.41 & -0.35 & 86.79 & 86.92 & +0.13 & 81.86 & 81.62 & -0.24 & -0.15 \\
KGModel & 94.35 & 94.06 & -0.29 & 92.35 & 92.24 & -0.11 & 87.30 & 87.53 & +0.23 & -0.06 \\
KGModel-RL & 94.53 & 93.54 & -0.99 & 92.41 & 91.60 & -0.81 & 87.69 & 86.91 & -0.78 & -0.86 \\
KGModel-Repaired & 95.58 & 95.03 & -0.55 & 93.05 & 93.66 & -0.61 & 89.26 & 88.88 & -0.38 & -0.56 \\
KGModel-Repaired-RL & 96.66 & 96.02 & -0.64 & 93.90 & 93.98 & +0.08 & 90.12 & 89.06 & -1.06 & -0.54 \\
CGModel & 95.85 & 95.41 & -0.44 & 93.92 & 93.51 & -0.41 & 89.96 & 88.81 & -1.15 & -0.67 \\
CGModel-RL & 95.99 & 94.83 & -1.16 & 94.35 & 93.70 & -0.65 & 90.21 & 87.89 & -0.80 & -0.87 \\
CGModel-Repaired & 96.89 & 95.83 & -1.06 & 94.25 & 94.38 & -0.13 & 91.81 & 90.87 & -0.94 & -0.71 \\
CGModel-Repaired-RL & 98.79 & 98.85 & +0.06 & 97.51 & 96.48 & -1.03 & 93.95 & 93.07 & -0.88 & -0.62 \\
\bottomrule
\end{tabular*}
\renewcommand{\arraystretch}{1.0}
\setlength{\tabcolsep}{6pt}
\end{table}

\end{document}